\pdfoutput=1 

\documentclass[journal,fleqn]{IEEEtran}

\usepackage{lipsum}

\usepackage[resetlabels,labeled]{multibib}

\newcites{Sup}{References}

\usepackage[ruled,,linesnumbered]{algorithm2e}
\usepackage{algpseudocode}
\usepackage{mathtools}
\usepackage{subcaption}
\usepackage{blindtext}
\usepackage{verbatim} 
\usepackage{dcolumn}
\usepackage{multirow}
\usepackage[table]{xcolor}
\usepackage{color,soul}
\usepackage[normalem]{ulem} 
\let\oldnl\nl
\newcommand{\nonl}{\renewcommand{\nl}{\let\nl\oldnl}}

\newenvironment{proc}[1][htb]
  {
   \begin{algorithm}[#1]%
  }{\end{algorithm}}

\usepackage{amssymb}

\usepackage{graphicx}

\usepackage[colorlinks=true,urlcolor=blue,citecolor=blue,linkcolor=blue,bookmarks=true]{hyperref}

\usepackage{float}
\newcommand{\minus}{\scalebox{0.5}[1.0]{$-$}}

\usepackage{tikz}
\usepackage{textcomp}

\newcommand\copyrighttext{%
  \fontsize{7}{1}\selectfont \textcopyright 2022 IEEE. Personal use of this material is permitted.  Permission from IEEE must be obtained for all other uses, in any current or future media, including reprinting/republishing this material for advertising or promotional purposes, creating new collective works, for resale or redistribution to servers or lists, or reuse of any copyrighted component of this work in other works. 
  DOI: \href{https://doi.org/10.1109/TCYB.2022.3164399}{10.1109/TCYB.2022.3164399}}
\newcommand\copyrightnotice{%
\begin{tikzpicture}[remember picture,overlay]
\node[anchor=south,yshift=10pt] at (current page.south) {\fbox{\parbox{\dimexpr\textwidth-\fboxsep-\fboxrule\relax}{\copyrighttext}}};
\end{tikzpicture}%
}

\begin{document}

%
\title{Scalable Transfer Evolutionary Optimization: Coping with Big Task-Instances}
%
%
%

\author{Mojtaba~Shakeri, Erfan~Miahi, Abhishek~Gupta, \emph{Senior Member}, \emph{IEEE}, and Yew-Soon~Ong, \emph{Fellow}, \emph{IEEE} 
\thanks{This work was supported by the A*STAR Cyber-Physical Production System (CPPS) -- Towards Contextual and Intelligent Response Research Program, under the RIE2020 IAF-PP Grant A19C1a0018, and Model Factory@SIMTech. Abhishek Gupta was also supported in part by the AI3 seed grant C211118016 on Upside-Down Multi-Objective Bayesian Optimization for Few-Shot Design. \emph{(Corresponding author: Abhishek Gupta)}}
\thanks{M. Shakeri is with Singapore Institute of Manufacturing Technology, Singapore 138634, and also with Computer Engineering Department, Faculty of Engineering, University of Guilan, Rasht 13769-41996, Iran (e-mail: Mojtaba\_Shakeri@simtech.a-star.edu.sg; shakeri@guilan.ac.ir).}
\thanks{E. Miahi is with Department of Computing Science, Faculty of Science, University of Alberta, Canada T6G 2E8 (e-mail: miahi@ualberta.ca).}
\thanks{A. Gupta is with Singapore Institute of Manufacturing Technology, Singapore 138634 (abhishek\_gupta@simtech.a-star.edu.sg).}
\thanks{Y.-S. Ong is with the School of Computer Science and Engineering, Nanyang Technological University, Singapore, and also with the Agency for Science, Technology and Research (A*STAR), Singapore (e-mail: asysong@ntu.edu.sg; ongyewsoon@hq.a-star.edu.sg).}
}

%
%

\markboth{Published in IEEE TRANSACTIONS ON CYBERNETICS}%
{Shell \MakeLowercase{\textit{et al.}}: Bare Demo of IEEEtran.cls for IEEE Journals}
%




\maketitle
\copyrightnotice
\begin{abstract}
In today's digital world, we are faced with an explosion of data and models produced and manipulated by numerous large-scale cloud-based applications. Under such settings, existing transfer evolutionary optimization frameworks grapple with \emph{simultaneously} satisfying two important quality attributes, namely (1) \emph{scalability} against a growing number of source tasks and (2) \emph{online learning agility} against sparsity of relevant sources to the target task of interest. Satisfying these attributes shall facilitate practical deployment of transfer optimization to scenarios with big task-instances, while curbing the threat of negative transfer. While applications of existing algorithms are limited to tens of source tasks, in this paper, we take a quantum leap forward in enabling more than two orders of magnitude scale-up in the number of tasks; i.e., we efficiently handle scenarios beyond 1000 source task-instances. We devise a novel transfer evolutionary optimization framework comprising two co-evolving species for joint evolutions in the space of source knowledge and in the search space of solutions to the target problem. In particular, co-evolution enables the learned knowledge to be orchestrated on the fly, expediting convergence in the target optimization task. We have conducted an extensive series of experiments across a set of practically motivated discrete and continuous optimization examples comprising a large number of source task-instances, of which only a small fraction indicate source-target relatedness. The experimental results show that not only does our proposed framework scale efficiently with a growing number of source tasks but is also effective in capturing relevant knowledge against sparsity of related sources, fulfilling the two salient features of scalability and online learning agility. 
\end{abstract}

\begin{IEEEkeywords}
Transfer evolutionary optimization, big task-instances, scalability, online learning agility.   
\end{IEEEkeywords}

%
\IEEEpeerreviewmaketitle

\section{Introduction}\label{SecIntro}
There have been sustained attempts at designing algorithms that are able to automatically transfer and reuse learned knowledge across datasets, problems, and domains. The main objective is to promote the \emph{generalization capacity} of intelligent systems in a way that performance efficiency and effectiveness is not only restricted to an individual (narrow) task, but can also be reproduced in other \emph{related} tasks by reusing \emph{computationally encoded knowledge priors} \cite{ong2019air}. This idea has particularly strong implications in the field of evolutionary computation. As evolutionary optimization algorithms are population-based, their sample complexity, i.e., the number of evaluations needed to find good solutions, can become prohibitive when searching from scratch. In this regard, \emph{transfer evolutionary optimization} (TrEO) is a promising approach that enhances computational tractability, reusing evolved skills drawn from previously tackled \emph{source} tasks to speed up the solving of related \emph{target} tasks at hand \cite{gupta2017insights, wong2021can,tan2021evolutionary}.


Reusing knowledge priors for sample efficient problem-solving constitutes what has broadly been referred to as \emph{transfer optimization} in the literature \cite{gupta2017insights}. There have been a growing number of successful examples of the transfer optimization paradigm in areas such as neuro-evolution \cite{gupta2018memetic}, multi-objective continuous optimization \cite{min2017multiproblem,feng2017autoencoding,zhang2019multi}, combinatorial optimization \cite{feng2014memetic,feng2015memes,ardeh2019novel,lim2019can}, dynamic optimization \cite{jiang2017transfer,liu2019neural,jiang2020individual}, training hyperheuristics  \cite{zhang2021surrogate,zhang2021multitask}, and learning classifier systems \cite{iqbal2012extracting,iqbal2013reusing,iqbal2017cross} with real-world applications from the composites manufacturing industry \cite{feng2017autoencoding}, engineering design \cite{min2017multiproblem,zhang2019multi,min2017knowledge}, last-mile logistics \cite{feng2020towards}, image feature learning \cite{bi2020learning}, automated machine learning \cite{min2020generalizing,joy2019flexible}, reinforcement learning \cite{karimpanal2020learning}, to name a few. 

In the class of \emph{sequential transfer optimization} (where a set of tasks are solved sequentially by utilizing information from those already solved), two ways of reusing priors are normally employed. One is the \emph{exact storage} and reuse of past solutions (either directly \cite{louis2004learning,cunningham1997case,kaedi2011biasing} or after passing through a mapping function \cite{feng2017autoencoding,didi2016multi}) for subsequent injection into the search space of the target problem. The other is \emph{model-based} transfer where reuse of priors has recently been realized by sampling from probabilistic models of elite candidate solutions drawn from previously solved optimization tasks \cite{friess2020improving,da2018curbing}. Irrespective of which approach is being applied, the challenge is to \emph{efficiently} curb the threat of \emph{negative transfer} when there are \emph{big} source task-instances and \emph{many} are not relevant to the target problem. (Note that an \emph{instance} represents a single optimization problem.) Measuring inter-task similarity, however, requires utilization of problem-specific data that may not be a priori known before the onset of the search, particularly in black-box evolutionary optimization settings which is the focus of this paper. This promoted \emph{online source-target similarity learning} as an effective strategy to mitigate the risks of negative transfer on the fly during the course of evolutionary optimization \cite{zhang2019multi,da2018curbing,shakeri2019coping}. In \cite{zhang2019multi}, Zhang \emph{et al.} introduced a multi-source selective transfer framework where inter-task similarity is captured by measuring the Wasserstein distance of the respective source and target probabilistic search models. The final source population is then formed by applying heuristic rules based on the max and variance of similarities across different sources, and, through a mapping function, injects the candidate solutions into an evolutionary algorithm. The work utilized Feng \emph{et al.}'s single layer denoising autoencoder \cite{feng2017autoencoding,feng2018evolutionary} as the mapping function to connect source and target populations in continuous search spaces. Their approach, however, requires similarity computations and comparisons in every generation of the evolution to effectively mandate the process of transfer. This incurs significant computational cost when the number of sources starts to grow; see complexity analysis in Section \ref{MdlOnTrfAlgo}.

The threat of negative transfer was also tackled by Da \emph{et al.} \cite{da2018curbing}. They developed an adaptive model-based transfer evolutionary algorithm (AMTEA) in which the measure of source-target similarity is defined as the extent of overlap between the optimized source and target search distributions captured via stacked density estimation of the target population. The AMTEA, however, requires a learning algorithm to be nested within the target evolutionary search to reveal latent source-target similarities during the course of the search. This impedes the approach from \emph{scaling} efficiently when large number of source probabilistic models are to be stacked with the target model. Shakeri \emph{et al.} \cite{shakeri2019coping} addressed the scaling burden of the AMTEA by incorporating a source selection mechanism, based on the theory of multi-armed bandits (MAB), to stack only a single source model to the target each time the similarity learning procedure is launched. However, a shortcoming of the resultant MAB-AMTEA is that identification of useful sources can become slow under sparsity of relevant source models. The uniformly random sampling of sources at initial stages of the MAB-AMTEA often prevents the algorithm from \emph{quickly converging} to the most relevant source. 

Given this background, the present paper aims at \emph{simultaneously} tackling two critical issues that have only been addressed in isolation by today's transfer evolutionary optimization frameworks, namely (1) lack of \emph{scalability} when faced with a growing number of source instances and (2) lack of \emph{online learning agility} against sparsity of relevant sources to the target task of interest. While applications of most existing algorithms are limited to the order of tens of source tasks, this paper takes a quantum leap forward to handle scenarios beyond 1000 source task-instances. The impact of such scalability is becoming increasingly apparent given large-scale cloud-based platforms of today, where online services catering to thousands of clients worldwide have emerged. For example, consider a cloud service that recommends \emph{packages} to \emph{clients}. For any new query raised by a client, the process of finding customized packages to recommend can be cast as the NP-hard 0/1 knapsack problem \cite{xie2010breaking} (a classical example in combinatorial optimization that we also study in Section \ref{KnpPrb}). Although NP-hard problems are known to be difficult to solve, especially when starting the search from scratch, the idea of TrEO makes uniquely possible the reuse of past problem solving experiences pertaining to the same or other similar clients. (In this setting, every experience is seen as a solved source task.) The ability to efficiently and effectively cope with thousands of experiences could then facilitate real-time generation of the best recommendation package, augmenting service level and throughput of cloud services as a whole.

To this end, we devise a novel \emph{scalable transfer evolutionary optimization} (sTrEO) framework comprising two co-evolving species for joint evolutions in the space of source priors and in the space of solutions to the target problem. The proposed co-evolutionary transfer mechanism enables the learned knowledge to be rapidly orchestrated by applying a \emph{two-membered evolution strategy}, i.e., (1+1)-(ES), with a mutation function that can adjust the level of transfer based on the relatedness of sources to the target. The key contribution of this work can thus be summarized as follows.
\begin{enumerate}
	\item We establish a scalable transfer evolutionary optimization framework with two salient features of scalability and online learning agility through co-evolution in the knowledge and target search spaces.
	\item We incorporate a novel mutation mechanism in the (1+1)-ES based source-target similarity learning algorithm that enables efficient and effective utilization of relevant source task-instances beyond two orders of magnitude larger in number than those encountered in existing works.
	\item We empirically verify the efficiency and effectiveness of the algorithm through an extensive and rigorous experimental study across a set of practically motivated discrete and continuous optimization examples.
\end{enumerate}

The rest of the paper is organized as follows. Section \ref{SecPri} contains preliminaries on probabilistic model-based transfer evolutionary optimization and conducts a complexity comparison for some of the state-of-the-art algorithms. In Section \ref{AMCTfrm}, we first present formal definitions for scalability and online learning agility before describing the sTrEO and its (1+1)-ES based source-target similarity learning module. Extensive experiments supplemented with rigorous analyses are conducted in Section \ref{SecExpSty} for assessing the performance of sTrEO across three case studies ranging from discrete to continuous search spaces. Finally, Section \ref{secconc} concludes the paper and highlights potential future research directions.

\section{Preliminaries}\label{SecPri}
\subsection{Basics of Probabilistic Model-Based Transfer}\label{MdlTrf}
The probabilistic model-based expression of a typical optimization problem $\mathcal{T}$ with a maximization function $f(\pmb{x})$ (we adopt $\minus f(\pmb{x})$ if the underlying optimization problem is one of minimization) over a search space $\mathcal{X}$ can be mathematically formulated as \cite{wierstra2014natural},
\begin{equation}\label{eq1}
\max_{p(\pmb{x})}\int_{\mathcal{X}}f(\pmb{x}).p(\pmb{x}).d\pmb{x},
\end{equation}
where, $p(\pmb{x})$ is the probabilistic search distribution model of candidate solutions in $\mathcal{X}$. Given a global optimum fitness value $f^{*}$, an optimized model $p^{*}(\pmb{x})$ is one of low entropy that generates high-quality (near optimal) solutions satisfying,
\begin{equation}\label{eq2}
\int_{\mathcal{X}}f(\pmb{x}).p^{*}(\pmb{x}).dx\geq f^{*}-\varepsilon,
\end{equation}
where, $\varepsilon$ represents a small positive convergence threshold. Note that if we let $\varepsilon$ be $0$, $p^{*}(\pmb{x})$ could then become a degenerate distribution that collapses to just a single point; this is not what we want.

In the sequential transfer optimization setting, when the objective is to optimize \emph{target} task $\mathcal{T}_{T}$, we assume that we have solved $T-1$ \emph{source} tasks $\mathcal{T}_{1}, \dots, \mathcal{T}_{T-1}$ giving low entropy knowledge priors $p^{*}_{1}(\pmb{x}), \dots, p^{*}_{T-1}(\pmb{x})$ over a \emph{unified} search space $\mathcal{X}_{u}$ that encodes solutions to all $T$ tasks. A complete description of unified search spaces can be found in \cite{gupta2018memetic}; details are left out herein for the sake of brevity. Under this setting, the transfer evolutionary optimization of the target problem from the viewpoint of probabilistic model-based search is described as \cite{gupta2018memetic}, 
{\small \begin{equation}\label{eq3}
\begin{split}
\max_{w_{1},\dots,w_{T-1},w_{T},p_{T}(\pmb{x})}\int_{\mathcal{X}_{u}}f_{T}(\pmb{x}).\Big[\sum_{s=1}^{T-1}w_{s}.p_{s}^{*}(\pmb{x})+w_{T}.p_{T}(\pmb{x})\Big].d\pmb{x} \\
    \textrm{such that} \sum_{i=1}^{T}w_{i}=1 \textrm{ and } w_{i}\geq 0, \forall i\in \{1,\dots,T-1,T\}.
\end{split}
\end{equation}}

In this formulation, $w_{1}, \dots, w_{T-1}, w_{T}$ are referred to as \emph{transfer coefficients} in the probabilistic mixture model $\big[\sum_{s=1}^{T-1}w_{s}.p_{s}^{*}(\pmb{x})+w_{T}.p_{T}(\pmb{x})\big]$, and determine the extent of knowledge transfer from each source to the target. The mixture model is trained to capture the true underlying distribution of the target population, from which candidate solutions are sampled to adaptively bias the target search. A proper adjustment of the coefficients is challenging nonetheless as there is no prior knowledge on the optimal search distribution $p_{T}^{*}(\pmb{x})$ of the target problem to help identify related and unrelated sources offline. The consequences of improper assignments are twofold. On one hand, designating an unreasonably high transfer coefficient to an unrelated source model will result in injecting poor solutions into the search space of the target that may significantly impede the target search performance. On the other hand, the inability to assign an adequately high value to a relevant source model suppresses the transfer of useful solutions to improve optimization efficiency in the target task. The challenge is more severe in black box optimization. Due to the absence of an algebraic model of the system to be optimized in such settings, we can only rely on online data generated during the course of target optimization to discover inter-task similarities. It is thus crucial for a transfer optimization algorithm to be able to accurately orchestrate the mixture transfer coefficients of (\ref{eq3}) based on the data generated during the target optimization search. 

\subsection{State-of-the-Art Transfer EAs}\label{MdlOnTrfAlgo}
It can be inferred from (\ref{eq3}) that the optimization of the target probabilistic model as well as adjustment of the mixture transfer coefficients are two intertwined components of the online model-based transfer optimization. The adaptive model-based transfer EA framework (AMTEA), as proposed in \cite{da2018curbing}, captures the latter via stacked density estimation of the target population \cite{smyth1998stacked}. At any generation of the target EA where the source-target similarity learning procedure is launched, a mixture model comprising a linear composition of $T$ available probabilistic models is learned from $n$ solutions (of dimensionality $d$) of the current generation by finding the optimal configuration of the transfer coefficients to each model. The learning process is carried out by applying the classical \emph{expectation-maximization} (EM) algorithm \cite{moon1996expectation}, which, in the context of stacked density estimation, entails the construction of an $n\times T$ probability matrix to capture the likelihood of each solution to all the $T$ probabilistic models \cite{da2018curbing}. Overall, this requires $\mathcal{O}(dnT)$ computational steps for a variable-wise factored distribution model, in addition to the complexity of executing the EM algorithm itself. The cost for learning \emph{optimal} transfer coefficient values is that AMTEA does not scale efficiently when the number of source task-instances grows to hundreds and beyond.

The lack of scalability of the AMTEA was tackled in \cite{shakeri2019coping} via a source selection algorithm that runs in $\mathcal{O}(T)$ to choose only one source model at a time for stacked density estimation. This reduces the computational steps required for constructing the probability matrix to $\mathcal{O}(dn)$, which is a significant speedup when there are thousands of source models available. The selection strategy is formulated based on a popular variant of the multi-armed bandit problem, known as the adversarial MAB \cite{auer1995gambling} which can be solved to optimality using the \emph{Exponential-weight algorithm for Exploration and Exploitation} (EXP3) \cite{auer2002nonstochastic}. In this setting, each source corresponds to an arm of a bandit whose reward, when chosen (pulled), is proportional to its level of similarity to the target. A source model which can generate fitter individuals, as evaluated by the target objective function, receives a higher reward measure and is more likely to be chosen in future generations. However, as was noted in Section \ref{SecIntro}, the uniformly random selection of just a single source at initial stages of the MAB often hinders the algorithm from quickly converging to the most relevant source, especially when related sources are sparse. An extensive series of experiments and analyses to evaluate the above algorithms and claims will be carried out in Section \ref{SecExpSty}. We denote the MAB-enabled AMTEA developed in \cite{shakeri2019coping} as MAB-AMTEA for the remainder of the paper.

There is another class of sequential transfer methods that builds a single layer mapping function to inject solutions of past search experiences into the current target population while the optimization search progresses \cite{feng2017autoencoding}. The mapping function is built each time knowledge transfer occurs. Given two solution populations (one pertaining to a previously solved source task and the other to the target problem) both of which consist of $n$ solutions of $d$ dimensions, a basic implementation requires $\mathcal{O}(nd^{2}+d^{3})$ computational steps to construct the mapping function. (Detailed computational analyses are not presented here for the sake of brevity.) The above process, however, is not scalable when the number of sources is large (i.e., $T\gg1$), as it needs to be repeated per source to build the corresponding mapping function; this makes the complexity scale as $\mathcal{O}(T(nd^{2}+d^{3}))$. In addition, memory consumption is another resource bottleneck as it is required to archive the solutions data of all source instances for all generations, which can quickly become intractable. Although the challenge of multiple sources was addressed in \cite{zhang2019multi} by proposing a source selection mechanism to form a single population to construct the mapping function, the memory consumption bottleneck remains. In contrast, in the probabilistic model-based sequential transfer methods, we only need to store the learned search distribution models, comprising only few distribution parameters. Finally, note that the existing single layer autoencoder method only applies to continuous search spaces.

\section{Scalable Transfer Evolutionary Optimization}\label{AMCTfrm}
The  main  motivation  behind  proposing the sTrEO is to \emph{simultaneously} achieve  scalability  and  online  learning  agility. For any transfer evolutionary algorithm, these two measures can be defined as follows:

  \textit{\textbf{Definition 1}} \textit{(Scalability in sTrEO)}: This represents internal algorithmic efficiency. It is measured by the time complexity, with respect to the number of sources, of each iteration of the source-target similarity learning procedure. 
  
  \textit{\textbf{Definition 2}} \textit{(Online Learning Agility in sTrEO)}: This represents algorithmic effectiveness or convergence rate. It is measured by the number of iterations taken by the source-target similarity learning procedure to correctly identify those sources that are most related to the target task. 

Based on these two definitions, we find that the existing AMTEA tends to compromise on scalability in favor of online learning agility, whereas the MAB-AMTEA tends to compromise online learning agility in favor of scalability.

\begin{figure}
    \centering
    \includegraphics[width=0.49\textwidth]{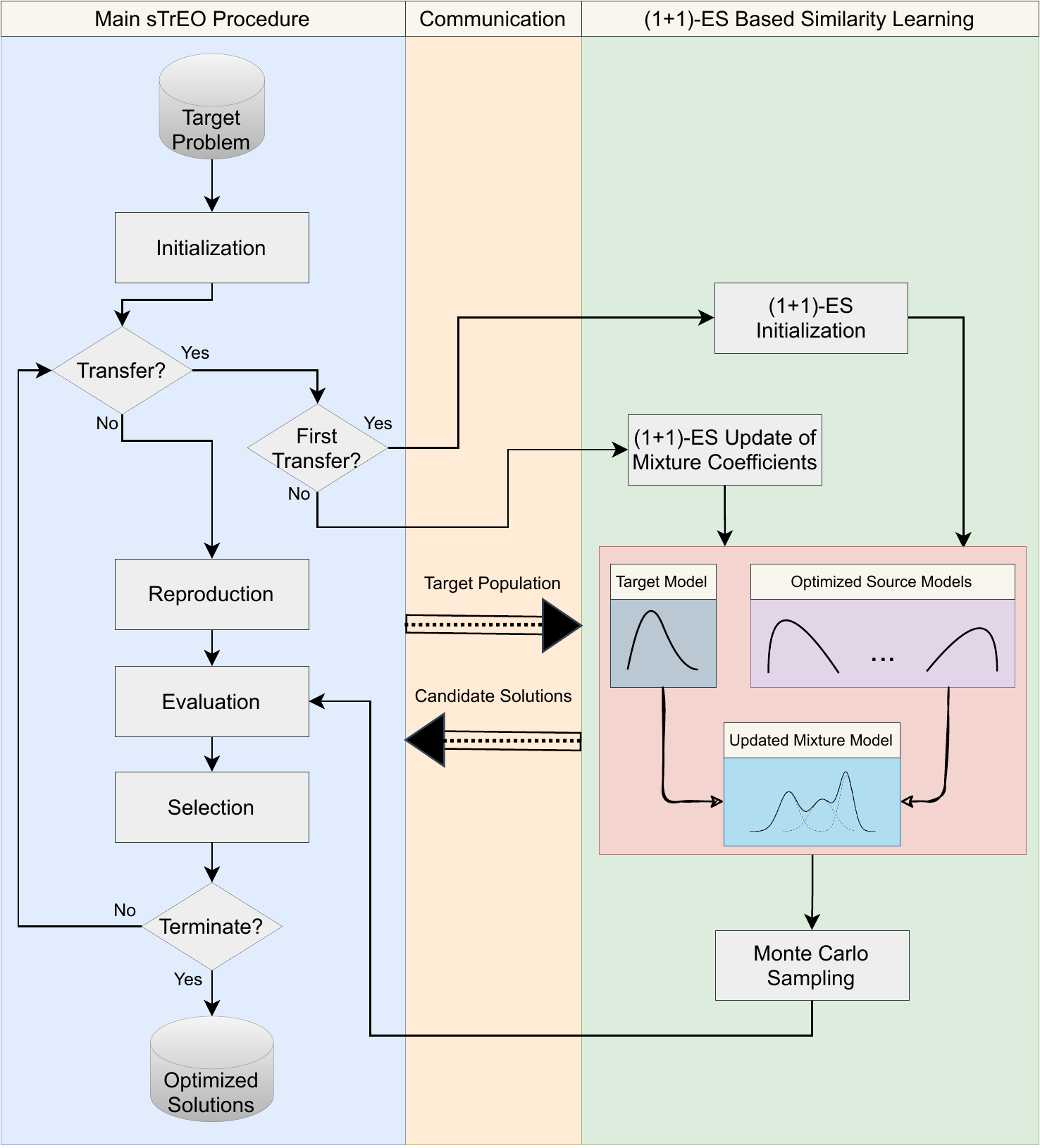}
    \caption{Flowchart of the sTrEO. Unlike the AMTEA algorithm \cite{da2018curbing} which incorporates computationally intensive stacked density estimation, the sTrEO uses a fast (1+1)-ES algorithm to learn the transfer coefficients.}
    \label{FigsTrEO}
\end{figure}

The sTrEO comprises two co-evolving species for joint evolutions in the space of source knowledge and in the search space of solutions to the target problem. The proposed co-evolutionary transfer mechanism enables efficient and effective utilization of the learned knowledge beyond 1000 source task-instances by applying a two-membered evolution strategy, i.e., (1+1)-(ES), with a novel mutation function capable of adjusting the level of transfer in accordance with the relatedness of sources to the target. The advantage of the (1+1)-ES compared to other population-based evolutionary algorithms is that evolution is carried out with only one individual, eliminating the complexity of evaluating and evolving an entire population. This property is in line with the paper's objective to ensure scalability in sTrEO. In this work, the species to evolve the solutions to the target problem is just a canonical EA (although any other preferred algorithm can be used). Fig. \ref{FigsTrEO} depicts a flowchart illustrating the steps of knowledge transfer in the proposed framework. The next two sections describe the components of sTrEO in detail.

\subsection{Source Knowledge Evolution in sTrEO}\label{SecMemeEvol}
The core component of sTrEO is the source-target similarity learning module which is based on the (1+1)-ES. In the (1+1)-ES, one solution is evolved per generation by being mutated. The fitness of the two solutions (i.e., the parent and its offspring) is compared and the best of the two is chosen as the individual for the next generation. In our setting, the \emph{chromosome} is encoded as the transfer coefficients in the mixture model and the \emph{fitness} is defined as the mean fitness of candidate solutions generated from the corresponding mixture model and evaluated by the objective function of the target problem. The algorithm evolves for one generation every time it is launched by the target EA, the frequency of which is determined by a \emph{transfer interval} (denoted by $\Delta$ generations in our setting -- see Procedure \ref{TEvolProcess}).

In the standard (1+1)-ES, the parent is mutated according to a mutation radius, a random variable following a normal distribution. In this paper, however, we propose a different mutation mechanism according to the characteristics of the transfer optimization paradigm as formulated in (\ref{eq3}). Given that $\int_{\mathcal{X}}f(\pmb{x}).p(\pmb{x}).d\pmb{x}$ indicates the expected objective value of a search distribution model $p(\pmb{x})$ against an objective function $f(\pmb{x})$ over a search space $\mathcal{X}$, we rephrase the formulation in (\ref{eq3}) as follows, 

\begin{equation}\label{eq3b}
\begin{split}
   \max_{w_{1},\dots,w_{T-1},w_{T}, p_{T}(\pmb{x})} & w_{1}.E_{\pmb{x}\sim p_{1}^{*}(\pmb{x})}\big[f_{T}(\pmb{x})\big]+\dots\\
  &  +w_{T-1}.E_{\pmb{x}\sim p_{T-1}^{*}(\pmb{x})}\big[f_{T}(\pmb{x})\big]\\
  &  +w_{T}.E_{\pmb{x}\sim p_{T}(\pmb{x})}\big[f_{T}(\pmb{x})\big],
\end{split}
\end{equation}
where $E_{\pmb{x}\sim p_{s}^{*}(\pmb{x})}[f_{T}(\pmb{x})]$ is the expected fitness of an optimized source probabilistic model $p_{s}^{*}(\pmb{x})$ relative to the target objective function $f_{T}(\pmb{x})$. Also note that $\sum_{i=1}^{T}w_{i}=1$ and $w_{i}\geq 0$, $\forall i\in \{1,\dots,T\}$.

As the overall objective in (\ref{eq3b}) is formulated as that of maximization, a higher expected value for each model should be reinforced with a higher transfer coefficient. This forms the basis for our novel mutation method as a means of guiding the evolution toward maximizing positive transfer by giving higher weights to related sources as well as suppressing negative transfer by neutralizing weights to unrelated ones. We implement this by defining a mutation vector of size $T$ such that the first $T-1$ elements correspond to the source models and the last refers to the target. Each entry in the mutation vector estimates the expected fitness of its corresponding probabilistic model. This is done by recording the up-to-date average fitness of candidate solutions that have been generated from the model and evaluated by the target objective function during the course of source-target similarity learning. Note that as the knowledge extraction progresses, the samples generated from the models accumulate, leading to more accurate estimations of their expected fitness. After mutation is carried out per generation, a new mixture model is formed from which a sample population is generated for knowledge transfer. For clarity, we present the details of the (1+1)-ES based source-target similarity learning module in pseudocode in Procedure \ref{CMTkExtracProcess} for the case of a maximization problem. 

\begin{proc}
\SetAlgoVlined
\KwIn{Optimized source models $p^{*}_{1}(\pmb{x}),\dots,p^{*}_{T-1}(\pmb{x})$, temperature $\lambda$, learning rate $\eta$, neutralization threshold $\varepsilon$, target task $\mathcal{T}_{T}$, transfer counter $t$, target population $P_{t}$, mean target fitness $\bar{f}_{T,t}$;} 
\eIf{$t==0$}{
Initialize transfer coefficients $w_{t}[1:T]$ with $1/T$\;
Initialize mutation vector $\pi[1:T]$ with zeroes\;
Build target model $p_{T,t}(\pmb{x})$ using $P_{t}$\;
Form mixture model $M_{t}$ from $p^{*}_{1}(\pmb{x}),\dots,p^{*}_{T-1}(\pmb{x})$, $p_{T,t}(\pmb{x})$ and $w_{t}[1:T]$\;
Generate candidate transfer solutions $X_{t}$ from $M_{t}$ using the Monte Carlo sampling method\;
Evaluate solutions in $X_{t}$ using target objective function $f_{T}(\pmb{x})$ and record the mean fitness as $\bar{f}_{t}$\;
}{
\For{$s=1$ \KwTo $T-1$}{
$\pi[s]\gets$ average fitness of candidate solutions in $\bigcup_{i=0}^{t-1}X_{i}$ that have been sampled from $p^{*}_{s}(\pmb{x})$\;
 }
$\pi[T]\gets \bar{f}_{T,t}$\;
\For{$i=1$ \KwTo $T$}{
$\pi^{\prime}[i]\gets \pi[i]/\textrm{max}(\pi[1:T])$\;
 }
\For{$i=1$ \KwTo $T$}{
$\pi^{\prime\prime}[i]\gets e^{\frac{\pi^{\prime}[i]}{\lambda}}/\sum_{j=1}^{T}e^{\frac{\pi^{\prime}[j]}{\lambda}}$\;
$w^{\prime}[i]\gets (1-\eta)\times w_{t-1}[i]+\eta\times\pi^{\prime\prime}[i]$\;
 }
\For{$i\leftarrow 1$ \KwTo $T$}{
\lIf{$w^{\prime}[i]\leq \varepsilon$}{
Neutralize $w^{\prime}[i]$ by $0$
  }
 }
\For{$i\leftarrow 1$ \KwTo $T$}{
$w^{\prime\prime}[i]\gets w^{\prime}[i]/\sum_{j=1}^{T}w^{\prime}[j]$\;
 }
Build target model $p_{T,t}(\pmb{x})$ using $P_{t}$\;
Form mixture model $M_{t}$ from $p^{*}_{1}(\pmb{x}),\dots,p^{*}_{T-1}(\pmb{x})$, $p_{T,t}(\pmb{x})$ and $w^{\prime\prime}[1:T]$\;
Generate candidate transfer solutions $X_{t}$ from $M_{t}$ using the Monte Carlo sampling method\;
Evaluate solutions in $X_{t}$ using target objective function $f_{T}(\pmb{x})$ and record the mean fitness as $\bar{f}^{\prime}$\;
\eIf{$\bar{f}^{\prime}\geq \bar{f}_{t-1}$}{
$w_{t}[1:T]\gets w^{\prime\prime}[1:T]$\;
$\bar{f}_{t}\gets \bar{f}^{\prime}$;
}{
$w_{t}[1:T]\gets w_{t-1}[1:T]$\;
$\bar{f}_{t}\gets \bar{f}_{t-1}$\;
 }
}
\Return{$X_{t}$} 
\caption{(1+1)-ES Based Similarity Learning}\label{CMTkExtracProcess}
\end{proc}

As already stated, Procedure \ref{CMTkExtracProcess} is called by the target EA each time knowledge transfer occurs. This is traced by a \emph{transfer counter} $(t)$ which is input to the procedure and indicates the current generation of the learning process. Accordingly, $t=0$ represents the initialization phase (lines 2--7) in which the initial chromosome is formed by assigning identical values (here, $1/T$) to the transfer coefficients. This implies that each source model, as well as the initial target model, has an equal probability to transfer candidate solutions to the target optimization algorithm. These individuals are evaluated by the target objective function and the mean fitness is recorded as the fitness of the initial chromosome. 

For the case of $t>0$, the similarity learning process proceeds with updating the mutation vector (lines 9--11). As already discussed, each entry in the mutation vector estimates the expected fitness of its corresponding model relative to the target problem and is measured by taking the average fitness of candidate solutions that have been sampled from that model. The average fitness to a source model is updated each time a new solution is added. This is done incrementally in a constant time without the need to keep the previous solutions in memory. Assume that the mutation entry $\pi[s]$, corresponding to the source model $p^{*}_{s}(\pmb{x})$, holds the average fitness over the first $i-1$ solutions; the updated average after adding the $i$th candidate solution $\pmb{x}_{i}$ sampled from $p^{*}_{s}(\pmb{x})$ is calculated as follows,
\begin{equation}\label{eq6}
\pi[s]\gets \pi[s]+\frac{f_{T}(\pmb{x}_{i})-\pi[s]}{i}.
\end{equation}
Note that unlike the source probabilistic models which remain fixed during the target optimization, the target model is non-stationary and is rebuilt each time the source-target similarity learning procedure is launched. This hinders the incremental update of the average fitness (as done for the source models) from being applied to the target. However, since the solutions to build the target model are evaluated in the main sTrEO procedure before being passed to the (1+1)-ES, it is possible to compute their mean fitness $\bar{f}_{T,t}$ beforehand and pass it along with the target population $P_{t}$ to Procedure \ref{CMTkExtracProcess}. 

Thereafter, a two-stage transformation is carried out on the mutation vector to help online source-target similarity to be captured more efficiently and effectively. This is realized by applying the \emph{maximum absolute scaling} and \emph{softmax} functions, respectively (lines 13 and 15). Here, the softmax function is used with \emph{temperature} $(\lambda)$, a hyper-parameter to affect the final generated probabilities. Lowering the temperature imposes more contrast across sources based on their relevance to the target problem. Eventually, mutation is carried out in line 16 to generate the offspring from the parent chromosome and mutation vector. A \emph{learning rate} ($\eta$) in the interval $[0,1]$ is used to determine the step-size of transfer coefficient updates in the offspring.

The updated transfer coefficients in the offspring form a new mixture model that should be used to generate new candidate solutions. This, however, is deferred until a neutralization function is carried out on each element of the offspring (line 18). If the value is below a positive {\it neutralization threshold} ($\varepsilon$), then the entry (which entails the transfer coefficient to a source model) is neutralized to zero, meaning that the corresponding source is identified unrelated and hence, should not be sampled for the remaining generations. The rationale behind such adjustment lies in the Monte Carlo sampling method being used, as it ensures that at least one sample is generated from every source model whose corresponding transfer coefficient is positive (no matter how close the value is to zero). Consequently, in situations where the number of sources is significantly greater than the population size, there could be no chance for those models with a slightly larger transfer coefficient to have their sample(s) chosen as candidate solutions for transfer. In this regard, neutralizing the sources with extremely low transfer coefficients is deemed an effective strategy to remedy this undesirable property. 

The last part of the code, i.e., lines 25--30, deals with evolution in which the parent $w_{t-1}[1:T]$ at iteration $t-1$ is replaced with the offspring $w^{\prime\prime}[1:T]$ if its fitness $\bar{f}_{t-1}$ is not better than the offspring's fitness $\bar{f}^{\prime}$. Note that all solutions in $X_{t}$ (line 23) are returned to the main sTrEO procedure (line 31) to guide the target search process. 


\subsection{Target Solutions Evolution in sTrEO}\label{SecTargetEvol}
Our proposed (1+1)-ES can be incorporated as a nested module within any canonical or state-of-the-art EA to optimize the target task. Procedure \ref{TEvolProcess} presents such integration for a canonical genetic algorithm (CGA) in pseudocode. In this setting, \emph{transfer interval} ($\Delta$) is the frequency of calling the source-target similarity learning module in number of generations. Upon the call, the (1+1)-ES advances for one generation. 

According to the procedure, after the initial population is configured and evaluated (lines 1 and 2), the iterative routine to optimize the target solutions follows two different branches of evolution. Whenever knowledge transfer occurs, the offspring is formed from the candidate solutions returned by Procedure \ref{CMTkExtracProcess} (lines 6--9). Otherwise, it is generated by applying the common reproduction operators (line 11). After being evaluated, the offspring is combined with the current population to form the next generation by using a selection mechanism (lines 12 and 13). Note that before Procedure \ref{CMTkExtracProcess} is launched in line 8, the mean fitness of solutions in the current population is computed in line 7 and passed as the input parameter ($\bar{f}_{T,t}$) along with the transfer counter $t$ to Procedure \ref{CMTkExtracProcess}. The above steps continue until some stopping criteria are eventually met. 
\begin{proc}
\SetAlgoVlined
\KwIn{Target task $\mathcal{T}_{T}$, transfer interval $\Delta$;}
\KwOut{Optimized target probabilistic model;}
Configure initial population $P_{0}$ at random\;
Evaluate solutions in $P_{0}$\;
$i\gets 0$ \tcc*{generation counter}
$t\gets 0$ \tcc*{transfer counter}
\While{stopping criteria not satisfied}{
\eIf{$mod(i,\Delta)==0$ {\bf and} $i>1$}{
$\bar{f}_{T,t}\gets$ mean fitness of solutions in $P_{i}$\;
Call Procedure \ref{CMTkExtracProcess} to form offspring $O$\;
$t\gets t+1$\;
}{
Generate offspring $O$ by reproduction of $P_{i}$\;
}
Evaluate solutions in $O$\;
Select next population $P_{i+1}$ from $P_{i}\cup O$\;
$i\gets i+1$;
}
\caption{Pseudocode of sTrEO for a CGA}\label{TEvolProcess}
\end{proc}

\subsection{Complexity Comparison of sTrEO}\label{CMTEAComp}
The learning process of the transfer coefficients in the sTrEO is carried out by the mutation mechanism of the (1+1)-ES. To this end, the $T$ entries in the mutation vector are updated each time a new candidate solution is added from the mixture model. According to (\ref{eq6}), this is done in a constant time per added solution, which, overall, requires $\mathcal{O}(n)$ computational steps for $n$ solutions. (Note that computing the mean fitness of the target population, to update the last entry of the mutation vector, also requires $\mathcal{O}(n)$ steps for a population size of $n$.) Thereafter, the two-stage transformation as well as the generation of the offspring (that encodes the updated transfer coefficients) are carried out in $\mathcal{O}(T)$. Thus, the total complexity of the mutation mechanism becomes $\mathcal{O}(n+T)$. As is clear, the (1+1)-ES, and hence the sTrEO as a whole, scales much more efficiently than the existing AMTEA with the learning complexity $\mathcal{O}(dnT)$.


\section{Experimental Study}\label{SecExpSty}
We conducted extensive experiments to assess the performance of sTrEO with regard to \emph{scalability} and \emph{online learning agility} across a set of practically motivated discrete and continuous optimization examples. Following our complexity analysis in Section \ref{MdlOnTrfAlgo}, we should note that the single layer autoencoder method proposed in \cite{feng2017autoencoding} and \cite{zhang2019multi} requires archiving the solutions data of all source instances for all generations, to construct the mapping function whenever the source-target similarity learning procedure is launched. This entails a significantly large and growing memory cost for evaluation scenarios with 1000 sources and beyond. As a more practically feasible alternative, we compare sTrEO with state-of-the-art probabilistic model-based transfer methods that only need to store the learned search distribution models, which comprises only few distribution parameters and hence consumes low memory. Our comparative evaluations are accordingly conducted against AMTEA \cite{da2018curbing} and the MAB-enabled AMTEA (i.e., MAB-AMTEA) \cite{shakeri2019coping} methods, along with a number of modern solvers without transfer capability. This includes a canonical GA (CGA) enhanced with a local solution repair heuristic for discrete optimization and two powerful variants of the recently proposed \emph{natural evolution strategies} (NESs) \cite{wierstra2014natural}, namely sNES and xNES, for continuous examples.

\subsection{Experimental Setup}\label{SecExpSU}
All the compared algorithms were implemented in Python 3.7\footnote{The source code is available at \href{https://github.com/erfanMhi/Transfer-Optimization}{https://github.com/erfanMhi/Transfer-Optimization}.} and run on a 64-bit Intel(R) Core$^{\textrm{TM}}$ i7-7700HQ with 4 cores and 8 logical processors. The OS was Ubuntu 20.04 with 16GB memory. Throughout all experiments, we used an \emph{elitist} selection strategy in the three transfer evolutionary algorithms, i.e., AMTEA, MAB-AMTEA and sTrEO, as well as the CGA. The (unified) search space representation together with the choice of probabilistic models were determined according to the characteristics of the underlying problem. For discrete optimization, the following general settings were adopted in our compared algorithms: 
\begin{enumerate}
	\item Representation: binary coded.
	\item Population size: $50$.
	\item Maximum function evaluations: $5000$.
	\item Evolutionary operators: 
	\begin{enumerate}
	    \item uniform crossover with probability $\rho_{c}=1$,
	    \item bit-flip mutation with probability $\rho_{m}=1/d$, where $d$ is the chromosome length.
    \end{enumerate}
	\item Probabilistic model: univariate marginal frequency (factored Bernoulli distribution) \cite{muhlenbein1997equation}.
\end{enumerate}

On the other hand, for the chosen set of continuous optimization examples, the following configuration was used:
\begin{enumerate}
	\item Representation: real-valued coded. 
	\item Population size: $50$.
	\item Maximum function evaluations: $5000$.
	\item Evolutionary operators for AMTEA, MAB-AMTEA and sTrEO: 
	\begin{enumerate}
	    \item simulated binary crossover \cite{deb1995simulated} with $\rho_{c}=1$ and distribution index $\eta_{c}=10$,
	    \item polynomial mutation \cite{deb2014analysing} with $\rho_{m}=1/d$ ($d$ is the chromosome length) and distribution index $\eta_{m}=10$.
    \end{enumerate}
	\item Probabilistic model: multivariate Gaussian distribution \cite{do2008multivariate}. 
\end{enumerate}

For fairness of comparison, we considered identical population size and maximum function evaluations for both sNES and xNES algorithms. The other parameters of the two solvers were configured with default settings from \cite{pybrain2010jmlr}. Note that the above configuration was adopted following the experimental study carried out in \cite{da2018curbing} for the assessment of the AMTEA approach. With regard to assessing the MAB-AMTEA, there is an egalitarianism factor $\gamma$ in the EXP3 that was set to $0.1$ according to \cite{shakeri2019coping}. 

As can be recalled from Section \ref{AMCTfrm}, we introduced four hyper-parameters to control the sTrEO as follows:
\begin{itemize}
  \item {\bf Temperature $(\lambda)$} affects the final probabilities generated by the softmax function in the mutation vector in the (1+1)-ES algorithm, contrasting source models based on their relevance to the target problem. 
  \item {\bf Learning rate $(\eta)$} determines the step-size of transfer coefficient updates in the (1+1)-ES algorithm. 
  \item {\bf Neutralization threshold $(\varepsilon)$} is the minimum value of the transfer coefficient for any source task to be considered for knowledge extraction in the (1+1)-ES algorithm.  
  \item {\bf Transfer interval $(\Delta)$} is the frequency of launching the nested (1+1)-ES algorithm in the main target procedure.
\end{itemize}

Sensitivity analyses were conducted to identify an appropriate yet general setting for each hyper-parameter. The relevant experimental study is presented in Section \ref{SenAnal} in the supplemental material. The resultant configuration is shown in \tableautorefname \ \ref{tabESSet}. Note that the neutralization threshold ($\varepsilon$) is a function of $T$ which is the total number of probabilistic models including the target model. Lastly, we should note that the setting for the transfer interval also applies to both AMTEA and MAB-AMTEA. 

\begin{table}
  \begin{center}
    \caption{Hyper-parameter setting for the (1+1)-ES.}\label{tabESSet}
    \begin{tabular}{p{4cm}p{4cm}}
    \hline
    Hyper-parameter & Setting \\ \hline
    Temperature $(\lambda)$ & $0.01$ \\ 
    Learning rate $(\eta)$ & $0.9$ \\ 
    Neutralization threshold $(\varepsilon)$ & $\frac{10^{-2}}{T}$ \\ 
    Transfer interval $(\Delta)$ & $2$ \\ \hline
    \end{tabular}
  \end{center}
\end{table}

\subsection{The Discrete 0/1 Knapsack Problem}\label{KnpPrb}
The knapsack problem (KP) is a well-known NP-hard problem in combinatorial (discrete) optimization and has been studied extensively by the operations research community. There are many variations of the problem with applications in supply chain and logistics optimization, recommender systems, wireless communications and investment decision making,  among many others. The basic version of the problem is the 0/1 KP in which there are $d$ number of items (representing the problem dimensionality), each with a value $v_{i}$ and a weight $w_{i}$, and a knapsack of a finite capacity $C$. The objective is to find a subset of items such that they can be placed into the knapsack while their total value is maximized. We can formulate the problem mathematically as follows,
\begin{align}
 \begin{gathered}
     \max \sum_{i=1}^{d}v_{i}x_{i}\\
     \textrm{s.t. } \sum_{i=1}^{d}w_{i}x_{i}\leq C, \textrm{ and } x_{i}\in \{0,1\},
 \end{gathered}   
\end{align}\label{KPeq}where, $x_{i}$ is a binary decision variable and equals $1$ if the $i$th item is selected and $0$, otherwise. It should be noted that solving KP instances using EAs may cause the capacity constraint to be violated for some individuals during the course of evolution. On such occasions, a repair mechanism is invoked to satisfy the constraint in accordance with Dantzig’s greedy approximation algorithm \cite{michalewicz1994genetic}.

We generate synthetic KP instances with same characteristics as \cite{da2018curbing} and \cite{shakeri2019coping}. Depending on the relationships between $w_{i}$ and $v_{i}$, the instances are randomly generated following three categories: (1) \emph{uncorrelated} (un), where $w_{i}=$ uniformly random real number $[1,10]$ and $v_{i}=$ uniformly random real number $[1,10]$; (2) \emph{weakly correlated} (wc), where $w_{i}=$ uniformly random real number $[1,10]$ and $v_{i}=w_{i}+$ uniformly random real number $[-5,5]$ (if, for some $i$, $v_{i}\leq0$, the value is ignored and sampling is repeated until $v_{i}>0$); and (3) \emph{strongly correlated} (sc), where $w_{i}=$ uniformly random real number $[1,10]$ and $v_{i}=w_{i}+5$. The second categorization defines two types of knapsacks: (1) \emph{restrictive capacity} (rc), where $C=20$; and (2) \emph{average capacity} (ac), where $C=0.5\sum_{i=1}^{d}w_{i}$. 

The knapsack problem types introduced above are representatives of \emph{composite recommendations} \cite{xie2010breaking} where each recommendation comprises a set of items with a value (rating) and a cost. Any recommended set of items is associated with a maximum total cost (budget) specified by the user where the constraint on the maximum number of items recommended can model the restrictive knapsack. In this setting, every query raised by any user is deemed to be an optimization problem instance. We define a $d=1000$-D KP configuration in which the target is ``KP\_uc\_ac'' and the source instance types are ``KP\_uc\_rc'', ``KP\_wc\_rc'', ``KP\_sc\_rc'' and ``KP\_sc\_ac''. Here, the related source type is ``KP\_sc\_ac'' implying that a relatively large number of items need to be selected as also must be done for the target.

\begin{figure}
  \begin{subfigure}[b]{0.49\columnwidth}
  \centering
    \includegraphics[width=\textwidth]{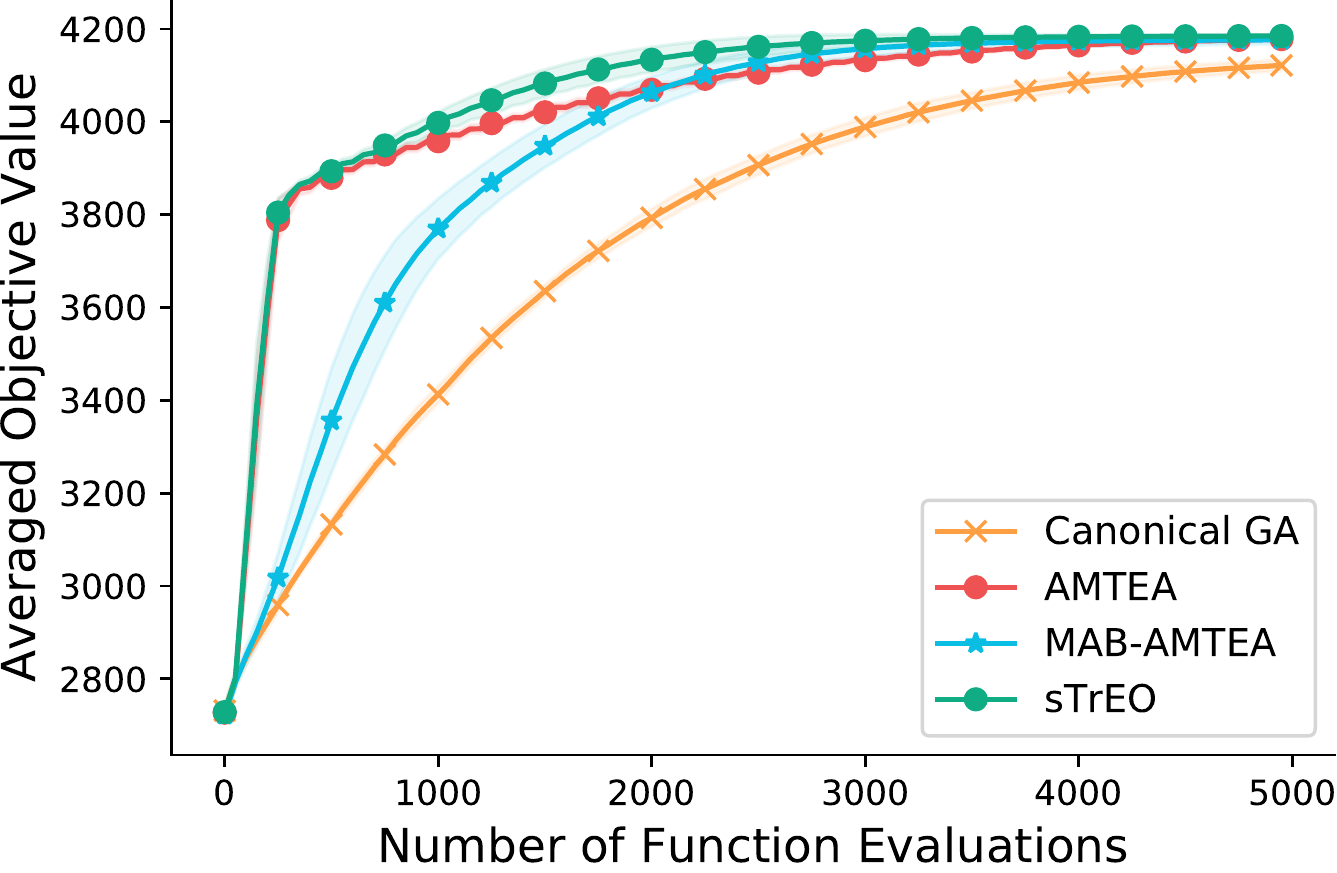}
    \caption{\scriptsize Convergence trends (250 related)}\label{FigKP1000-250G}
 \end{subfigure}
 \begin{subfigure}[b]{0.49\columnwidth}
 \centering
   \includegraphics[width=\textwidth]{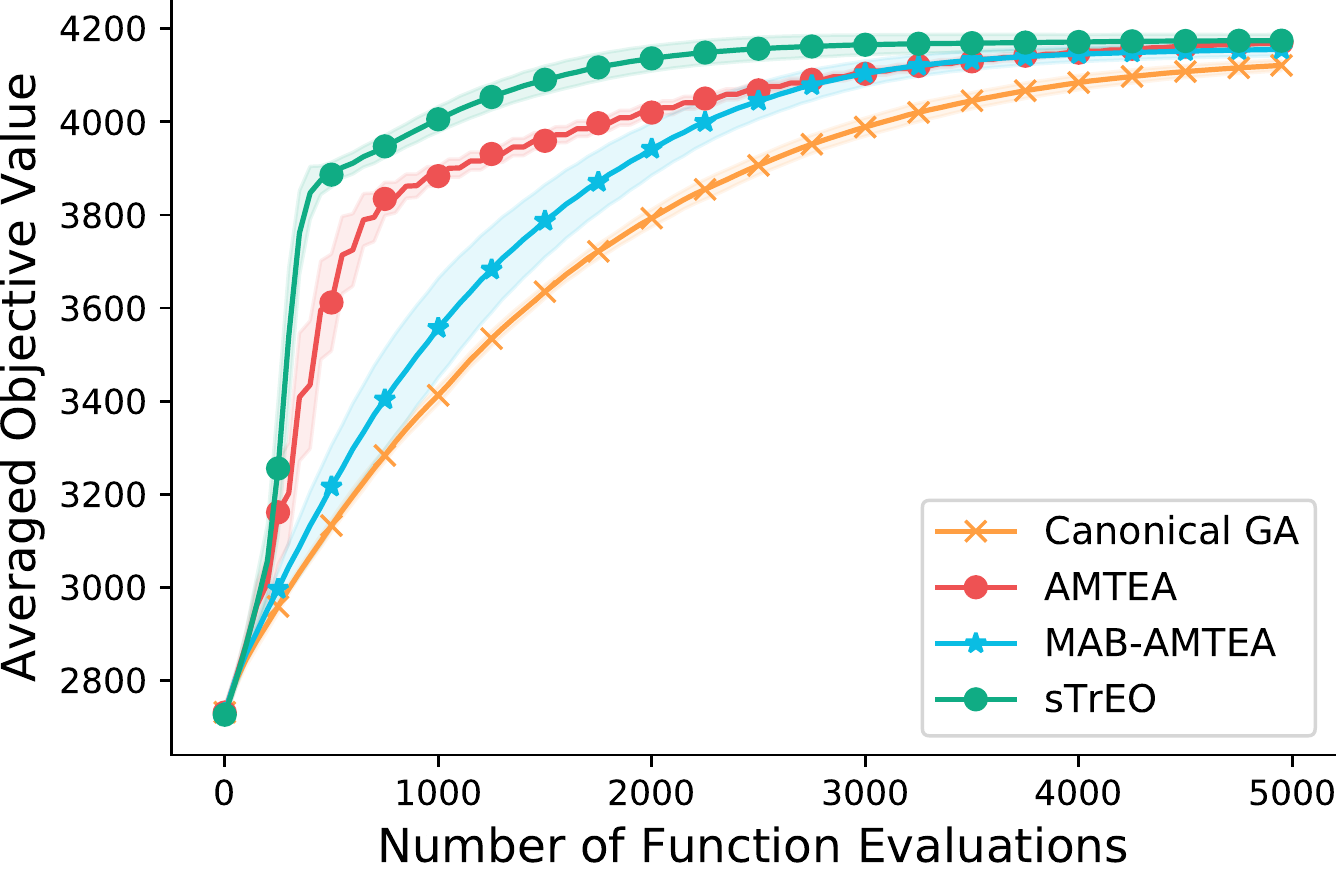}
  \caption{\scriptsize Convergence trends (40 related)}\label{FigKP1000-40G}
 \end{subfigure}
 \begin{subfigure}[b]{0.49\columnwidth}
  \centering
    \includegraphics[width=\textwidth]{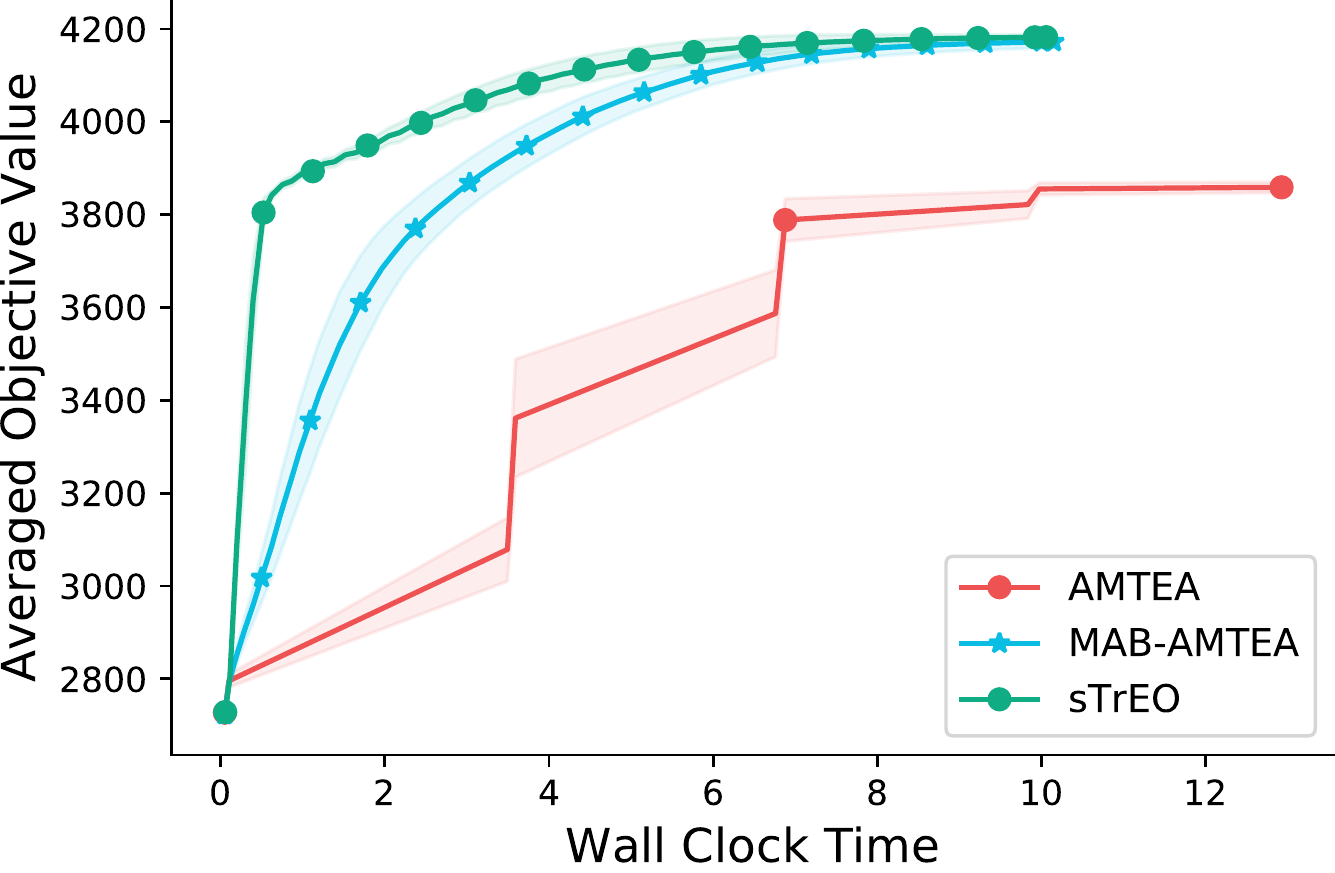}
    \caption{\scriptsize Convergence times (250 related)}\label{FigKP1000-250T}
 \end{subfigure}
 \begin{subfigure}[b]{0.49\columnwidth}
 \centering
   \includegraphics[width=\textwidth]{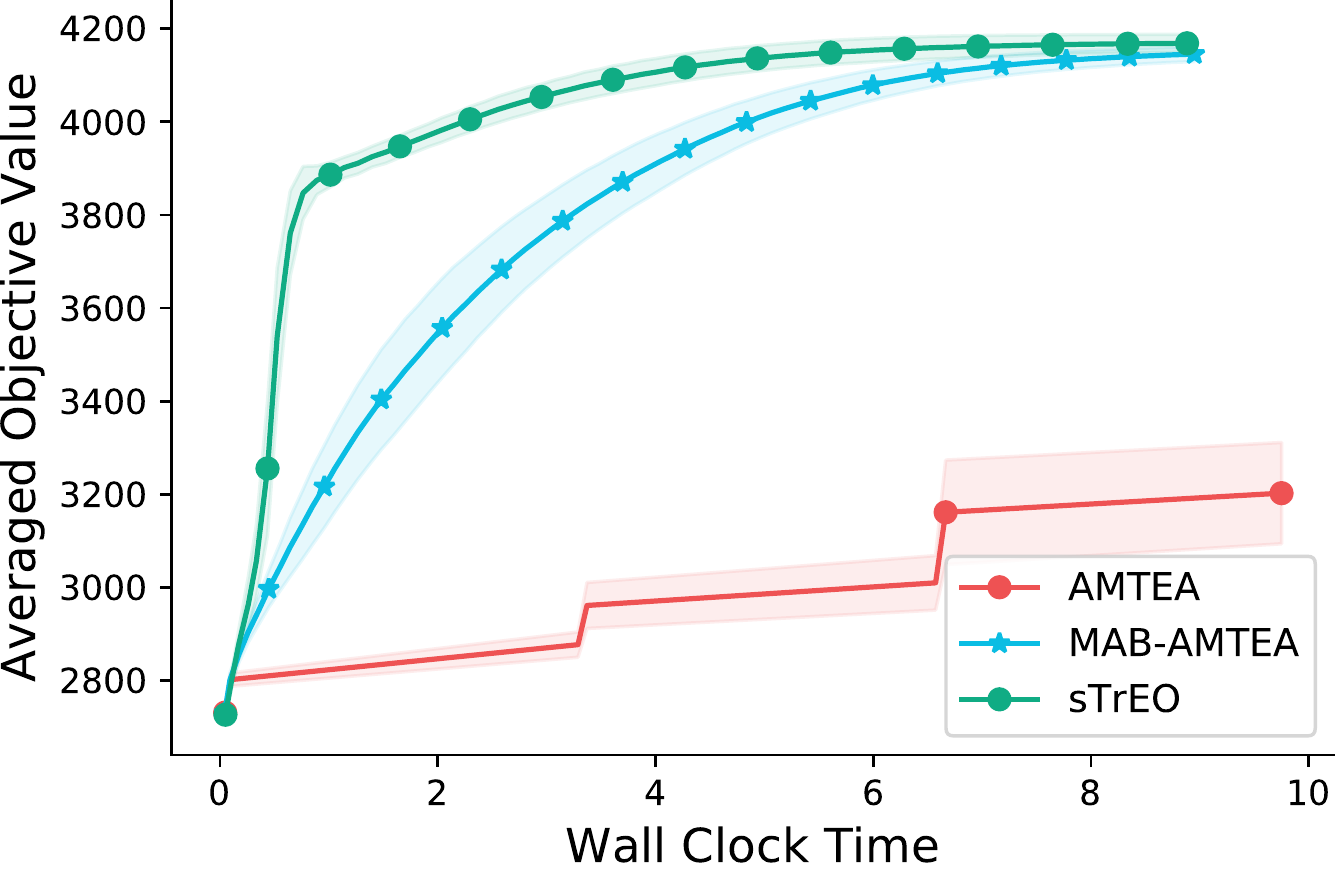}
  \caption{\scriptsize Convergence times (40 related)}\label{FigKP1000-40T}
 \end{subfigure}
\caption{Averaged performance results over 30 independent runs for the 0/1 knapsack problem with 1000 source instances. Convergence trends relative to the number of function evaluations for (a) 250 related sources and (b) 40 related sources. Convergence trends relative to sTrEO's convergence time for (c) 250 related sources and (d) 40 related sources. The shaded region indicates standard deviations either side of the mean.}\label{FigKP}
\end{figure}

\begin{figure}
  \begin{subfigure}[b]{0.49\columnwidth}
  \centering
    \includegraphics[width=\textwidth]{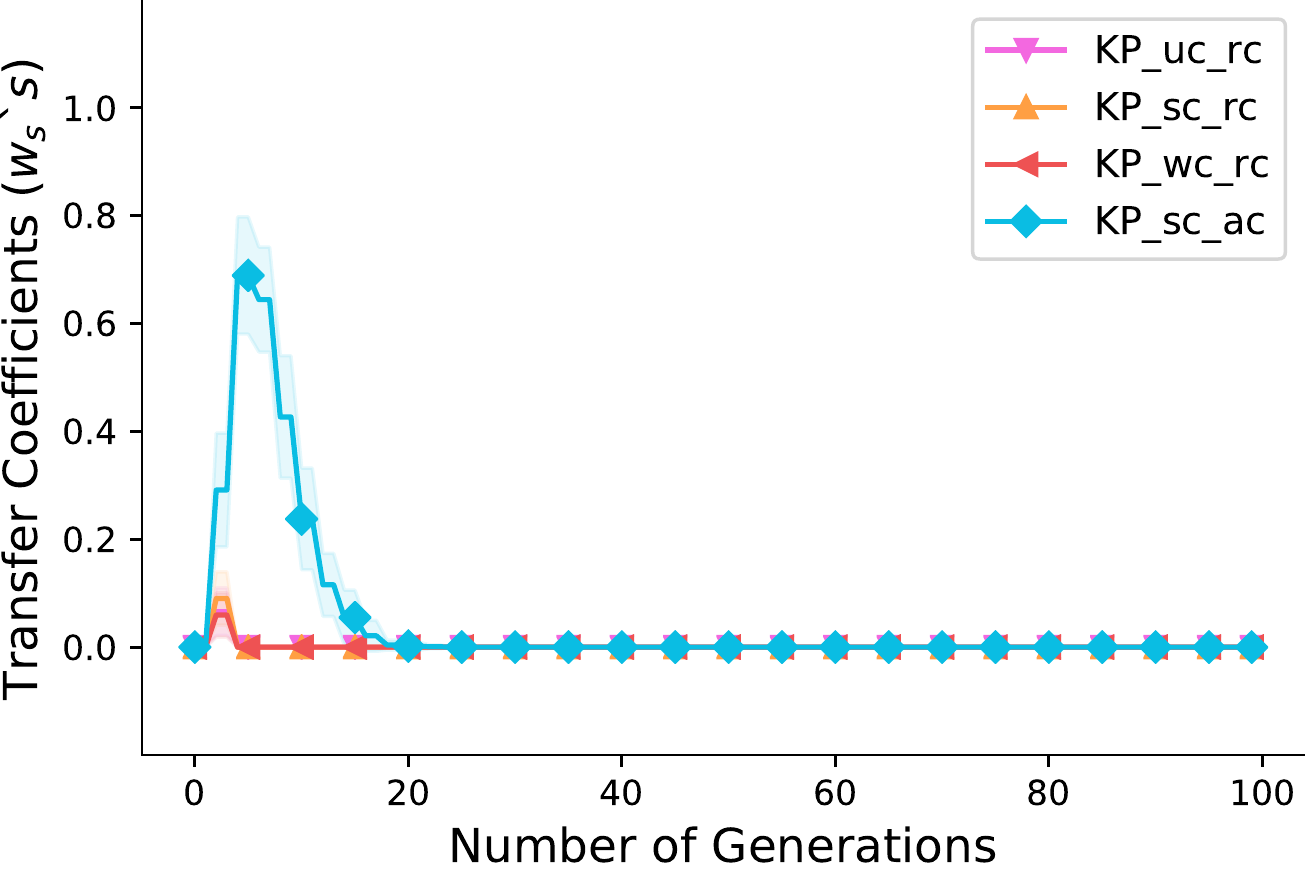}
    \caption{\scriptsize AMTEA's learned $w_{s}$'s (250 related)}\label{AMTEALrn1000-250}
 \end{subfigure}
 \begin{subfigure}[b]{0.49\columnwidth}
 \centering
   \includegraphics[width=\textwidth]{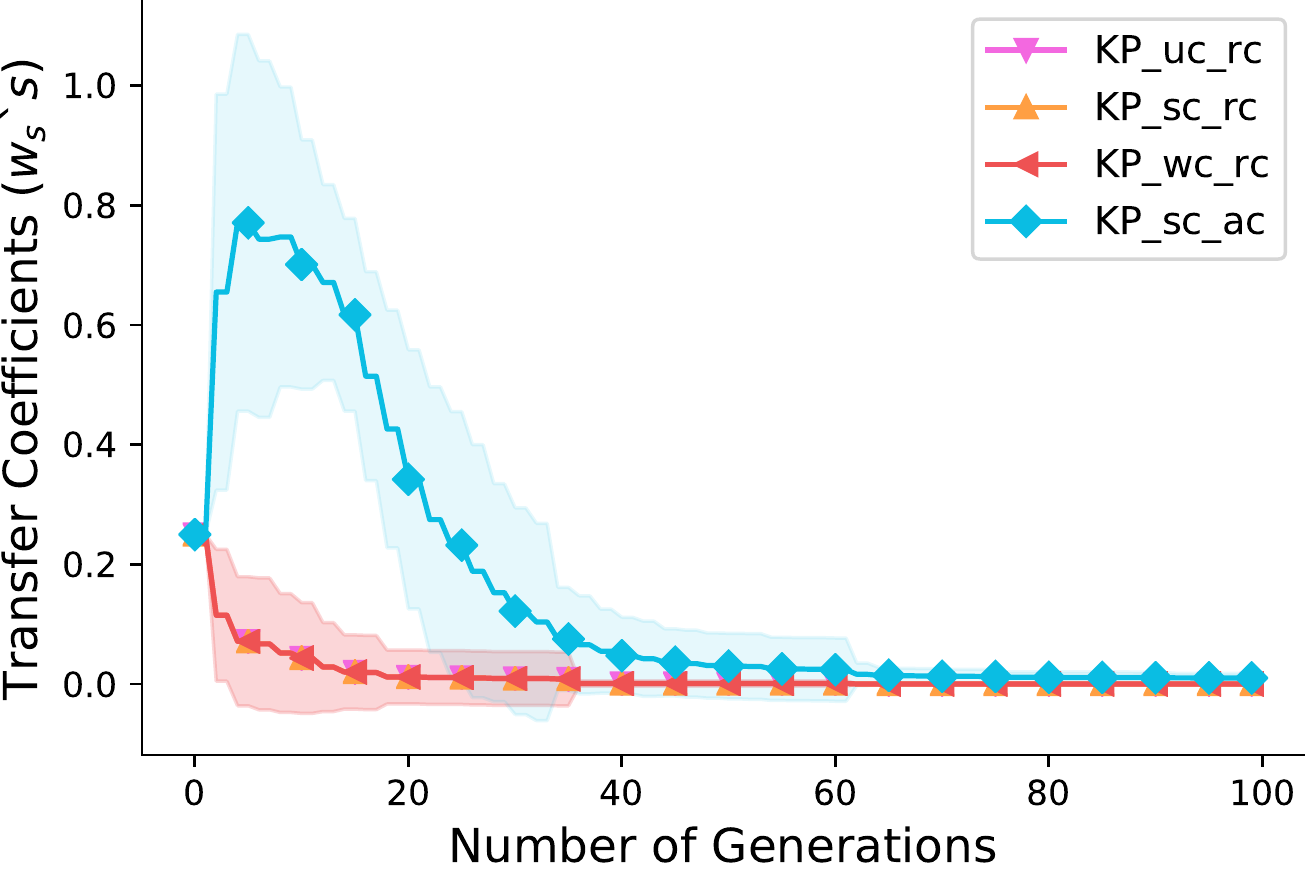}
  \caption{\scriptsize sTrEO's learned $w_{s}$'s (250 related)}\label{CMTEALrn1000-250}
 \end{subfigure}
  \begin{subfigure}[b]{0.49\columnwidth}
  \centering
    \includegraphics[width=\textwidth]{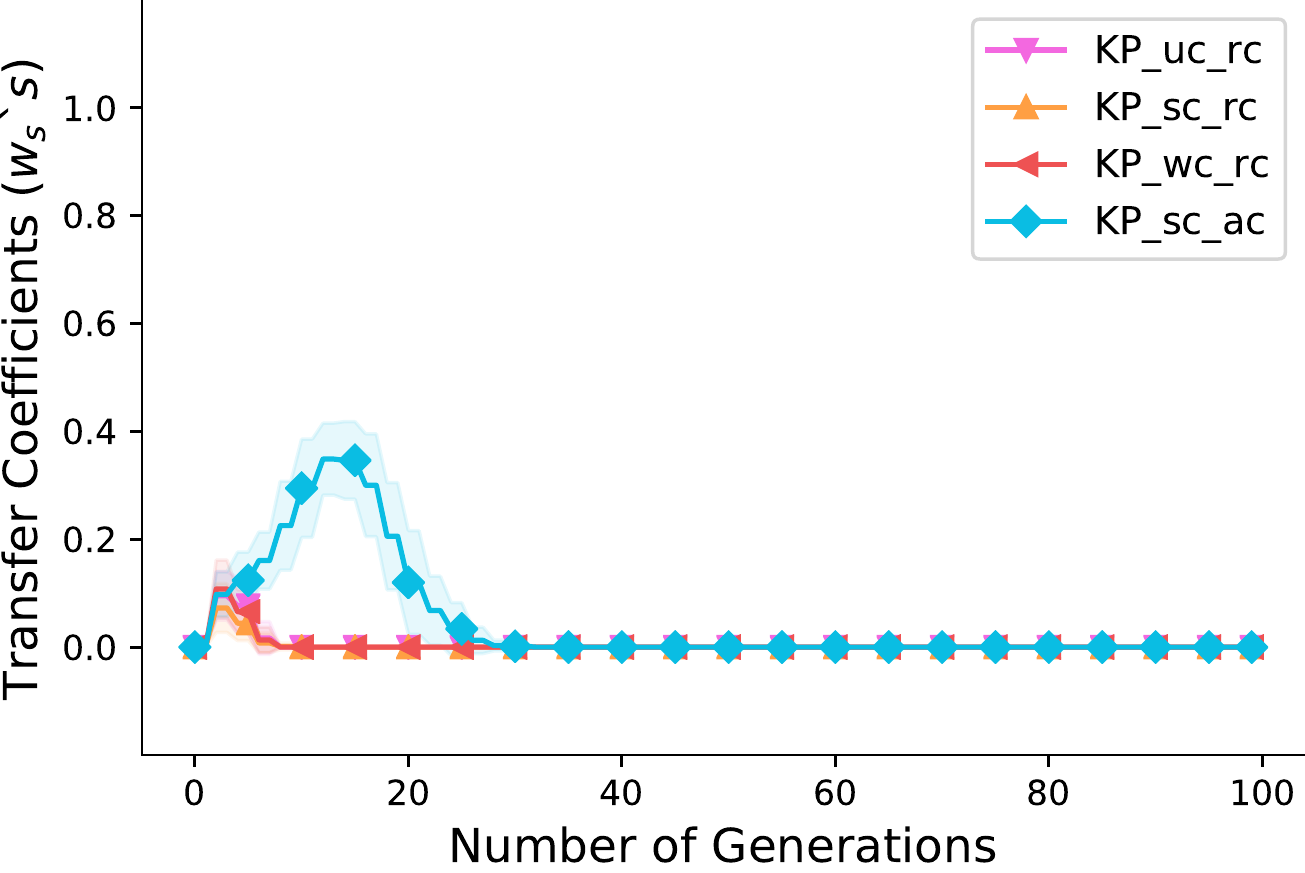}
    \caption{\scriptsize AMTEA's learned $w_{s}$'s (40 related)}\label{AMTEALrn1000-40}
 \end{subfigure}
 \begin{subfigure}[b]{0.49\columnwidth}
 \centering
   \includegraphics[width=\textwidth]{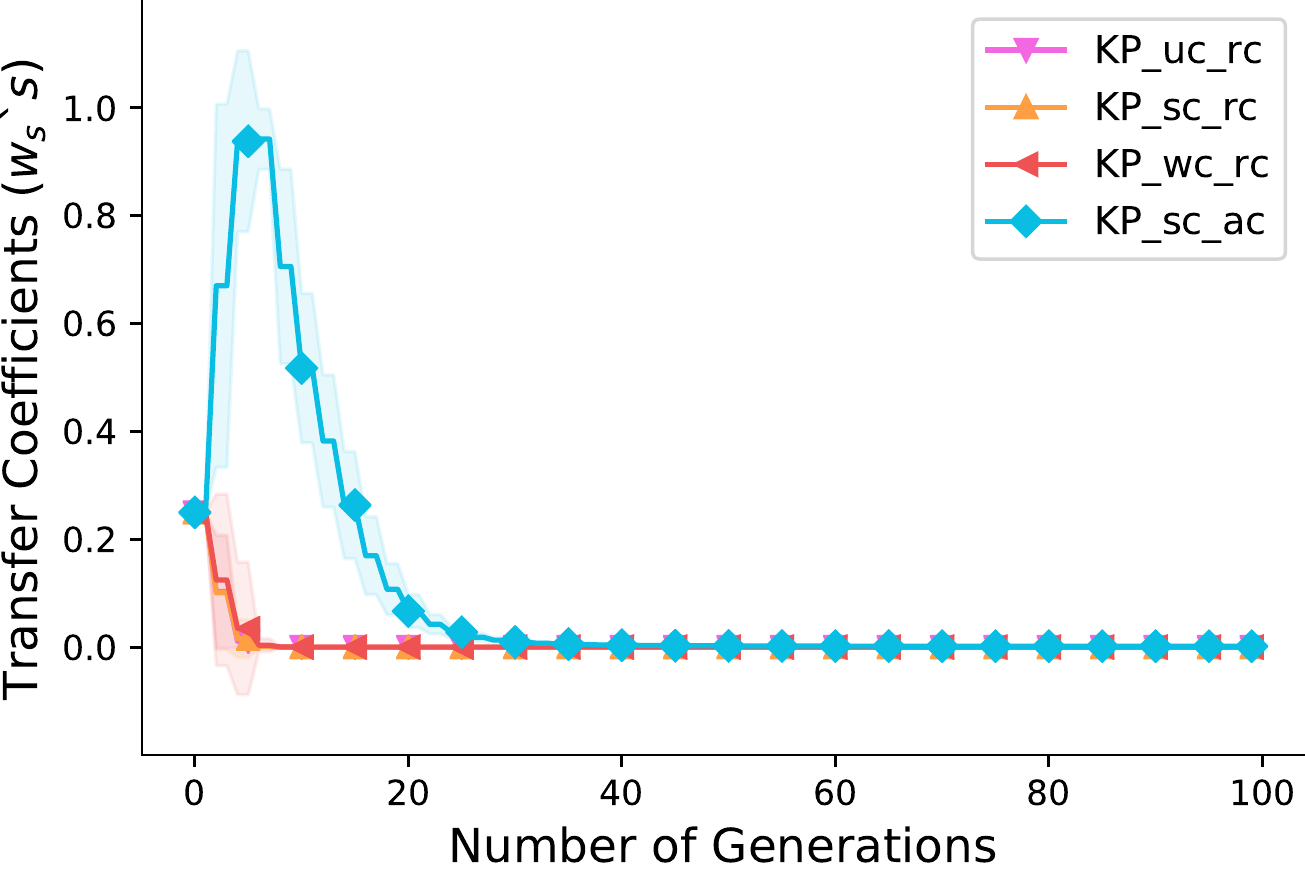}
  \caption{\scriptsize sTrEO's learned $w_{s}$'s (40 related)}\label{CMTEALrn1000-40}
 \end{subfigure}
\caption{Learned transfer coefficients of AMTEA and sTrEO while solving the 0/1 knapsack problem with 1000 sources. The target is ``KP\_uc\_ac'' and the source types are ``KP\_uc\_rc'', ``KP\_wc\_rc'', ``KP\_sc\_rc'' and ``KP\_sc\_ac''. (a) AMTEA's learned $w_{s}$'s with 250 related sources. (b) sTrEO's learned $w_{s}$'s with 250 related sources. (c) AMTEA's learned $w_{s}$'s with 40 related sources. (d) sTrEO's learned $w_{s}$'s with 40 related sources. Results were averaged over 30 independent runs. The shaded region indicates standard deviations either side of the mean.}\label{FigKPLearn}
\end{figure}

Two evaluation scenarios were defined to investigate the performance of sTrEO with respect to scalability and online learning agility. The scenarios are described as follows:
\begin{itemize}
	\item \emph{Scenario (A):} where the number of sources is $1000$ with $250$ instances of the related type and $750$ instances equally distributed among the three unrelated types (the source-target relatedness ratio is $0.25$).
	\item \emph{Scenario (B):} where the number of sources is $1000$ with only $40$ instances of the related type and $960$ instances equally distributed among the three unrelated types (the source-target relatedness ratio is $0.04$).
\end{itemize}

All source instances were first optimized by a binary CGA where the final derived solutions were used to construct the probabilistic models. We then compared the performance of our proposed sTrEO with CGA, AMTEA, MAB-AMTEA for both evaluation scenarios. The averaged results obtained over 30 independent runs are shown in Fig. \ref{FigKP}. With respect to convergence efficiency, Figs. \ref{FigKP1000-250G} and \ref{FigKP1000-40G} present the convergence trends of the compared algorithms for Scenario (A) with 250 related sources and Scenario (B) with 40 related sources, respectively. It can be seen from the convergence trends that utilization of relevant knowledge from previously solved experiences in essence does improve optimization efficiency in the target problem. The effectiveness in learning source-target similarities, is more noticeable for AMTEA and sTrEO given that both acquire a steeper convergence curve than MAB-AMTEA in early function evaluations. The reason for such different convergence trends lies in how agile the online source-target similarity learning mechanism of the compared methods is to discern related sources and exploit useful knowledge accordingly. We can therefore conclude that MAB-AMTEA fails to reach that level of agility as opposed to AMTEA and sTrEO. This is more evident for Scenario (B) given sparsity of relevant sources. The advantage of such capability (i.e., to achieve convergence in fewer number of function evaluations) would be more significant for applications where the evaluation of the objective function is very costly.

\begin{figure*}
  \begin{subfigure}[b]{0.33\textwidth}
    \includegraphics[width=0.95\textwidth]{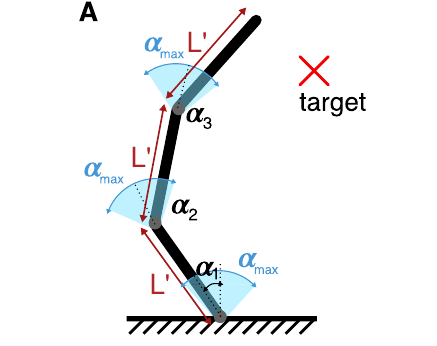}
    \caption{\scriptsize Kinematic Arm}\label{FigKinArm}
  \end{subfigure}  
  \begin{subfigure}[b]{0.33\textwidth}
    \includegraphics[width=\textwidth]{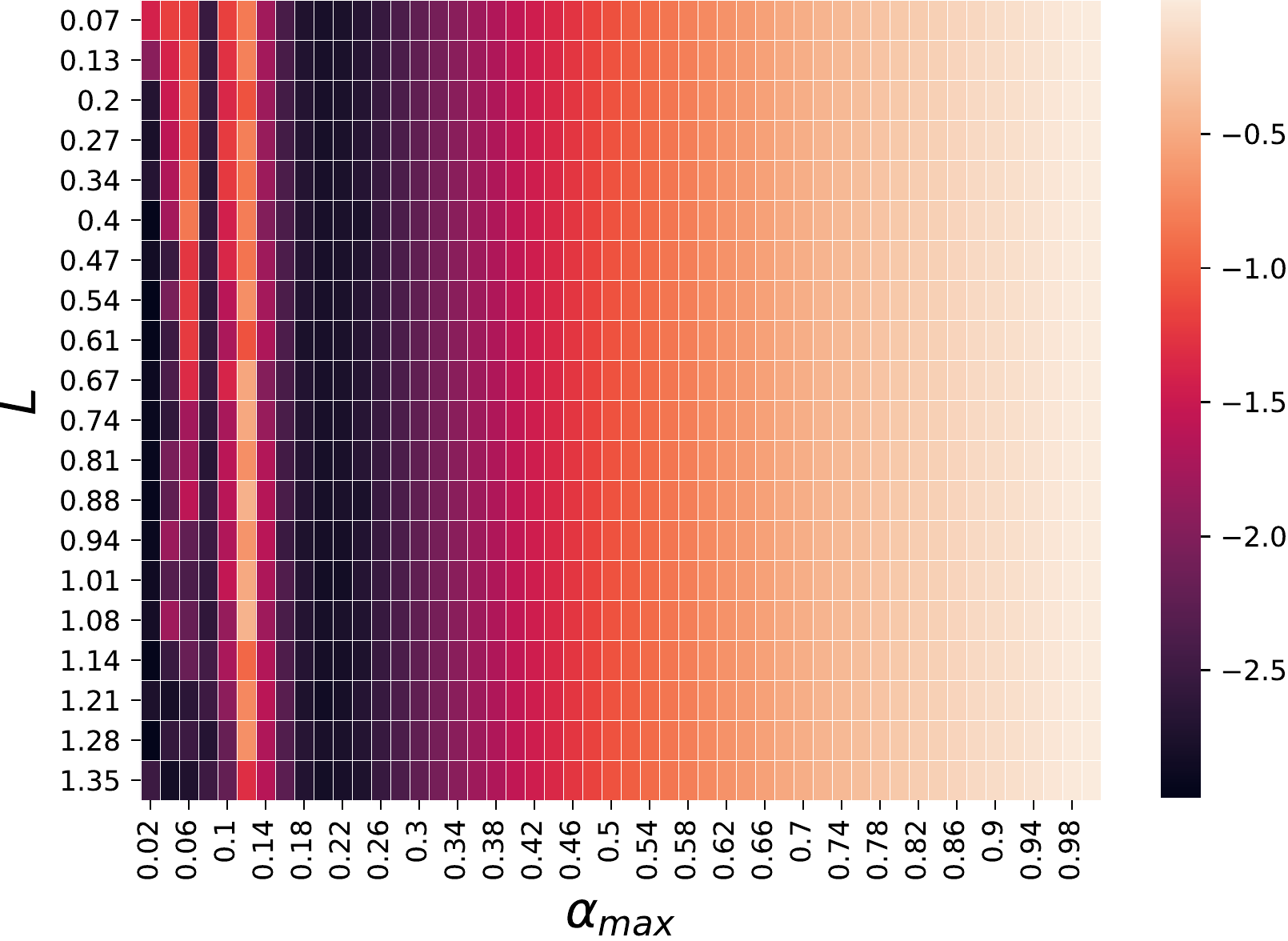}
    \caption{\scriptsize 10 joints}\label{FigHM10Joint}
  \end{subfigure}
  \begin{subfigure}[b]{0.33\textwidth}
    \includegraphics[width=\textwidth]{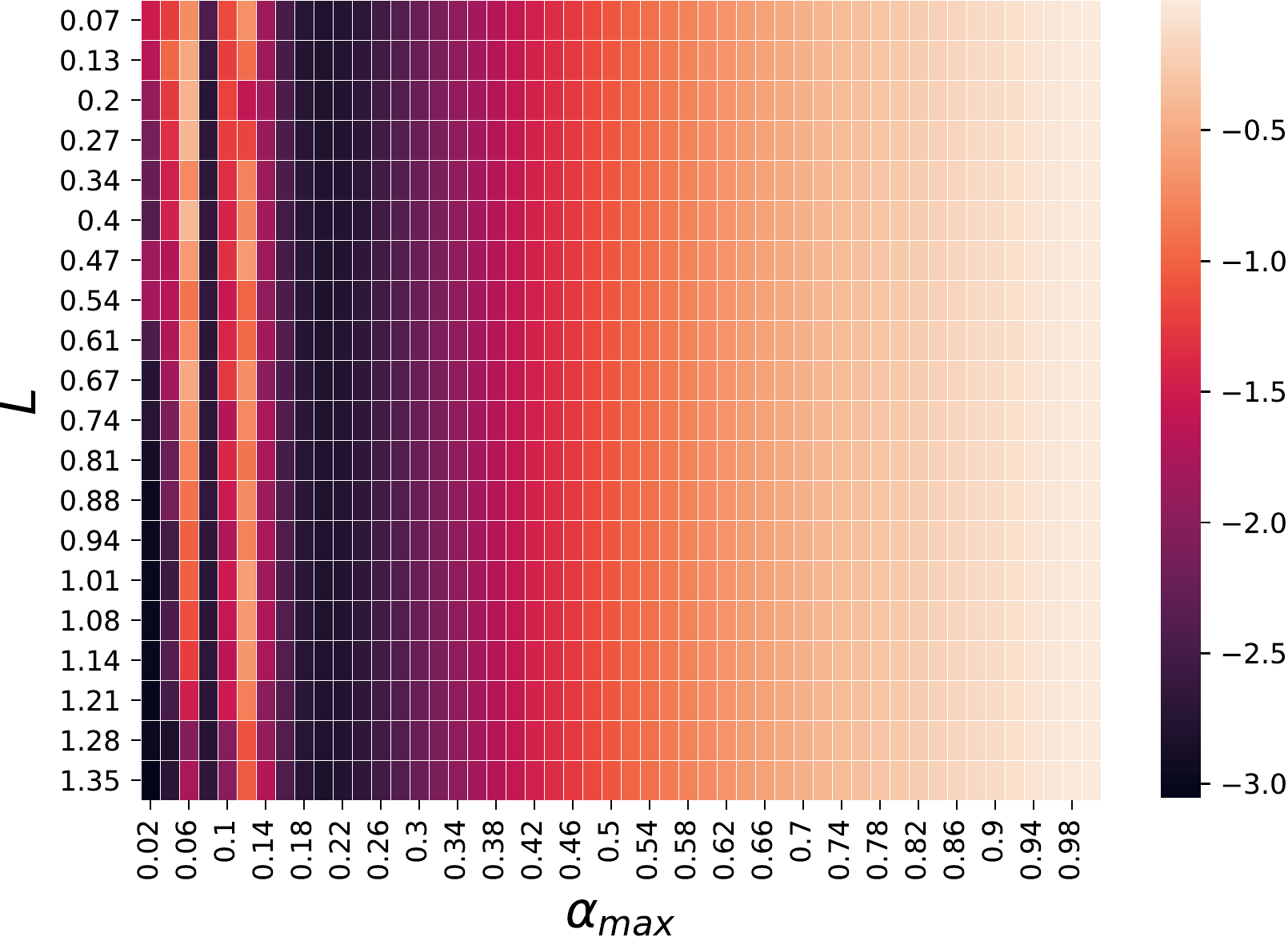}
    \caption{\scriptsize 20 joints}\label{FigHM20Joint}
    \label{fig:2}
  \end{subfigure}
 \caption{(a) A planar robotic arm with three joints (and equal links) \cite{mouret2020quality}. The objective is to find the angle of each joint (here, $\alpha_{1},\alpha_{2},\alpha_{3}$) such that the end-effector is as close as possible to a predefined target in the plane. Heatmaps of source-target relatedness for diverse combinations of $L$ and $\alpha_{\textrm{max}}$, ranging between $(0,\sqrt{2})$ and $(0,1]$, respectively, for (b) $10$ joints and (c) $20$ joints. According to the sidebar on the right which marks the fitness range, the hotter cells indicate more relatedness to the target (as their mean fitness values are greater) whereas the colder ones show less or no relatedness.}\label{FigKinArmDef}
\end{figure*}

We also present the convergence trend of the three compared TrEO methods within the convergence time of sTrEO as shown in Figs. \ref{FigKP1000-250T} and \ref{FigKP1000-40T} for Scenarios (A) and (B), respectively. 
While the MAB-AMTEA is able to converge within this time limit\footnote{Note that the rationale behind proposing MAB-AMTEA in \cite{shakeri2019coping} is to tackle the scaling burden of AMTEA when exposed to big task-instances.}, AMTEA falls much behind the other two and requires a considerably longer time to eventually converge to optimality, particularly for Scenario (B) with the fewer number of relevant sources. The above observations indicate that neither AMTEA nor MAB-AMTEA are able to ensure scalability as well as online learning agility \emph{simultaneously}. The AMTEA lacks scalability due to its computationally intensive EM-based source-target similarity learning module whereas the MAB-AMTEA's source selection strategy, based on the EXP3, is not effective to quickly identify and utilize related sources. 
Lastly, we present the trajectory of the learned transfer coefficients of AMTEA and sTrEO (aggregated per source knapsack type) in Fig. \ref{FigKPLearn}. As can be seen from Figs. \ref{AMTEALrn1000-250} to \ref{CMTEALrn1000-40}, sTrEO can learn qualitatively similar trends of transfer coefficient values as AMTEA. (Recall that AMTEA is provably optimal whereas sTrEO follows a randomized (1+1)-ES procedure.) The results show that despite compromising on optimality, sTrEO achieves comparable performance, while simultaneously achieving significantly lower computational cost.

\subsection{Robotic Arm with Variable Morphology}\label{SecKinArm}
We consider a planar robotic arm adopted from \cite{mouret2020quality} as our second case study. In this setting, there is an arm of length $L$ with $d$ joints as equal as the number of its links. (Here $d$ specifies the dimensionality of the problem.) Each joint can rotate to a maximum angle $\alpha_{\textrm{max}}$, encoded in $(0,1]$, which is the same for all joints. Further, all the links have the identical length (here, $L^{\prime}$). The problem objective is to find the angle of each joint of the robot (i.e., $\alpha_{1},\dots,\alpha_{d}$) such that the tip of the arm (i.e., the end-effector) is as close as possible to a predefined target in the plane (see Fig. \ref{FigKinArm}).

We identify a task $\mathcal{T}$ by a particular combination of $L$ and $\alpha_{\textrm{max}}$ in which a candidate solution is encoded by a vector $\pmb{\alpha}=\alpha_{1},\dots,\alpha_{d}$. The objective function $f(\pmb{\alpha},\mathcal{T}_{L,\alpha_{\textrm{max}}})$ to evaluate the solution $\pmb{\alpha}$ to the task $\mathcal{T}_{L,\alpha_{\textrm{max}}}$ computes the Euclidean distance from the tip position $p_{d}$ to the target position $T$. The $p_{d}$ can be obtained recursively as follows \cite{mouret2020quality}:
\begin{equation}\label{eq7}
M_{0}=I,
\end{equation}
\begin{equation}\label{eq8}
M_{i} = M_{i-1}\cdot
 \begin{pmatrix}
  \cos{\alpha^{\prime}_{i}} & -\sin{\alpha^{\prime}_{i}} & 0 & L^{\prime} \\
  \sin{\alpha^{\prime}_{i}} & \cos{\alpha^{\prime}_{i}} & 0 & 0 \\
  0  & 0 & 1 & 0  \\
  0 & 0 & 0 & 1 
 \end{pmatrix},
\end{equation}
\begin{equation}\label{eq9}
p_{i}=M_{i}\cdot(0,0,0,1)^\intercal,
\end{equation}
where, $\alpha^{\prime}_{i}=2\pi\cdot\alpha_{\textrm{max}}\cdot(\alpha_{i}-0.5)$, $\forall i\in\{1,\dots,d\}$ and $L^{\prime}=L/d$. Given the above, the objective function, which calculates the distance between the end of the last link and the target position, can be presented as,
\begin{equation}\label{eq10}
f(\pmb{\alpha},\mathcal{T}_{L,\alpha_{\textrm{max}}})=-\parallel p_{d}-T\parallel.
\end{equation}
Here, we arbitrary set the target $T$ to $(1,1)$ in the plane. Note that the objective function is negated in (\ref{eq10}) so that the problem can be treated as one of maximization.

We consider two different scenarios with $10$ and $20$ joints in the experiments. The target task is defined as $\mathcal{T}_{\sqrt{2},1}$ for both settings, ending up with the largest solution space possible as each joint can rotate from $\minus\pi$ to $\pi$. To identify the sources and their relatedness to the target, we defined $1000$ source instances with diverse combinations of $L$ and $\alpha_{\textrm{max}}$, ranging between $(0,\sqrt{2})$ and $(0,1]$, respectively. One of the sources was first optimized by a continuous CGA where the final derived solutions were used to build the probabilistic model. The probabilistic models of the remaining sources were built after successively utilizing the AMTEA to optimize one source instance by exploiting the already solved ones. We then sampled $100$ solutions from each source model and evaluated them using the objective function of the target task. The average objective value to each source forms a cell of a heatmap, as shown in Figs. \ref{FigHM10Joint} and \ref{FigHM20Joint} for $10$ and $20$ joints, respectively. Here, the hotter cells indicate more relatedness to the target (as their mean fitness values are greater) whereas the colder ones show less or no relatedness. Having identified how source-target relatedness is shaped for each scenario, we randomly chose $15$ instances with $0<L<\sqrt{2}$ and $\alpha_{\textrm{max}}=1$ as \emph{related} and $985$ ones with $0<L<\sqrt{2}$ and $0.18<\alpha_{\textrm{max}}<0.26$ as \emph{unrelated} to form a total of $1000$ sources per scenario. We followed the same procedure, as already outlined above, to build the probabilistic models of the sources. 

Given that the source-target similarity learning module described in Procedure \ref{CMTkExtracProcess} is for the case of a maximization problem, we need some refinements to apply sTrEO for the examples whose underlying optimization problem is one of minimization and $\minus f_{T}(\pmb{x})$ is alternatively being used. The first amendment is on the initial values of the mutation vector. Rather than setting all the values to zeroes, we initialize the mutation vector with a \emph{lower bound} on the fitness value of the target problem. For the robotic arm case study, this was set to the fitness value of the worst feasible solution to the target, which is $\minus2\sqrt{2}$ when the the tip of the arm falls in the opposite direction to the target position. Such adjustment led to more effective exploitation of related sources as derived from our preliminary experiments. The second amendment is to shift the values of the mutation vector to the range $[0,+\infty)$ before applying the maximum absolute scaling function in line 13 in Procedure \ref{CMTkExtracProcess}. This is done by adding all entries by the absolute value of their minimum.  

\begin{figure}
  \begin{subfigure}[b]{0.49\columnwidth}
  \centering
    \includegraphics[width=\textwidth]{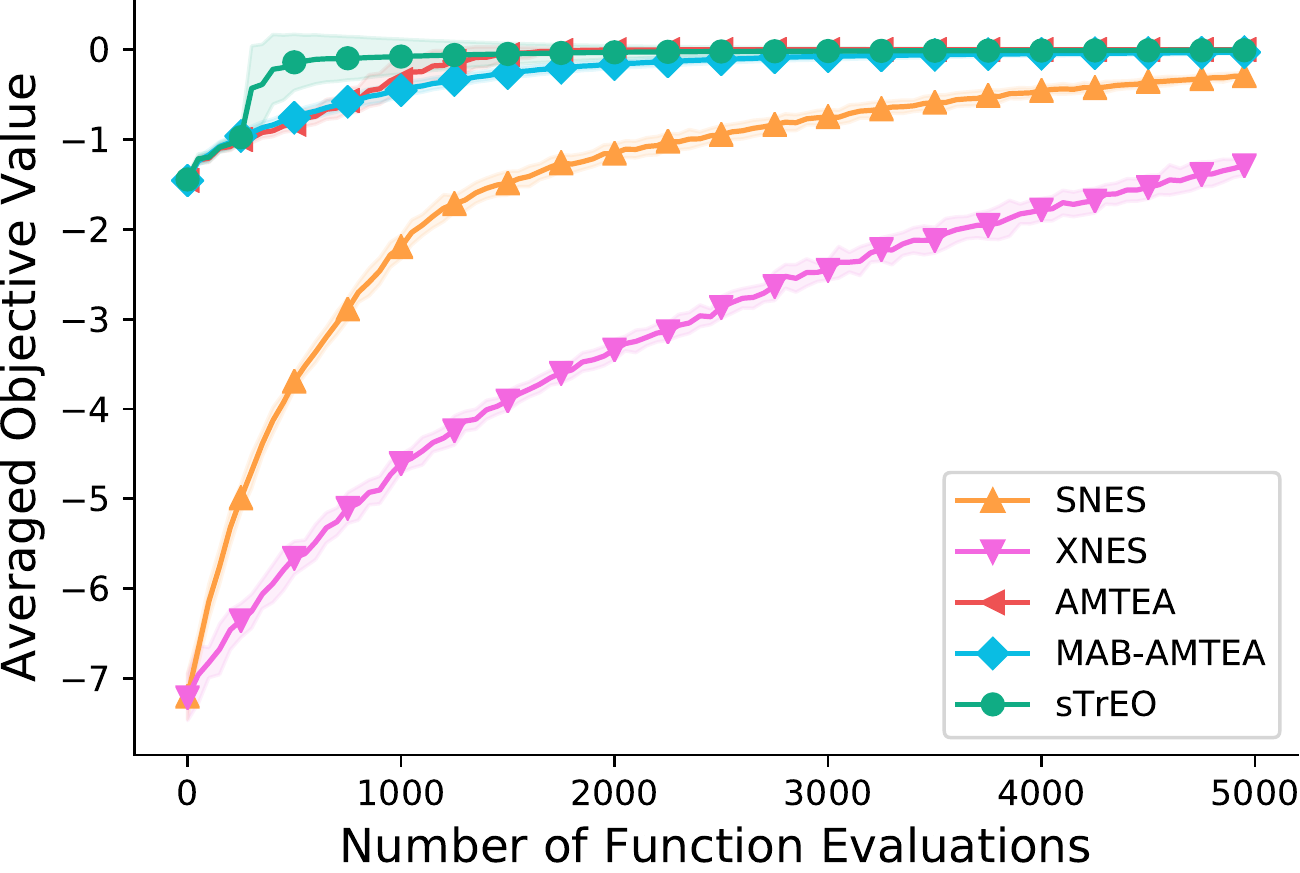}
    \caption{\scriptsize Convergence trends (10 joints)}\label{FigKA1000-15G10}
 \end{subfigure}
 \begin{subfigure}[b]{0.49\columnwidth}
 \centering
   \includegraphics[width=\textwidth]{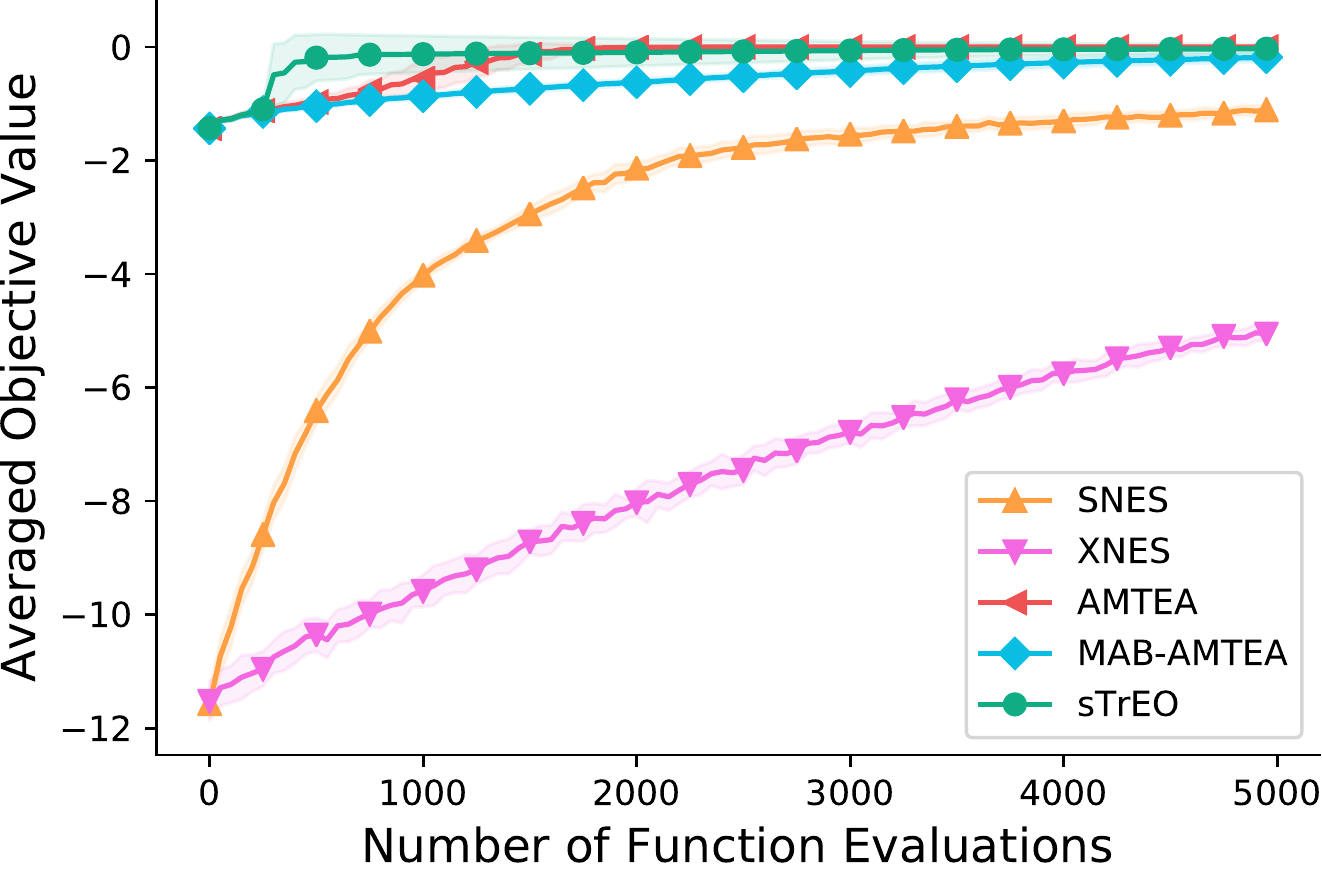}
  \caption{\scriptsize Convergence trends (20 joints)}\label{FigKA1000-15G20}
 \end{subfigure}
 \begin{subfigure}[b]{0.49\columnwidth}
  \centering
    \includegraphics[width=\textwidth]{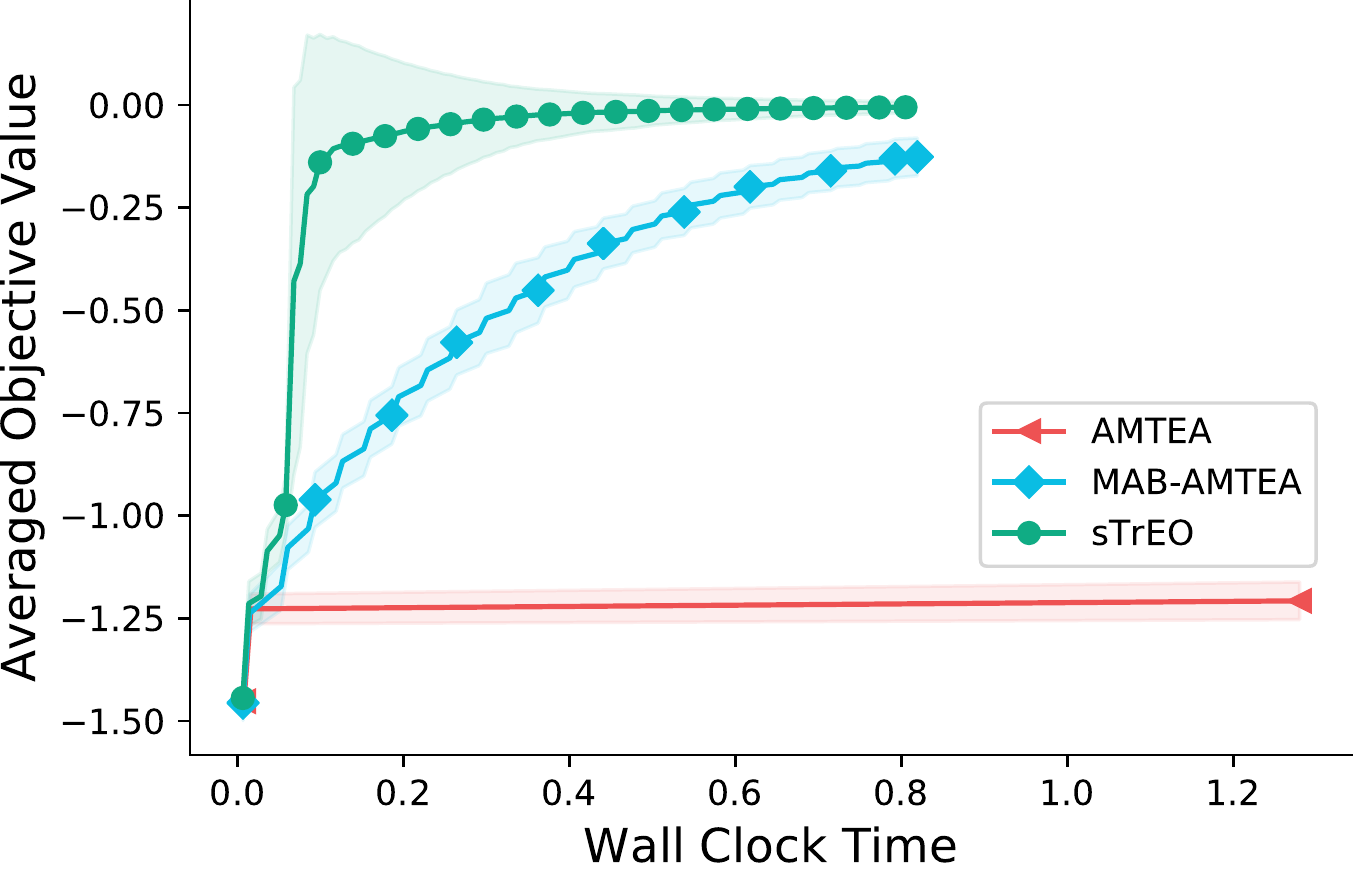}
    \caption{\scriptsize Convergence times (10 joints)}\label{FigKA1000-15T10}
 \end{subfigure}
 \begin{subfigure}[b]{0.49\columnwidth}
 \centering
   \includegraphics[width=\textwidth]{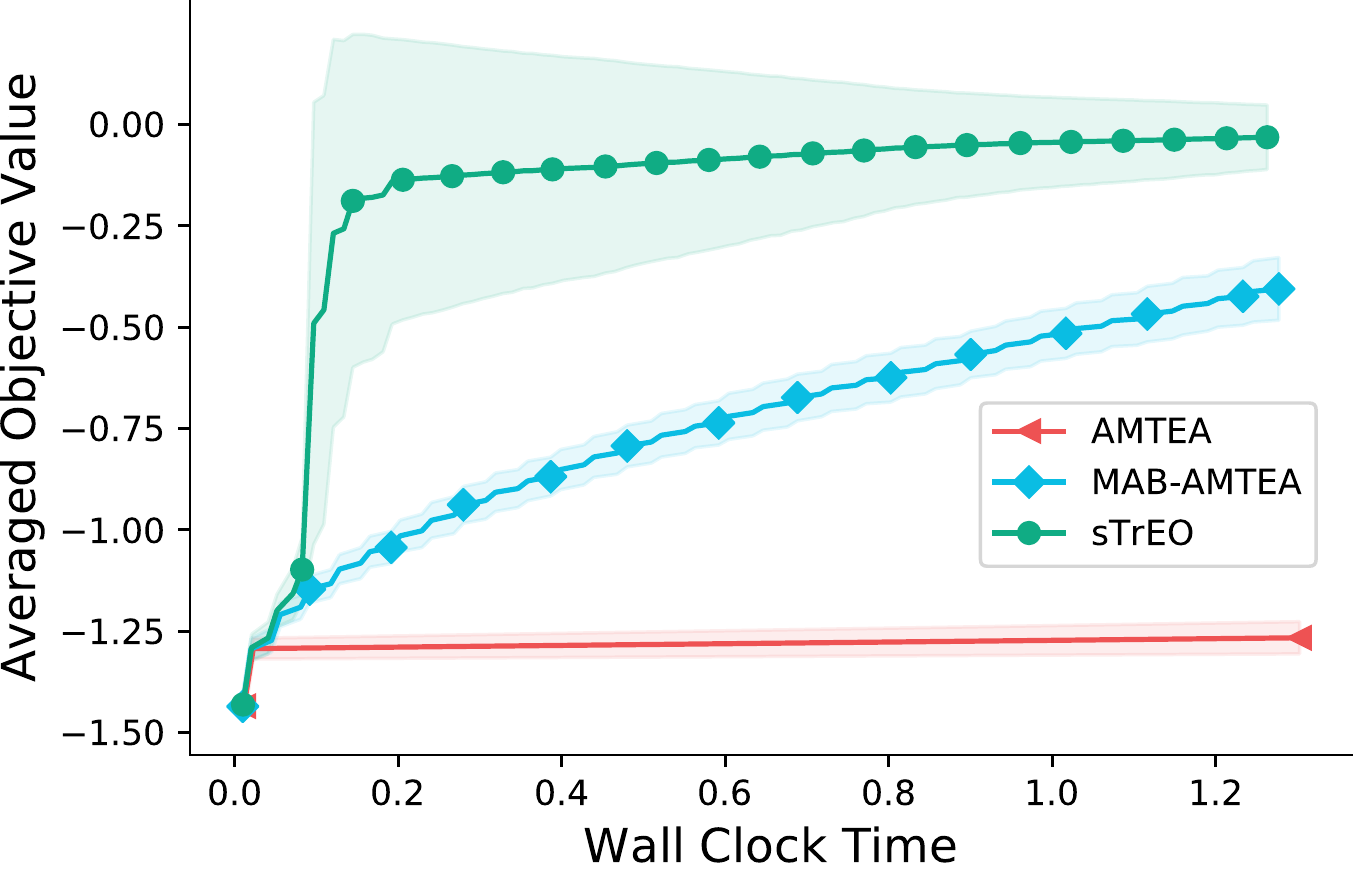}
  \caption{\scriptsize Convergence times (20 joints)}\label{FigKA1000-15T20}
 \end{subfigure}
\caption{Averaged performance results over 30 independent runs for the robotic arm example. The target task is $\mathcal{T}_{\sqrt{2},1}$. There are 1000 source instances, 15 of which are related. Convergence trends relative to the number of function evaluations for (a) 10 joints and (b) 20 joints. Convergence trends relative to sTrEO's convergence time for (c) 10 joints and (d) 20 joints. The shaded region indicates standard deviations either side of the mean.}\label{FigKA}
\end{figure}

The performance of sTrEO was compared with sNES and xNES, as algorithms without transfer capability, in addition to AMTEA and MAB-AMTEA for both 10- and 20-joint evaluation scenarios. A penalty function is embedded in all the compared algorithms to penalize those solutions that violate the $\alpha_{\textrm{max}}$ constraint. The averaged results obtained over 30 independent runs are shown in Fig. \ref{FigKA}. With regard to convergence efficiency, Figs. \ref{FigKA1000-15G10} and \ref{FigKA1000-15G20} present the convergence trends of the compared algorithms for $10$ and $20$ joints, respectively. Both plots strongly advocate the advantage of knowledge transfer to improve optimization efficiency, making possible the convergence to optimality within a limited computational budget. Aside from that, we can notice a small spike in averaged objective values in early function evaluations of sTrEO, further validating our previous findings on online learning agility of the nested (1+1)-ES relative to sparsity of related sources. Similar to the previous case study, we analyzed the scalability of sTrEO against AMTEA and MAB-AMTEA by presenting the convergence trend of the three algorithms within STrEO's convergence time, as shown in Figs. \ref{FigKA1000-15T10} and \ref{FigKA1000-15T20} for 10 and 20 joints, respectively. Interestingly, while consistent results can be observed given the computational efficiency of sTrEO, we can notice that the MAB-AMTEA, despite being scalable, fails at establishing optimality within STrEO's convergence time. This indicates that the bandit-based source selection mechanism to stack only a single source model to the target each time the similarity learning procedure is launched is not effective to quickly utilize relevant sources to ensure optimality. 

\begin{figure}
 \begin{subfigure}[b]{0.49\columnwidth}
  \centering
    \includegraphics[width=\textwidth]{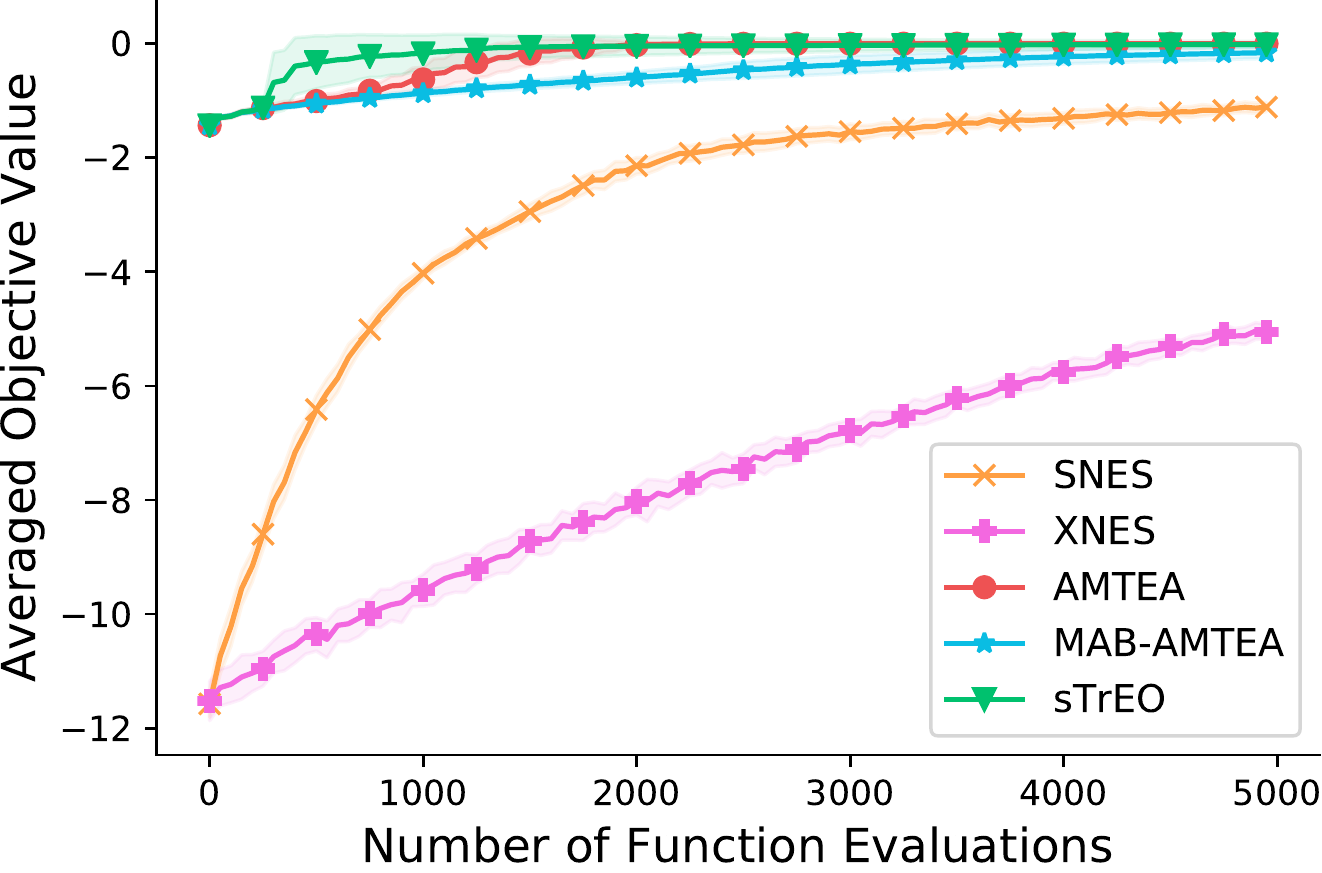}
    \caption{\scriptsize Convergence trends}\label{10000-150Gen}
 \end{subfigure}
  \begin{subfigure}[b]{0.49\columnwidth}
  \centering
    \includegraphics[width=\textwidth]{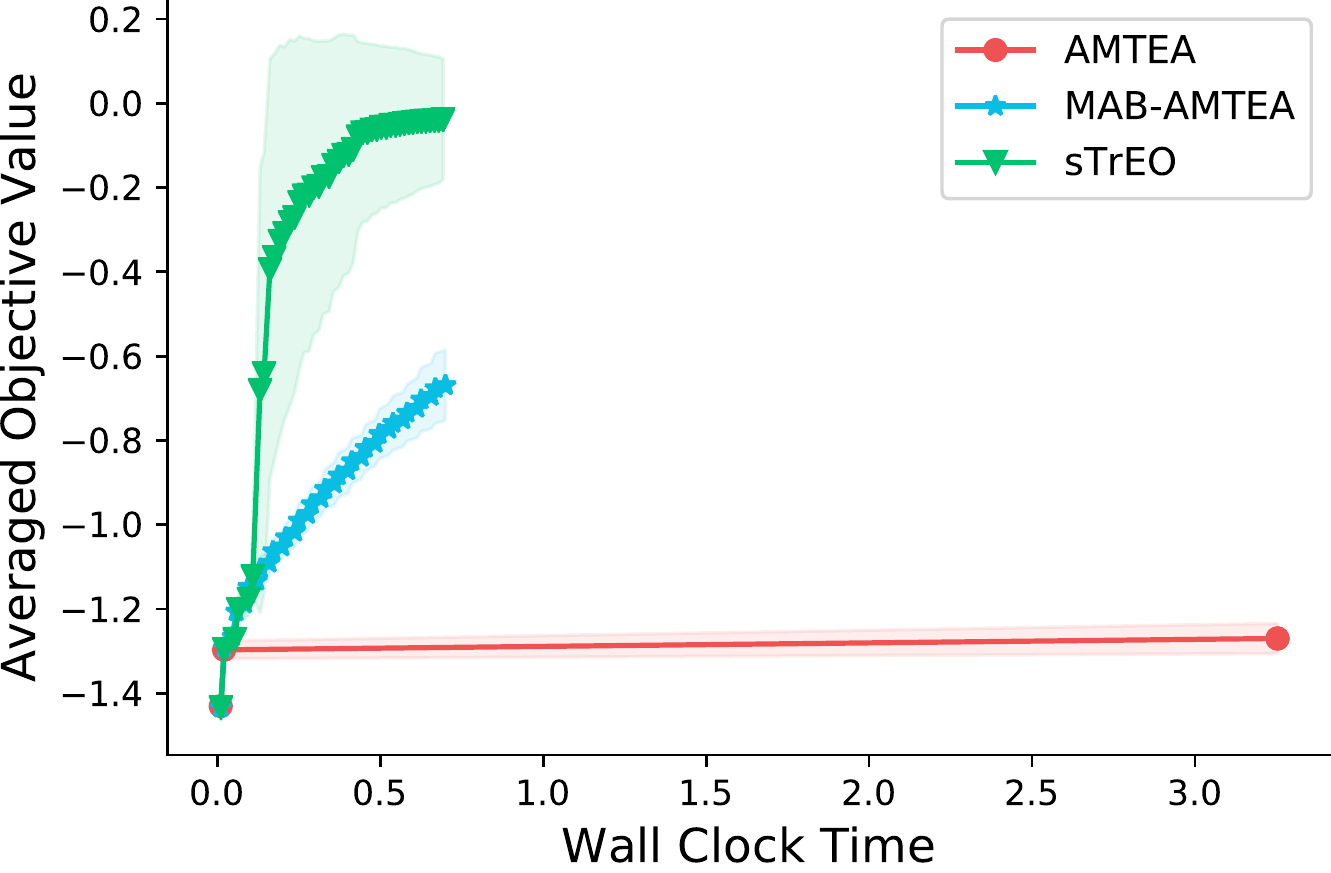}
    \caption{\scriptsize Convergence times}\label{10000-150Time}
 \end{subfigure}
 \caption{Averaged performance results over $30$ independent runs for the 20-joint robotic arm example. The target task is $\mathcal{T}_{\sqrt{2},1}$. There are 10000 source instances, 150 of which are related. Convergence trends relative to (a) number of function evaluations and (b) sTrEO's convergence time. The shaded region indicates standard deviations either side of the mean.}\label{Fig10000-150}
\end{figure}

Further experiments were carried out in Section \ref{FurtherExp} in the supplemental material to investigate the performance of sTrEO for more than 1000 source task-instances on the robotic arm with 20 joints. We leave out the details herein for the sake of brevity and present the results (averaged over 30 independent runs) for the total of 10000 sources with 150 related instances in Fig. \ref{Fig10000-150}. With respect to convergence efficiency, we can see from  Fig. \ref{10000-150Gen} that both AMTEA and sTrEO are able to utilize related sources in early function evaluations, further validating their online learning agility when exposed to big source task-instances with sparse related sources. With regard to scalability of the three compared TrEO algorithms, Fig. \ref{10000-150Time} shows their convergence trend within sTrEO's convergence time. It can be observed that unlike AMTEA, sTrEO's computational complexity is as efficient as MAB-AMTEA (despite the latter's failure in converging to optimality). AMTEA requires much longer time to converge to optimality. Following the above analysis, we can conclude that our proposed sTrEO is able to \emph{simultaneously} ensure scalability and online learning agility \emph{beyond} 1000 source task-instances.


\subsection{Double Pole Balancing Controller System}\label{SecDobPole}
Double pole balancing is a prototypical control problem commonly used as a case study for neuro-evolutionary and reinforcement learning algorithms. In this setting, two poles are hinged at the same point to a wheeled cart on a finite length track. The goal is to keep both poles balanced indefinitely by applying a force to the cart at regular intervals without causing the cart to move beyond the track boundaries. The task is more difficult than the basic single pole problem due to nonlinear interactions incurred between the two poles \cite{gomez2008accelerated}.

Six variables are defined to represent the state of the system: the angle of each pole from vertical, the angular velocity of each pole, the position of the wheeled cart on the track, and the velocity of the cart \cite{gomez1999solving}. Regarding the length of the two poles, the long one is fixed to $1$ meter whereas the short one is variable (denoted by $l$) and indicates a task $e_{l}$ under evaluation. The system is simulated according to the steps instructed in \cite{da2018curbing}. In this setting, a feedforward neural network (FNN), whose structure remains fixed during the simulation, is used to output a force to the wheeled cart periodically. The conditions for the system failure are either the cart goes out of the track boundaries or one of the two poles loses its balance and drops beyond a certain degree from the vertical. Here, a candidate chromosome represents the connection weights of neurons of the FNN controller and is assessed by measuring the number of time steps elapsed before the system fails. Clearly, the solution quality (fitness) is higher for longer time steps. We assert that a task is solved if a solution could be found such that its fitness is greater than a specified number of time steps. Following \cite{da2018curbing}, this value was set to $100,000$ time steps in our experiments, which is over $30$ minutes in simulated time. The architecture of the FNN controller is identical to one applied in \cite{da2018curbing}, i.e., a two-layer network with ten hidden neurons and no bias parameters. This forms a total of 70 weights encoded in the solution chromosome of the evolutionary optimizer.

An established fact concerning the double pole system is that the problem becomes harder to solve when the two poles are very close in length (here, $l$, the length of the shorter pole, approaches $1$m, which is the length of the longer pole). Previous studies suggest incremental learning techniques in which the length of the shorter pole is increased very gradually and then the resultant problem is solved incrementally \cite{gomez1999solving}. The analogy between such incremental problem solving and transfer optimization paradigm was realized in \cite{da2018curbing} to successfully apply the AMTEA to solve more difficult variants of the problem by utilizing the solutions to easier solved ones. Accordingly, we consider this practical problem as our third case study to assess how effective our proposed sTrEO is at solving increasingly intractable variants of this problem. In this regard, we aim at tackling $e_{0.825}$ as the target problem by building up source task-instances with an ascending order of difficulty starting from $e_{0.1}$ through $e_{0.775}$.

\begin{table}
  \begin{center}
    \caption{Performance comparison of sTrEO against sNES, xNES, AMTEA and MAB-AMTEA for the double pole balancing problem with $l=0.825$.}\label{tabDblPl}
    \begin{tabular}{p{2.5cm}p{2.5cm}p{2.5cm}}
    \hline
    Methods & Successes & Function Evaluations \\ \hline
    sNES & $0/50$ & NA \\ 
    xNES & $0/50$ & NA \\ 
    AMTEA & $34/50$ & $3739\pm 939$ \\ 
    MAB-AMTEA & $7/50$ & $4156\pm 972$ \\ 
    sTrEO & $35/50$ & $3108\pm 993$ \\ \hline
    \end{tabular}
  \end{center}
\end{table}

\begin{figure}
  \begin{subfigure}[b]{0.49\columnwidth}
  \centering
    \includegraphics[width=\textwidth]{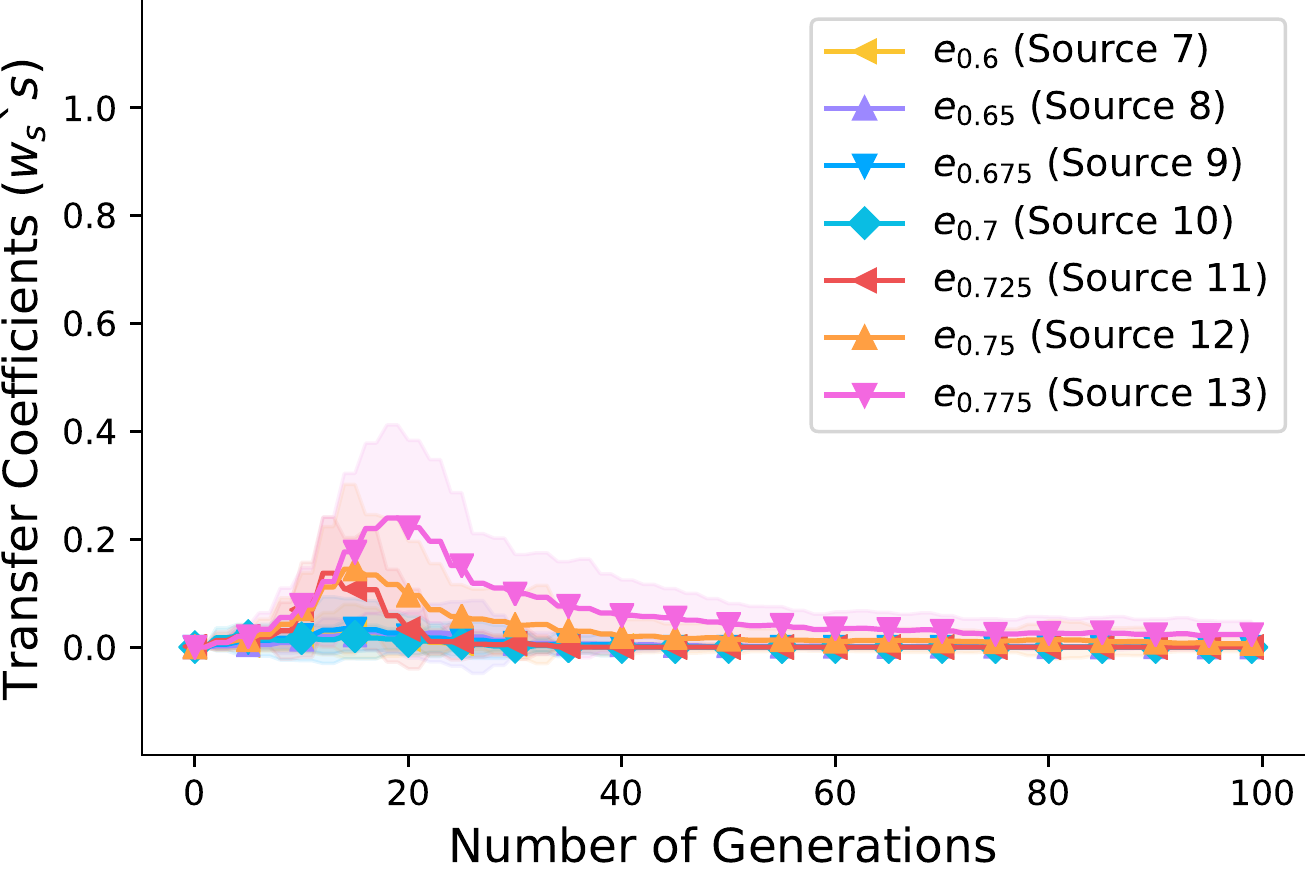}
    \caption{\scriptsize AMTEA's learned $w_{s}$'s}\label{AMTEALrnDblPole}
 \end{subfigure}
 \begin{subfigure}[b]{0.49\columnwidth}
  \centering
    \includegraphics[width=\textwidth]{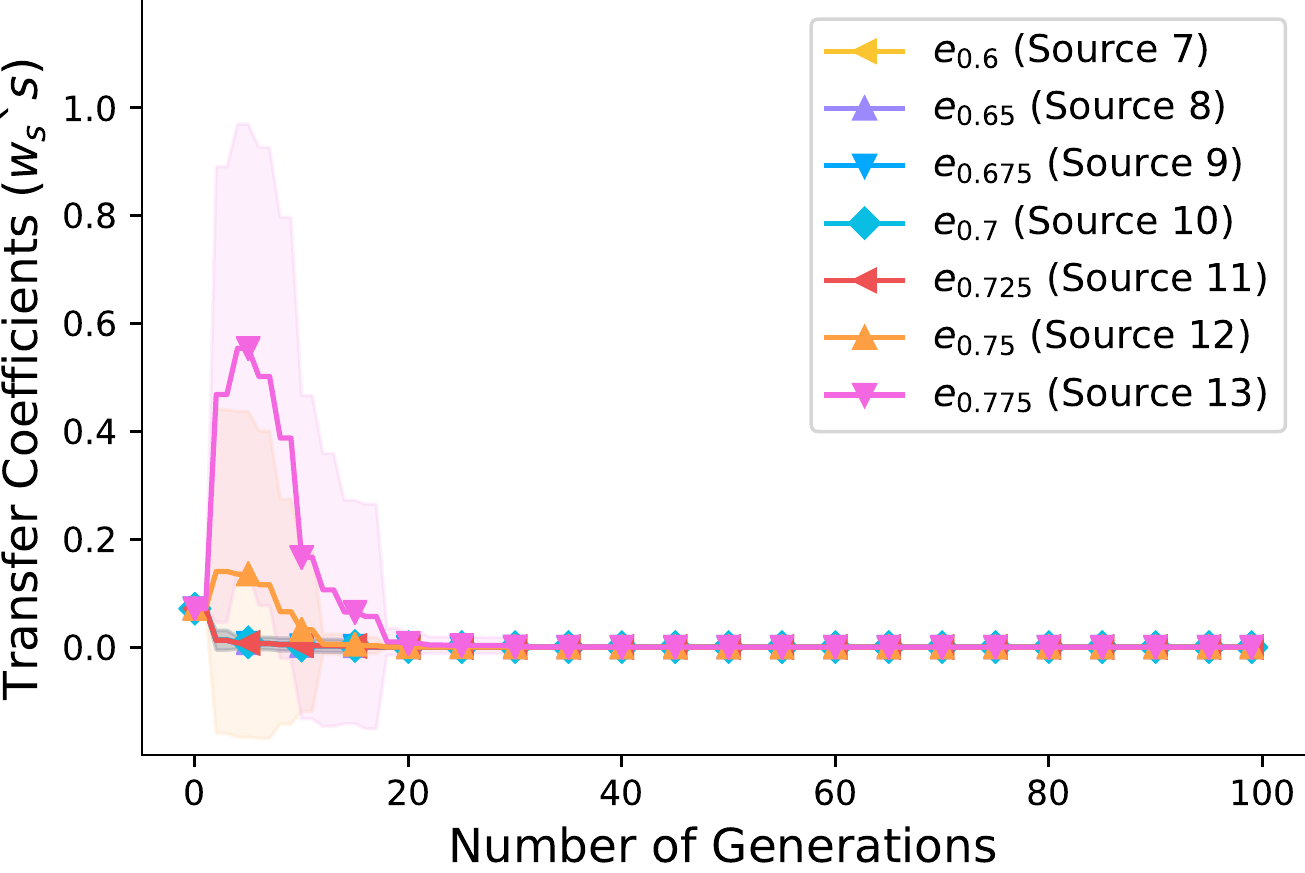}
    \caption{\scriptsize sTrEO's learned $w_{s}$'s}\label{CMTEALrnDblPole}
 \end{subfigure}
 \caption{Learned transfer coefficients of (a) AMTEA and (b) sTrEO for the double pole balancing problem with $l=0.825$. Results were averaged over successful runs. The shaded region indicates standard deviations either side of the mean.}\label{FigDblPole}
\end{figure}

To construct the optimized source probabilistic models, we first utilized a continuous CGA to solve $e_{0.1}$. Thereafter, the AMTEA was applied to solve the next instances in ascending order of $l$ by utilizing the previously solved ones. Following \cite{da2018curbing}, we adopt {\it success rate} as the performance metric for our experimental study. It is measured by counting the number of runs that each method is able to solve the problem relative to the total number of runs. We compared sTrEO with sNES, xNES, AMTEA and MAB-AMTEA by launching each for 50 independent runs. \tableautorefname \ \ref{tabDblPl} shows the success rate as well as the number of function evaluations averaged over successful runs for the compared algorithms. According to the table, neither sNES nor xNES are able to achieve success in any of their 50 independent runs, revealing the inherent difficulty of $e_{0.825}$. Conversely, both AMTEA and sTrEO can achieve significantly higher success rates but with noticeably different number of function evaluations, with that of sTrEO being fewer. The MAB-AMTEA, on the other hand, succeeded for only 7 runs, requiring a considerably higher number of evaluations than AMTEA and, in particular, sTrEO. Such notable performance of sTrEO is mainly due to online learning agility of its (1+1)-ES based source-target similarity leaning module, as showcased for the other two case studies. Lastly, we present the learned transfer coefficients of AMTEA and sTrEO averaged over their successful runs for seven mostly related sources in Figs. \ref{AMTEALrnDblPole} and \ref{CMTEALrnDblPole}, respectively. By observing the two trajectories, we can conclude that sTrEO is able to learn qualitatively similar trends of transfer coefficient values as AMTEA but with significantly lower computational cost. 


\section{Conclusion}\label{secconc}
This paper proposed a novel approach for scalable transfer evolutionary optimization, \emph{simultaneously} overcoming two critical shortcomings of today's transfer evolutionary optimization frameworks, namely (1) lack of \emph{scalability} when faced with a growing number of source task-instances and (2) lack of \emph{online learning agility} against sparsity of relevant sources to the target task. While applications of existing algorithms are limited to tens of source tasks, in this paper, we took a quantum leap forward to handle scenarios beyond 1000 source task-instances. We devised a scalable transfer evolutionary optimization framework, under the label of sTrEO, which comprises two co-evolving species for joint evolutions in the space of source knowledge and in the search space of solutions to the target problem. We were able to empirically verify the efficiency and effectiveness of sTrEO from two aspects of scalability and online learning agility across a set of practically motivated discrete and continuous optimization examples. Such notable performance arises due to,
\begin{enumerate}
    \item the scalability of the (1+1)-ES algorithm nested in sTrEO with the number of source instances, as only one individual evolves per generation and mutation is the sole reproduction operation;
    \item the novel mutation mechanism employed in the (1+1)-ES that quickly captures source-target similarities during the course of optimization. This is realized by estimating the expected fitness of source probabilistic models relative to the target objective function, enabling their exploration and exploitation;
    \item the proposed neutralization strategy in the (1+1)-ES that effectively filters out unrelated sources.
\end{enumerate}

Notably, estimating the expected fitness of each source probabilistic model relative to the target objective function enables source-target similarity matching to be carried out in an offline mode (as was done for the robotic arm example to construct the similarity heatmaps) given a unified search space and known target objective function. 

As future work, we intend to generalize sTrEO to a wider variety of real-world optimization problems in domains of manufacturing and operations research characterized by different solution representations; e.g., graph-based representations, scheduling problems, etc. In all such cases, sTrEO would be designed to be immersed in and leverage environments with as many as millions of prior optimization experiences (e.g., those derived from historical databases of interconnected, distributed micro-factories). We will also investigate how our proposed transfer optimization framework can be extended to support \emph{evolutionary multitasking} where mutual exchange of knowledge takes place among a set of tasks being simultaneously solved \cite{ong2016evolutionary,bali2020cognizant}. Notably, there is growing interest in large-scale multitasking where task numbers in the order of tens, hundreds or even thousands jointly occur \cite{huang2021towards,liang2021evolutionary}, hence, providing a stage for our scalable transfer methods to excel at addressing tractability concerns.


%





\ifCLASSOPTIONcaptionsoff
  \newpage
\fi

\bibliographystyle{IEEEtran}
\bibliography{IEEEabrv,Refs}

\begin{thebibliography}{10}
\providecommand{\url}[1]{#1}
\csname url@samestyle\endcsname
\providecommand{\newblock}{\relax}
\providecommand{\bibinfo}[2]{#2}
\providecommand{\BIBentrySTDinterwordspacing}{\spaceskip=0pt\relax}
\providecommand{\BIBentryALTinterwordstretchfactor}{4}
\providecommand{\BIBentryALTinterwordspacing}{\spaceskip=\fontdimen2\font plus
\BIBentryALTinterwordstretchfactor\fontdimen3\font minus
  \fontdimen4\font\relax}
\providecommand{\BIBforeignlanguage}[2]{{%
\expandafter\ifx\csname l@#1\endcsname\relax
\typeout{** WARNING: IEEEtran.bst: No hyphenation pattern has been}%
\typeout{** loaded for the language `#1'. Using the pattern for}%
\typeout{** the default language instead.}%
\else
\language=\csname l@#1\endcsname
\fi
#2}}
\providecommand{\BIBdecl}{\relax}
\BIBdecl

\bibitem{ong2019air}
Y.-S. Ong and A.~Gupta, ``Air 5: Five pillars of artificial intelligence
  research,'' \emph{IEEE Transactions on Emerging Topics in Computational
  Intelligence}, vol.~3, no.~5, pp. 411--415, 2019.

\bibitem{gupta2017insights}
A.~Gupta, Y.-S. Ong, and L.~Feng, ``Insights on transfer optimization: Because
  experience is the best teacher,'' \emph{IEEE Transactions on Emerging Topics
  in Computational Intelligence}, vol.~2, no.~1, pp. 51--64, 2017.

\bibitem{wong2021can}
J.~C. Wong, A.~Gupta, and Y.-S. Ong, ``Can transfer neuroevolution tractably
  solve your differential equations?'' \emph{IEEE Computational Intelligence
  Magazine}, vol.~16, no.~2, pp. 14--30, 2021.

\bibitem{tan2021evolutionary}
K.~C. Tan, L.~Feng, and M.~Jiang, ``Evolutionary transfer optimization-a new
  frontier in evolutionary computation research,'' \emph{IEEE Computational
  Intelligence Magazine}, vol.~16, no.~1, pp. 22--33, 2021.

\bibitem{gupta2018memetic}
A.~Gupta and Y.-S. Ong, \emph{Memetic Computation: The Mainspring of Knowledge
  Transfer in a Data-Driven Optimization Era}.\hskip 1em plus 0.5em minus
  0.4em\relax Springer, 2018, vol.~21.

\bibitem{min2017multiproblem}
A.~T.~W. Min, Y.-S. Ong, A.~Gupta, and C.-K. Goh, ``Multiproblem surrogates:
  Transfer evolutionary multiobjective optimization of computationally
  expensive problems,'' \emph{IEEE Transactions on Evolutionary Computation},
  vol.~23, no.~1, pp. 15--28, 2017.

\bibitem{feng2017autoencoding}
L.~Feng, Y.-S. Ong, S.~Jiang, and A.~Gupta, ``Autoencoding evolutionary search
  with learning across heterogeneous problems,'' \emph{IEEE Transactions on
  Evolutionary Computation}, vol.~21, no.~5, pp. 760--772, 2017.

\bibitem{zhang2019multi}
J.~Zhang, W.~Zhou, X.~Chen, W.~Yao, and L.~Cao, ``Multi-source selective
  transfer framework in multi-objective optimization problems,'' \emph{IEEE
  Transactions on Evolutionary Computation}, 2019.

\bibitem{feng2014memetic}
L.~Feng, Y.-S. Ong, M.-H. Lim, and I.~W. Tsang, ``Memetic search with
  interdomain learning: A realization between cvrp and carp,'' \emph{IEEE
  Transactions on Evolutionary Computation}, vol.~19, no.~5, pp. 644--658,
  2014.

\bibitem{feng2015memes}
L.~Feng, Y.-S. Ong, A.-H. Tan, and I.~W. Tsang, ``Memes as building blocks: a
  case study on evolutionary optimization+ transfer learning for routing
  problems,'' \emph{Memetic Computing}, vol.~7, no.~3, pp. 159--180, 2015.

\bibitem{ardeh2019novel}
M.~A. Ardeh, Y.~Mei, and M.~Zhang, ``A novel genetic programming algorithm with
  knowledge transfer for uncertain capacitated arc routing problem,'' in
  \emph{Pacific Rim International Conference on Artificial Intelligence}.\hskip
  1em plus 0.5em minus 0.4em\relax Springer, 2019, pp. 196--200.

\bibitem{lim2019can}
R.~Lim, Y.-S. Ong, H.~T.~H. Phan, A.~Gupta, and A.~N. Zhang, ``Can route
  planning be smarter with transfer optimization?'' in \emph{Proceedings of the
  Genetic and Evolutionary Computation Conference Companion}, 2019, pp.
  318--319.

\bibitem{jiang2017transfer}
M.~Jiang, Z.~Huang, L.~Qiu, W.~Huang, and G.~G. Yen, ``Transfer learning-based
  dynamic multiobjective optimization algorithms,'' \emph{IEEE Transactions on
  Evolutionary Computation}, vol.~22, no.~4, pp. 501--514, 2017.

\bibitem{liu2019neural}
X.-F. Liu, Z.-H. Zhan, T.-L. Gu, S.~Kwong, Z.~Lu, H.~B.-L. Duh, and J.~Zhang,
  ``Neural network-based information transfer for dynamic optimization,''
  \emph{IEEE transactions on neural networks and learning systems}, 2019.

\bibitem{jiang2020individual}
M.~Jiang, Z.~Wang, S.~Guo, X.~Gao, and K.~C. Tan, ``Individual-based transfer
  learning for dynamic multiobjective optimization,'' \emph{IEEE Transactions
  on Cybernetics}, 2020.

\bibitem{zhang2021surrogate}
F.~Zhang, Y.~Mei, S.~Nguyen, M.~Zhang, and K.~C. Tan, ``Surrogate-assisted
  evolutionary multitask genetic programming for dynamic flexible job shop
  scheduling,'' \emph{IEEE Transactions on Evolutionary Computation}, 2021.

\bibitem{zhang2021multitask}
F.~Zhang, Y.~Mei, S.~Nguyen, K.~C. Tan, and M.~Zhang, ``Multitask genetic
  programming-based generative hyperheuristics: A case study in dynamic
  scheduling,'' \emph{IEEE Transactions on Cybernetics}, 2021.

\bibitem{iqbal2012extracting}
M.~Iqbal, W.~N. Browne, and M.~Zhang, ``Extracting and using building blocks of
  knowledge in learning classifier systems,'' in \emph{Proceedings of the 14th
  annual conference on Genetic and evolutionary computation}, 2012, pp.
  863--870.

\bibitem{iqbal2013reusing}
------, ``Reusing building blocks of extracted knowledge to solve complex,
  large-scale boolean problems,'' \emph{IEEE Transactions on Evolutionary
  Computation}, vol.~18, no.~4, pp. 465--480, 2013.

\bibitem{iqbal2017cross}
M.~Iqbal, B.~Xue, H.~Al-Sahaf, and M.~Zhang, ``Cross-domain reuse of extracted
  knowledge in genetic programming for image classification,'' \emph{IEEE
  Transactions on Evolutionary Computation}, vol.~21, no.~4, pp. 569--587,
  2017.

\bibitem{min2017knowledge}
A.~T.~W. Min, R.~Sagarna, A.~Gupta, Y.-S. Ong, and C.~K. Goh, ``Knowledge
  transfer through machine learning in aircraft design,'' \emph{IEEE
  Computational Intelligence Magazine}, vol.~12, no.~4, pp. 48--60, 2017.

\bibitem{feng2020towards}
L.~Feng, Y.~Huang, I.~W. Tsang, A.~Gupta, K.~Tang, K.~C. Tan, and Y.-S. Ong,
  ``Towards faster vehicle routing by transferring knowledge from customer
  representation,'' \emph{IEEE Transactions on Intelligent Transportation
  Systems}, 2020.

\bibitem{bi2020learning}
Y.~Bi, B.~Xue, and M.~Zhang, ``Learning and sharing: A multitask genetic
  programming approach to image feature learning,'' \emph{arXiv preprint
  arXiv:2012.09444}, 2020.

\bibitem{min2020generalizing}
A.~T.~W. Min, A.~Gupta, and Y.-S. Ong, ``Generalizing transfer bayesian
  optimization to source-target heterogeneity,'' \emph{IEEE Transactions on
  Automation Science and Engineering}, 2020.

\bibitem{joy2019flexible}
T.~T. Joy, S.~Rana, S.~Gupta, and S.~Venkatesh, ``A flexible transfer learning
  framework for bayesian optimization with convergence guarantee,''
  \emph{Expert Systems with Applications}, vol. 115, pp. 656--672, 2019.

\bibitem{karimpanal2020learning}
T.~G. Karimpanal, S.~Rana, S.~Gupta, T.~Tran, and S.~Venkatesh, ``Learning
  transferable domain priors for safe exploration in reinforcement learning,''
  in \emph{2020 International Joint Conference on Neural Networks
  (IJCNN)}.\hskip 1em plus 0.5em minus 0.4em\relax IEEE, 2020, pp. 1--10.

\bibitem{louis2004learning}
S.~J. Louis and J.~McDonnell, ``Learning with case-injected genetic
  algorithms,'' \emph{IEEE Transactions on Evolutionary Computation}, vol.~8,
  no.~4, pp. 316--328, 2004.

\bibitem{cunningham1997case}
P.~Cunningham and B.~Smyth, ``Case-based reasoning in scheduling: reusing
  solution components,'' \emph{International Journal of Production Research},
  vol.~35, no.~11, pp. 2947--2962, 1997.

\bibitem{kaedi2011biasing}
M.~Kaedi and N.~Ghasem-Aghaee, ``Biasing bayesian optimization algorithm using
  case based reasoning,'' \emph{Knowledge-Based Systems}, vol.~24, no.~8, pp.
  1245--1253, 2011.

\bibitem{didi2016multi}
S.~Didi and G.~Nitschke, ``Multi-agent behavior-based policy transfer,'' in
  \emph{European Conference on the Applications of Evolutionary
  Computation}.\hskip 1em plus 0.5em minus 0.4em\relax Springer, 2016, pp.
  181--197.

\bibitem{friess2020improving}
S.~Friess, P.~Ti{\v{n}}o, S.~Menzel, B.~Sendhoff, and X.~Yao, ``Improving
  sampling in evolution strategies through mixture-based distributions built
  from past problem instances,'' in \emph{International Conference on Parallel
  Problem Solving from Nature}.\hskip 1em plus 0.5em minus 0.4em\relax
  Springer, 2020, pp. 583--596.

\bibitem{da2018curbing}
B.~Da, A.~Gupta, and Y.-S. Ong, ``Curbing negative influences online for
  seamless transfer evolutionary optimization,'' \emph{IEEE transactions on
  cybernetics}, vol.~49, no.~12, pp. 4365--4378, 2019.

\bibitem{shakeri2019coping}
M.~Shakeri, A.~Gupta, Y.-S. Ong, X.~Chi, and A.~Z. NengSheng, ``Coping with big
  data in transfer optimization,'' in \emph{2019 IEEE International Conference
  on Big Data (Big Data)}.\hskip 1em plus 0.5em minus 0.4em\relax IEEE, 2019,
  pp. 3925--3932.

\bibitem{feng2018evolutionary}
L.~Feng, L.~Zhou, J.~Zhong, A.~Gupta, Y.-S. Ong, K.-C. Tan, and A.~K. Qin,
  ``Evolutionary multitasking via explicit autoencoding,'' \emph{IEEE
  transactions on cybernetics}, vol.~49, no.~9, pp. 3457--3470, 2019.

\bibitem{xie2010breaking}
M.~Xie, L.~V. Lakshmanan, and P.~T. Wood, ``Breaking out of the box of
  recommendations: from items to packages,'' in \emph{Proceedings of the fourth
  ACM conference on Recommender systems}, 2010, pp. 151--158.

\bibitem{wierstra2014natural}
D.~Wierstra, T.~Schaul, T.~Glasmachers, Y.~Sun, J.~Peters, and J.~Schmidhuber,
  ``Natural evolution strategies,'' \emph{The Journal of Machine Learning
  Research}, vol.~15, no.~1, pp. 949--980, 2014.

\bibitem{smyth1998stacked}
P.~Smyth and D.~Wolpert, ``Stacked density estimation,'' in \emph{Advances in
  neural information processing systems}, 1998, pp. 668--674.

\bibitem{moon1996expectation}
T.~K. Moon, ``The expectation-maximization algorithm,'' \emph{IEEE Signal
  processing magazine}, vol.~13, no.~6, pp. 47--60, 1996.

\bibitem{auer1995gambling}
P.~Auer, N.~Cesa-Bianchi, Y.~Freund, and R.~E. Schapire, ``Gambling in a rigged
  casino: The adversarial multi-armed bandit problem,'' in \emph{Proceedings of
  IEEE 36th Annual Foundations of Computer Science}.\hskip 1em plus 0.5em minus
  0.4em\relax IEEE, 1995, pp. 322--331.

\bibitem{auer2002nonstochastic}
------, ``The nonstochastic multiarmed bandit problem,'' \emph{SIAM journal on
  computing}, vol.~32, no.~1, pp. 48--77, 2002.

\bibitem{muhlenbein1997equation}
H.~M{\"u}hlenbein, ``The equation for response to selection and its use for
  prediction,'' \emph{Evolutionary Computation}, vol.~5, no.~3, pp. 303--346,
  1997.

\bibitem{deb1995simulated}
K.~Deb, R.~B. Agrawal \emph{et~al.}, ``Simulated binary crossover for
  continuous search space,'' \emph{Complex systems}, vol.~9, no.~2, pp.
  115--148, 1995.

\bibitem{deb2014analysing}
K.~Deb and D.~Deb, ``Analysing mutation schemes for real-parameter genetic
  algorithms,'' \emph{International Journal of Artificial Intelligence and Soft
  Computing}, vol.~4, no.~1, pp. 1--28, 2014.

\bibitem{do2008multivariate}
C.~B. Do, ``The multivariate gaussian distribution,'' 2008.

\bibitem{pybrain2010jmlr}
T.~Schaul, J.~Bayer, D.~Wierstra, Y.~Sun, M.~Felder, F.~Sehnke,
  T.~R{\"u}ckstie{\ss}, and J.~Schmidhuber, ``{PyBrain},'' \emph{Journal of
  Machine Learning Research}, vol.~11, pp. 743--746, 2010.

\bibitem{michalewicz1994genetic}
Z.~Michalewicz and J.~Arabas, ``Genetic algorithms for the 0/1 knapsack
  problem,'' in \emph{International Symposium on Methodologies for Intelligent
  Systems}.\hskip 1em plus 0.5em minus 0.4em\relax Springer, 1994, pp.
  134--143.

\bibitem{mouret2020quality}
J.-B. Mouret and G.~Maguire, ``Quality diversity for multi-task optimization,''
  \emph{arXiv preprint arXiv:2003.04407}, 2020.

\bibitem{gomez2008accelerated}
F.~Gomez, J.~Schmidhuber, and R.~Miikkulainen, ``Accelerated neural evolution
  through cooperatively coevolved synapses,'' \emph{Journal of Machine Learning
  Research}, vol.~9, no. May, pp. 937--965, 2008.

\bibitem{gomez1999solving}
F.~J. Gomez and R.~Miikkulainen, ``Solving non-markovian control tasks with
  neuroevolution,'' in \emph{IJCAI}, vol.~99, 1999, pp. 1356--1361.

\bibitem{ong2016evolutionary}
Y.-S. Ong and A.~Gupta, ``Evolutionary multitasking: a computer science view of
  cognitive multitasking,'' \emph{Cognitive Computation}, vol.~8, no.~2, pp.
  125--142, 2016.

\bibitem{bali2020cognizant}
K.~K. Bali, A.~Gupta, Y.-S. Ong, and P.~S. Tan, ``Cognizant multitasking in
  multiobjective multifactorial evolution: Mo-mfea-ii,'' \emph{IEEE
  Transactions on Cybernetics}, 2020.

\bibitem{huang2021towards}
Y.~Huang, L.~Feng, A.~Qin, M.~Chen, and K.~C. Tan, ``Towards large-scale
  evolutionary multi-tasking: A gpu-based paradigm,'' \emph{IEEE Transactions
  on Evolutionary Computation}, 2021.

\bibitem{liang2021evolutionary}
Z.~Liang, X.~Xu, L.~Liu, Y.~Tu, and Z.~Zhu, ``Evolutionary many-task
  optimization based on multi-source knowledge transfer,'' \emph{IEEE
  Transactions on Evolutionary Computation}, 2021.

\end{thebibliography}

\vspace{-2.4cm}

\begin{IEEEbiography}[{\raisebox{0.7cm}{\includegraphics[width=1in,height=1.25in,clip,keepaspectratio]{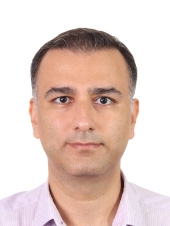}}}]{Mojtaba Shakeri}
  received the PhD degree in Computer Engineering from the Nanyang Technological University, Singapore in 2013. He has been a Scientist in the Singapore Institute of Manufacturing Technology, a research institute in Singapore’s Agency for Science, Technology and Research (A*STAR) from 2019 to 2022. He was also an assistant professor in the Islamic Azad University, Qazvin Branch and later in the University of Guilan, Iran, engaged with teaching AI-related courses for both bachelor and master students for nearly six years. He has supervised over 50 master dissertations in both universities. His core competencies are in combinatorial optimization with applications for supply chains and operations research mainly by developing nature-inspired evolutionary algorithms and mathematical models.
\end{IEEEbiography}

\vspace{-2.2cm}

\begin{IEEEbiography}[{\raisebox{0.7cm}{\includegraphics[width=1in,height=1.25in,clip,keepaspectratio]{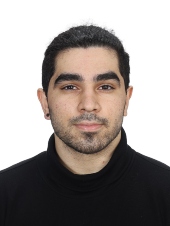}}}]{Erfan Miahi} is an MSc student at the University of Alberta and the Reinforcement Learning and Artificial Intelligence lab. He received the Bachelor's degree in Computer Engineering from the University of Guilan 
in 2020. Erfan has diverse research experience in the field of artificial intelligence, ranging from developing deep learning algorithms for medical image analysis to fundamental research in reinforcement learning and evolutionary optimization. His research areas of interest are reinforcement learning and representation learning. 
\end{IEEEbiography}

\vspace{-2.2cm}

\begin{IEEEbiography}[{\raisebox{0.7cm}{\includegraphics[width=1in,height=1.25in,clip,keepaspectratio]{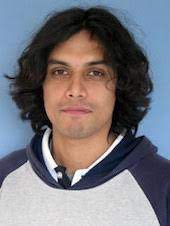}}}]{Abhishek Gupta}
  received the PhD degree in Engineering Science from the University of Auckland, New Zealand, in 2014. He is currently a Scientist and Technical Lead in the Singapore Institute of Manufacturing Technology, a research institute in Singapore’s Agency for Science, Technology and Research (A*STAR). Abhishek has diverse research experience in computational science, ranging from mathematical and numerical modelling in engineering to topics in computational intelligence. He is an editorial board member of the IEEE Transactions on Emerging Topics in Computational Intelligence, Complex \& Intelligent Systems journal, the Memetic Computing journal, and the Springer book series on Adaptation, Learning, and Optimization.
\end{IEEEbiography}

\vspace{-2.2cm}

\begin{IEEEbiography}[{\raisebox{0.7cm}{\includegraphics[width=1in,height=1.25in,clip,keepaspectratio]{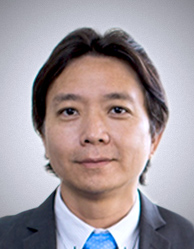}}}]{Yew-Soon Ong}
  (M’99-SM’12-F’18) received the Ph.D. degree in artificial intelligence in complex design from the University of Southampton, U.K., in 2003. He is President’s Chair Professor in Computer Science at Nanyang Technological University (NTU) and Chief Artificial Intelligence Scientist of the Agency for Science, Technology and Research Singapore. At NTU, he serves as co-Director of the Singtel-NTU Cognitive \& Artificial Intelligence Joint Lab. His research interest is in artificial and computational intelligence. He is founding Editor-in-Chief of the IEEE Transactions on Emerging Topics in Computational Intelligence and associate editor of IEEE Transactions on Neural Networks \& Learning Systems, IEEE Transactions on Cybernetics, IEEE Transactions on Artificial Intelligence.
\end{IEEEbiography}

\clearpage

\twocolumn[
    \begin{center}
      {\LARGE\sffamily Scalable Transfer Evolutionary Optimization: Coping with Big task-instances - Supplemental Material}\\
       \vspace{2ex}
   By \text{Mojtaba Shakeri, Erfan Miahi, Abhishek Gupta and Yew-Soon Ong}
    \end{center}
]

\setcounter{equation}{0}
\setcounter{figure}{0}
\setcounter{table}{0}
\setcounter{page}{1}
\setcounter{section}{0}
\makeatletter

\renewcommand{\thesection}{S.\Roman{section}}
\renewcommand{\theequation}{S.\arabic{equation}}
\renewcommand{\thetable}{S.\arabic{table}}
\renewcommand{\thefigure}{S.\arabic{figure}}

\section{Hyper-Parameter Sensitivity Analysis}\label{SenAnal}
\begin{figure}
 \begin{subfigure}[b]{0.49\columnwidth}
  \centering
    \includegraphics[width=\textwidth]{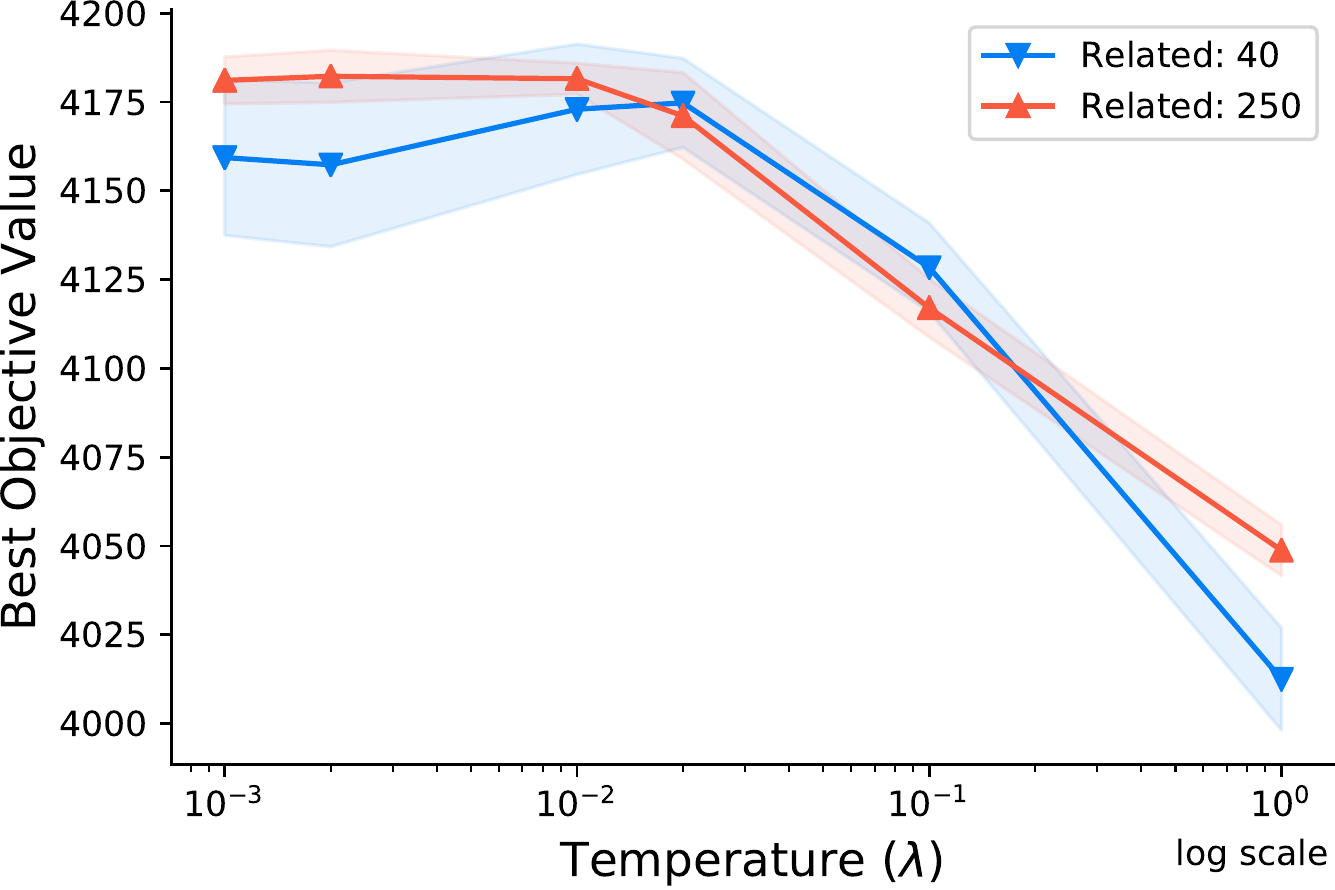}
    \caption{\scriptsize Knapsack}\label{TempSenAnalTimeKP}
 \end{subfigure}
 \begin{subfigure}[b]{0.49\columnwidth}
  \centering
    \includegraphics[width=\textwidth]{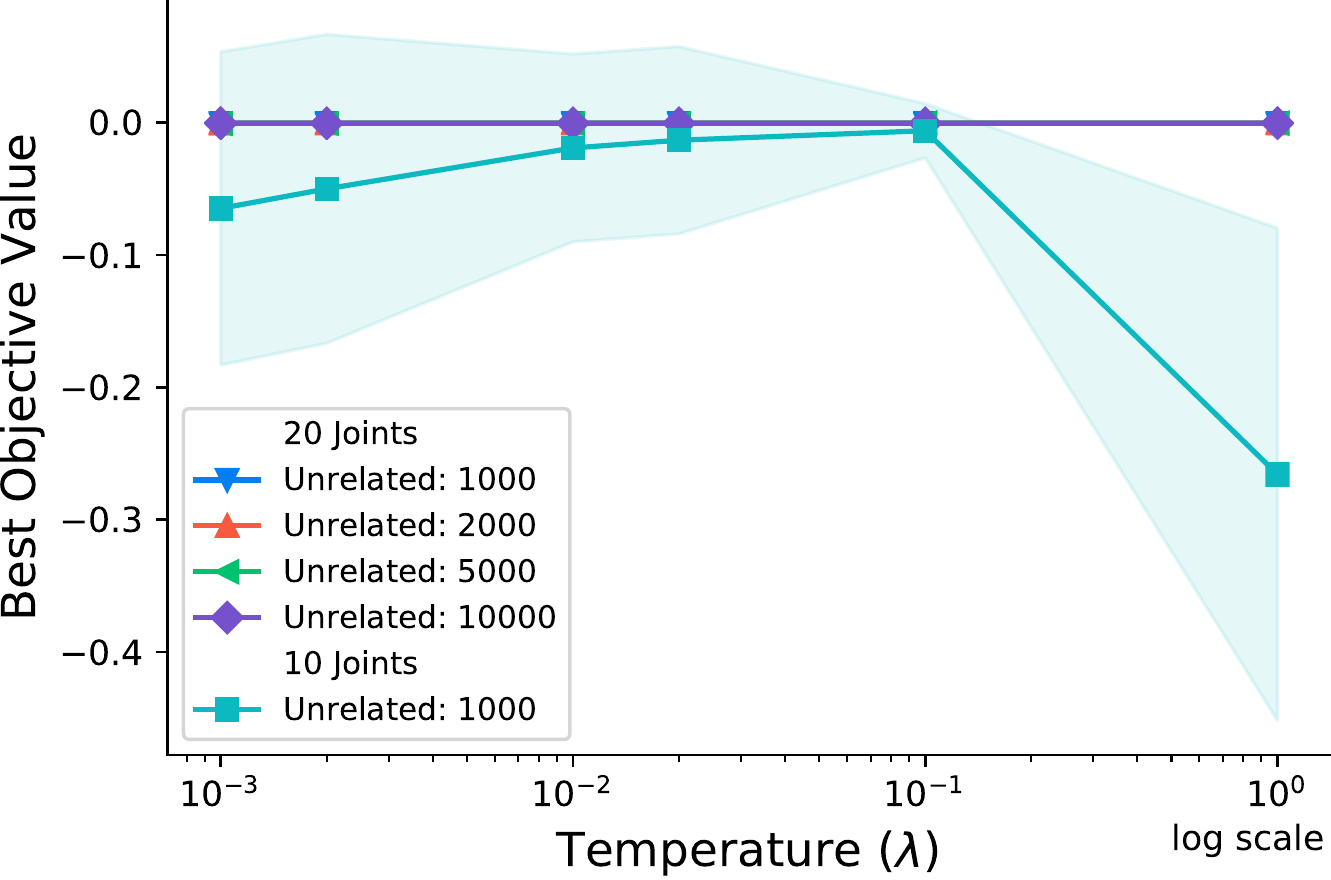}
    \caption{\scriptsize Robotic arm}\label{TempSenAnalTimeRA}
 \end{subfigure}
 \caption{Sensitivity analysis of \emph{temperature} ($\lambda$) for (a) two variations of the 0/1 knapsack problem (1000 sources with 250 and 40 related sources) and (b) five variations of the robotic arm example (10 joints with 1000 sources and 15 related sources, 20 joints with 1000, 2000, 5000 and 10000 sources and 15, 30, 75, 150 related sources, respectively). The shaded region indicates standard deviations either side of the mean.}\label{FigTempSenAnal}
\end{figure}

\begin{figure}
 \begin{subfigure}[b]{0.49\columnwidth}
  \centering
    \includegraphics[width=\textwidth]{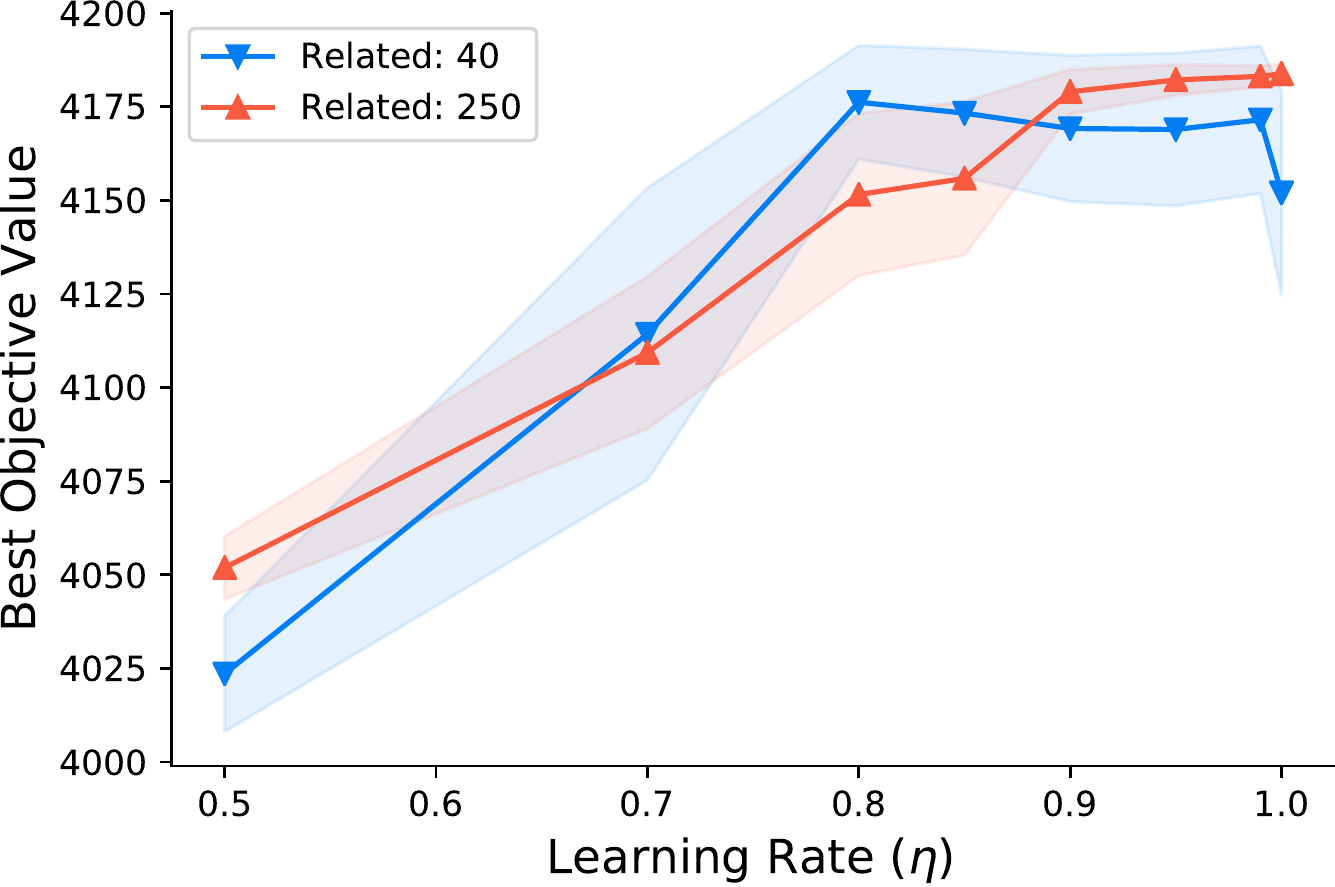}
    \caption{\scriptsize Knapsack}\label{LearnSenAnalTimeKP}
 \end{subfigure}
 \begin{subfigure}[b]{0.49\columnwidth}
  \centering
    \includegraphics[width=\textwidth]{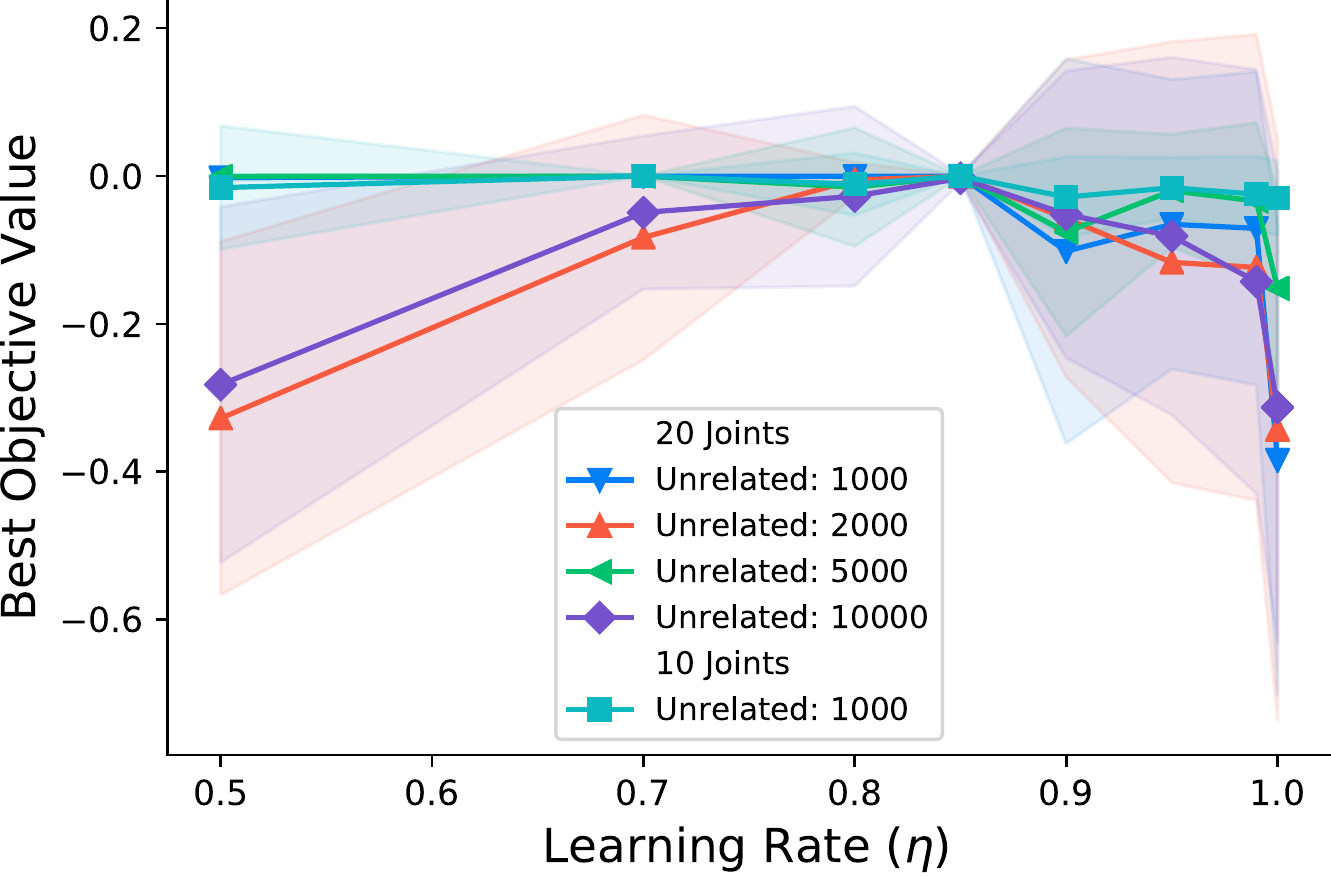}
    \caption{\scriptsize Robotic arm}\label{LearnSenAnalTimeRA}
 \end{subfigure}
 \caption{Sensitivity analysis of \emph{learning rate} ($\eta$) for (a) two variations of the 0/1 knapsack problem (1000 sources with 250 and 40 related sources) and (b) five variations of the robotic arm example (10 joints with 1000 sources and 15 related sources, 20 joints with 1000, 2000, 5000 and 10000 sources and 15, 30, 75, 150 related sources, respectively). The shaded region indicates standard deviations either side of the mean.}\label{FigLearnSenAnal}
\end{figure}

\begin{figure}
 \begin{subfigure}[b]{0.49\columnwidth}
  \centering
    \includegraphics[width=\textwidth]{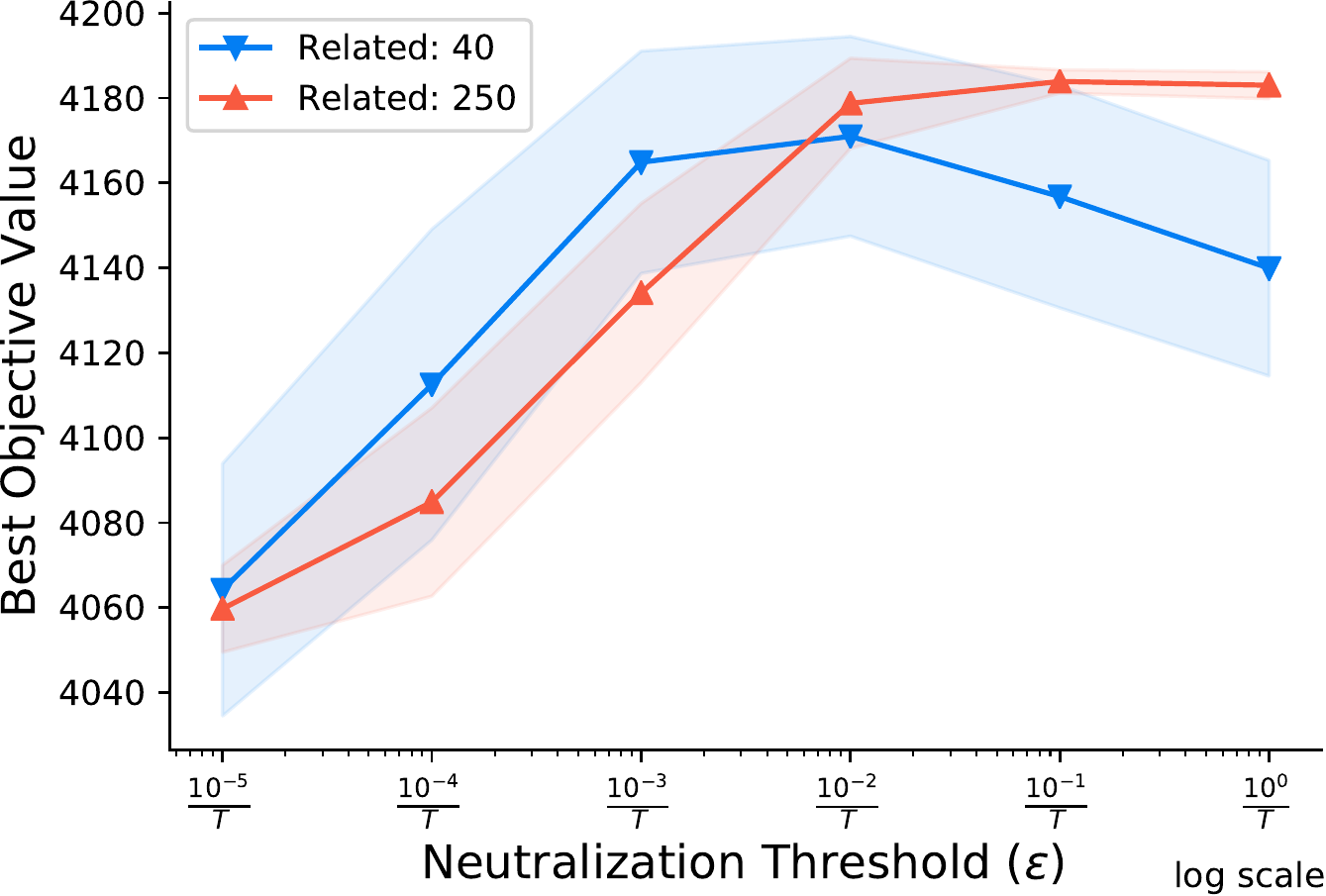}
    \caption{\scriptsize Knapsack}\label{NeutralSenAnalTimeKP}
 \end{subfigure}
 \begin{subfigure}[b]{0.49\columnwidth}
  \centering
    \includegraphics[width=\textwidth]{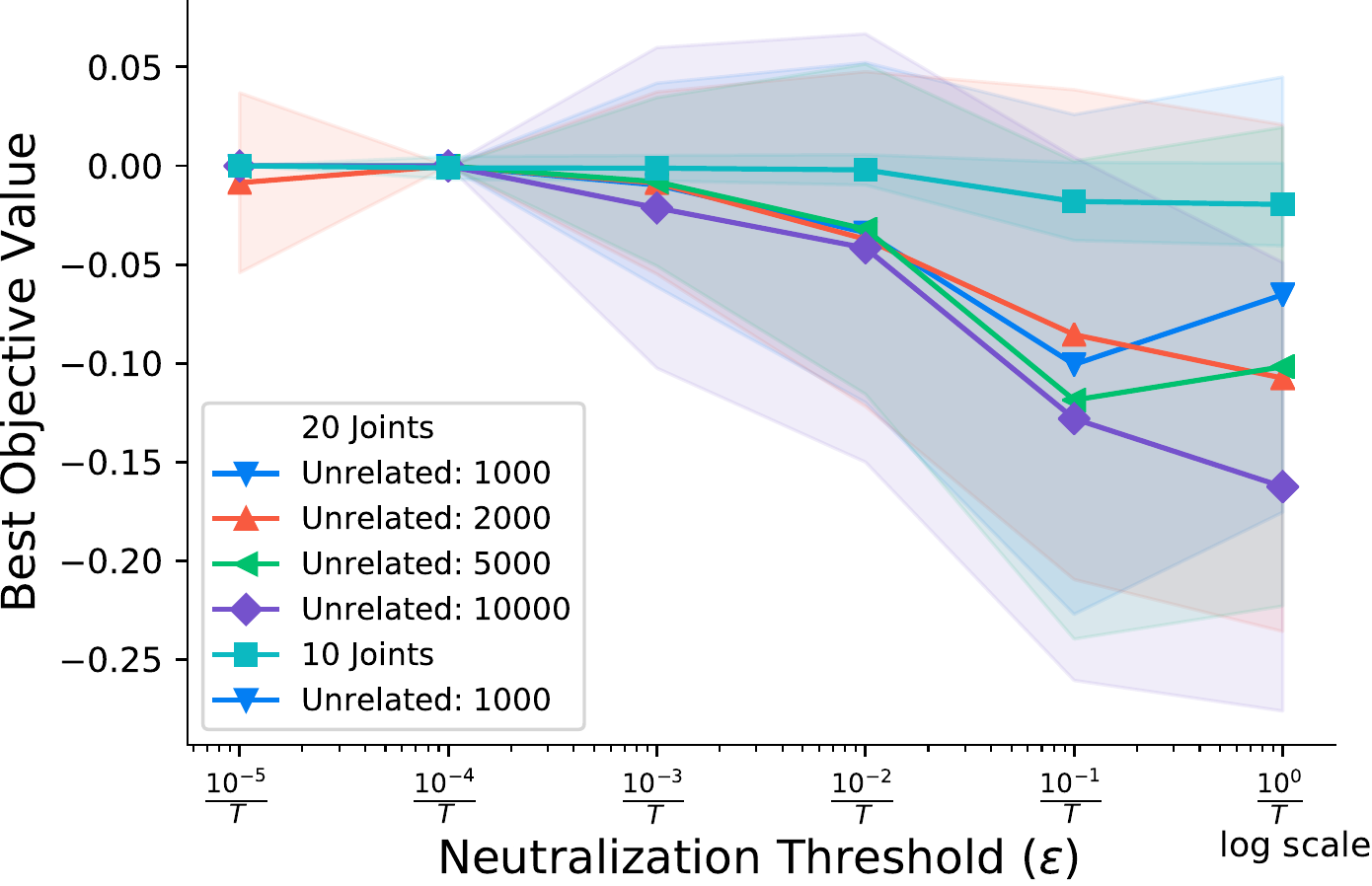}
    \caption{\scriptsize Robotic arm}\label{NeutralSenAnalTimeRA}
 \end{subfigure}
 \caption{Sensitivity analysis of \emph{neutralization threshold} ($\varepsilon$) for (a) two variations of the 0/1 knapsack problem (1000 sources with 250 and 40 related sources) and (b) five variations of the robotic arm example (10 joints with 1000 sources and 15 related sources, 20 joints with 1000, 2000, 5000 and 10000 sources and 15, 30, 75, 150 related sources, respectively). The shaded region indicates standard deviations either side of the mean.}\label{FigNeutralSenAnal}
\end{figure}

\begin{figure}
 \begin{subfigure}[b]{0.49\columnwidth}
  \centering
    \includegraphics[width=\textwidth]{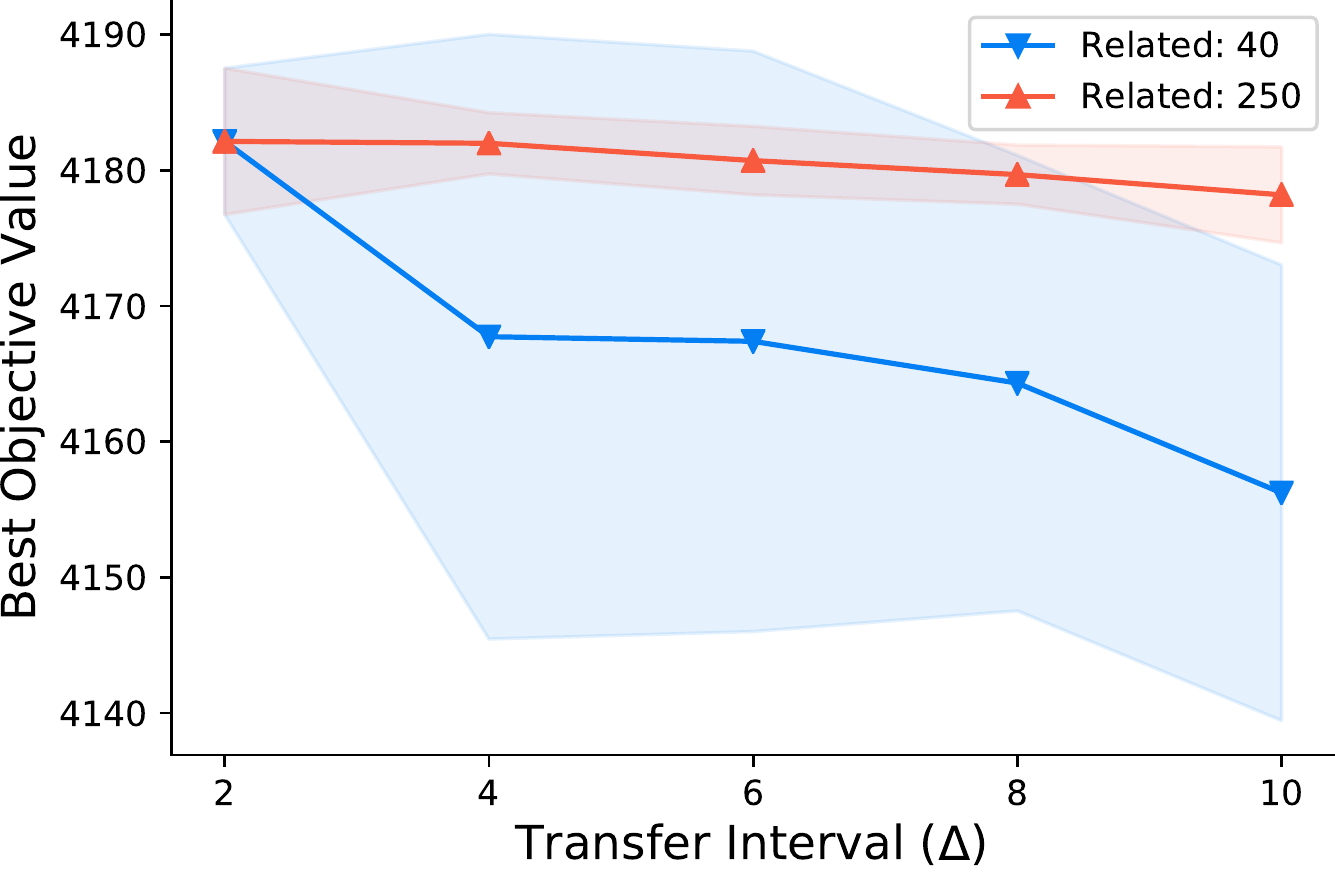}
    \caption{\scriptsize Knapsack}\label{TransferSenAnalTimeKP}
 \end{subfigure}
 \begin{subfigure}[b]{0.49\columnwidth}
  \centering
    \includegraphics[width=\textwidth]{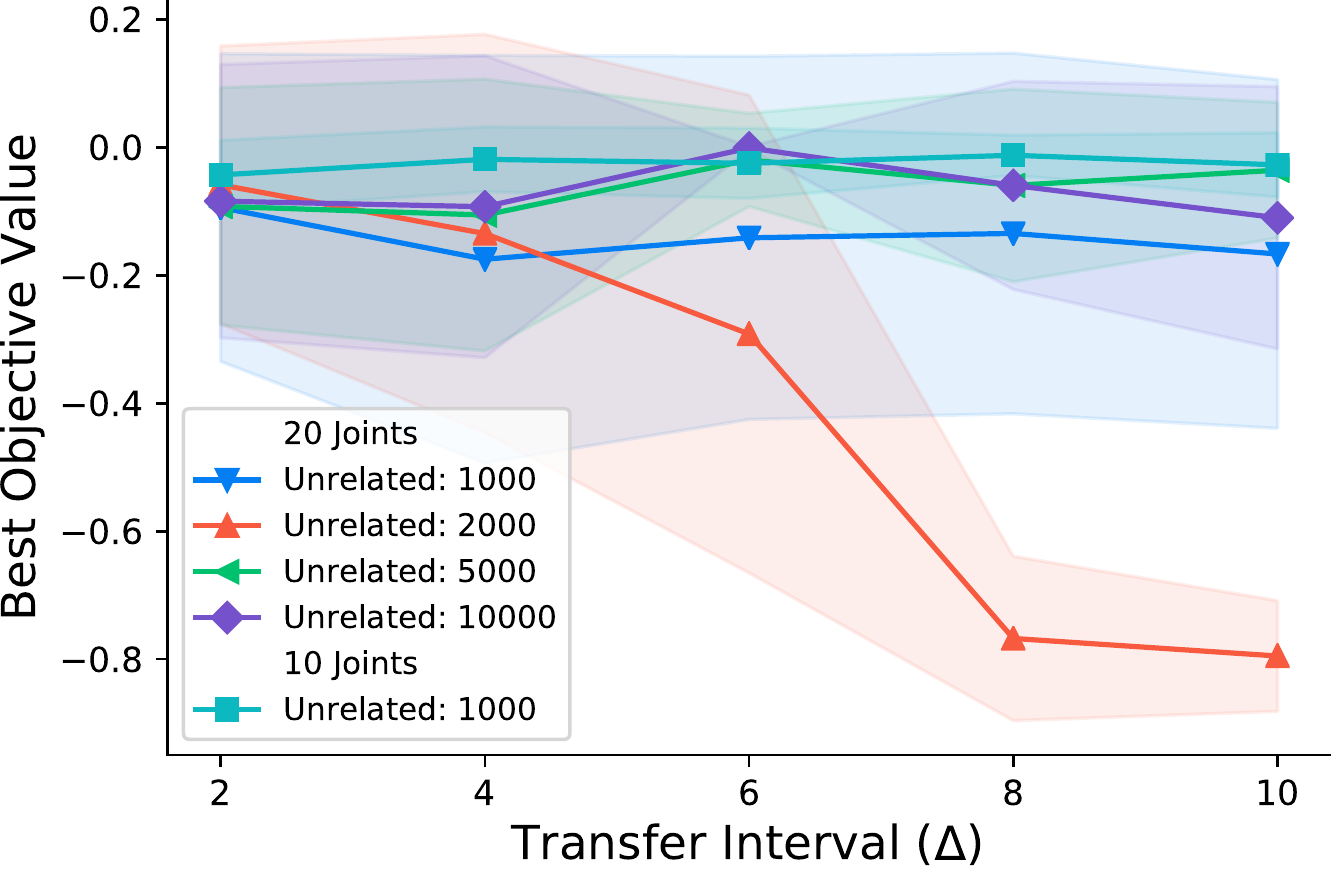}
    \caption{\scriptsize Robotic arm}\label{TransferSenAnalTimeRA}
 \end{subfigure}
 \caption{Sensitivity analysis of \emph{transfer interval} ($\Delta$) for (a) two variations of the 0/1 knapsack problem (1000 sources with 250 and 40 related sources) and (b) five variations of the robotic arm example (10 joints with 1000 sources and 15 related sources, 20 joints with 1000, 2000, 5000 and 10000 sources and 15, 30, 75, 150 related sources, respectively). The shaded region indicates standard deviations either side of the mean.}\label{FigTransferSenAnal}
\end{figure}

We carried out the sensitivity analysis for the four hyper-parameters of the sTrEO. Each hyper-parameter is treated individually by fixing the value of the others to the setting chosen for the experiments in the paper. \tableautorefname \ \ref{tabSenAnal} shows different settings used for each hyper-parameter. Note that the neutralization threshold $\varepsilon$ is a function of $T$ which is the total number of probabilistic models including the target model. Figs. \ref{FigTempSenAnal} to \ref{FigTransferSenAnal} shows the best objective value of sTrEO (averaged over 30 independent runs) relative to changes in the setting of each hyper-parameter. Two variations of the 0/1 knapsack problem (1000 sources with 250 and 40 related sources) as well as five variations of the robotic arm example (10 joints with 1000 sources and 15 related sources, 20 joints with 1000, 2000, 5000 and 10000 sources with 15, 30, 75, 150 related sources, respectively) were considered for the analysis. 

\begin{table}[b]
  \begin{center}
    \caption{Settings for the sensitivity analysis. The setting in bold was chosen for the experiments in the paper.}\label{tabSenAnal}
    \begin{tabular}{p{3.8cm}p{4.2cm}}
    \hline
    Hyper-parameter & Settings \\ \hline
    Temperature $(\lambda)$ & $0.001,0.002, \mathbf{0.01},0.02,0.1,1$ \\ 
    Learning rate $(\eta)$ & $0.5,0.7,0.8,0.85,\mathbf{0.9},0.95,0.99,1$ \\ 
    Neutralization threshold $(\varepsilon)$ & $\frac{10^{-5}}{T},\frac{10^{-4}}{T},\frac{10^{-3}}{T},\frac{\mathbf{10^{-2}}}{\boldsymbol{T}},\frac{10^{-1}}{T},\frac{1}{T}$ \\ 
    Transfer interval $(\Delta)$ & $\mathbf{2},4,6,8,10$ \\ \hline
    \end{tabular}
  \end{center}
\end{table}

The impact of \emph{temperature} ($\lambda$) on the performance of sTrEO is shown in Fig. \ref{FigTempSenAnal} for both knapsack and robotic arm examples. Given a fixed time budget (which is sTrEO's convergence time following the setting in \tableautorefname \ \ref{tabESSet}), Figs. \ref{TempSenAnalTimeKP} and \ref{TempSenAnalTimeRA} show the best objective value achieved for different settings of $\lambda$ for knapsack and robotic arm examples, respectively. As can be seen, setting the hyper-parameter in $[0.001, 0.02]$ yields superior results for both variations of the knapsack problem. For the robotic arm example, however, the range increases to $[0.001, 0.1]$. Based on the above analysis, the appropriate setting of $\lambda$ should lie in $[0.001, 0.02]$, which we set $\lambda=0.01$ in the experiments.


The impact of \emph{learning rate} ($\eta$) on the performance of sTrEO is shown in Fig. \ref{FigLearnSenAnal} for both knapsack and robotic arm examples. As can be seen from Fig. \ref{LearnSenAnalTimeKP}, setting the hyper-parameter in $[0.9, 0.99]$ yields superior results for both variations of the knapsack problem. For the robotic arm example, however, the range is in $[0.8, 0.9]$ according to Fig. \ref{LearnSenAnalTimeRA}. Based on the above analysis, we set $\eta=0.9$ in the experiments. 

The impact of \emph{neutralization threshold} ($\varepsilon$) on the performance of sTrEO is shown in Fig. \ref{FigNeutralSenAnal} for both knapsack and robotic arm examples. As can be seen from Fig. \ref{NeutralSenAnalTimeKP} setting the hyper-parameter around $\frac{10^{-2}}{T}$ yields superior results for both variations of the knapsack problem. This, however, is broader for the robotic arm example in $[\frac{10^{-5}}{T}, \frac{10^{-2}}{T}]$ following Fig. \ref{NeutralSenAnalTimeRA}. Based on the above analysis, we set $\varepsilon=\frac{10^{-2}}{T}$ in the experiments.


Lastly, the impact of \emph{transfer interval} ($\Delta$) on the performance of sTrEO is shown in Fig. \ref{FigTransferSenAnal} for both knapsack and robotic arm examples. As can be seen from Fig. \ref{TransferSenAnalTimeKP}, setting $\Delta=2$ yields superior results for both variations of the knapsack problem. This, however, increases to $[2, 4]$ for the robotic arm example following Fig. \ref{TransferSenAnalTimeRA}. Based on the above analysis, we set $\Delta=2$ in the experiments.


\section{Further Experiments}\label{FurtherExp}
We carried out further experiments to assess the performance of sTrEO beyond 1000 source task-instances by keeping the number of related sources fixed and increasing the number of unrelated sources. To this end, we set up an evaluation scenario on the more complex 20-joint robotic arm example by starting with 150 related sources and gradually increasing the number of unrelated sources from 850 to 9850. That is, the experiment starts with 1000 sources of which 150 instances are related. We then increase the total number of sources to 2000, 5000 and 10000 while keeping the number of related source instances to 150, thereby reaching the same source-target relatedness ratio of $0.015$ that we considered in Section \ref{SecKinArm}. We used the same setting as in \tableautorefname \ \ref{tabESSet} in the experiments. The assessment is from two aspects of \emph{online learning agility} and \emph{scalability} by comparing with the two existing TrEO methods, AMTEA and MAB-AMTEA.

\begin{figure}
  \begin{subfigure}[b]{0.49\columnwidth}
  \centering
    \includegraphics[width=\textwidth]{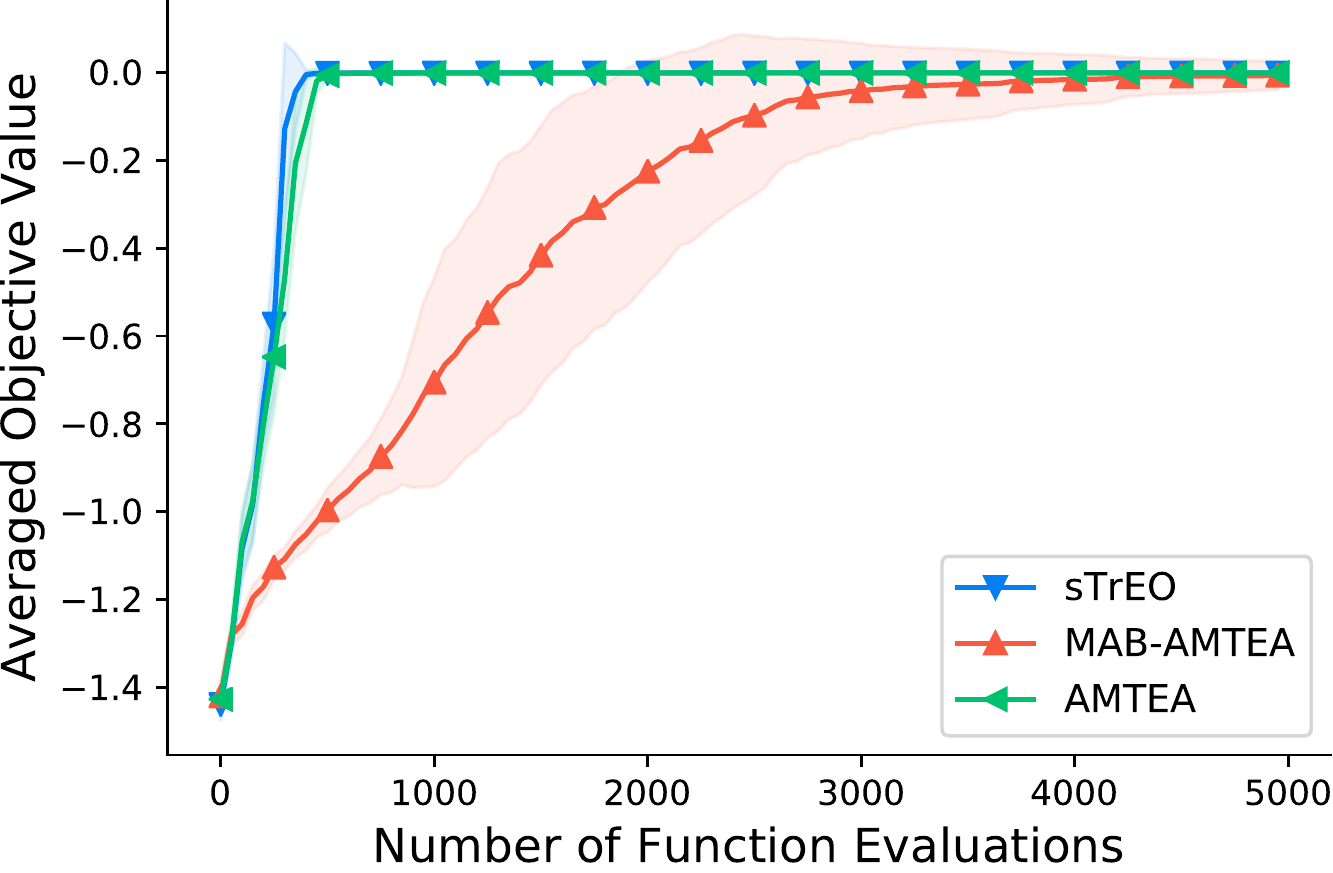}
    \caption{\scriptsize 1000 sources (150 related)}\label{GenRA1000_15}
 \end{subfigure}
 \begin{subfigure}[b]{0.49\columnwidth}
  \centering
    \includegraphics[width=\textwidth]{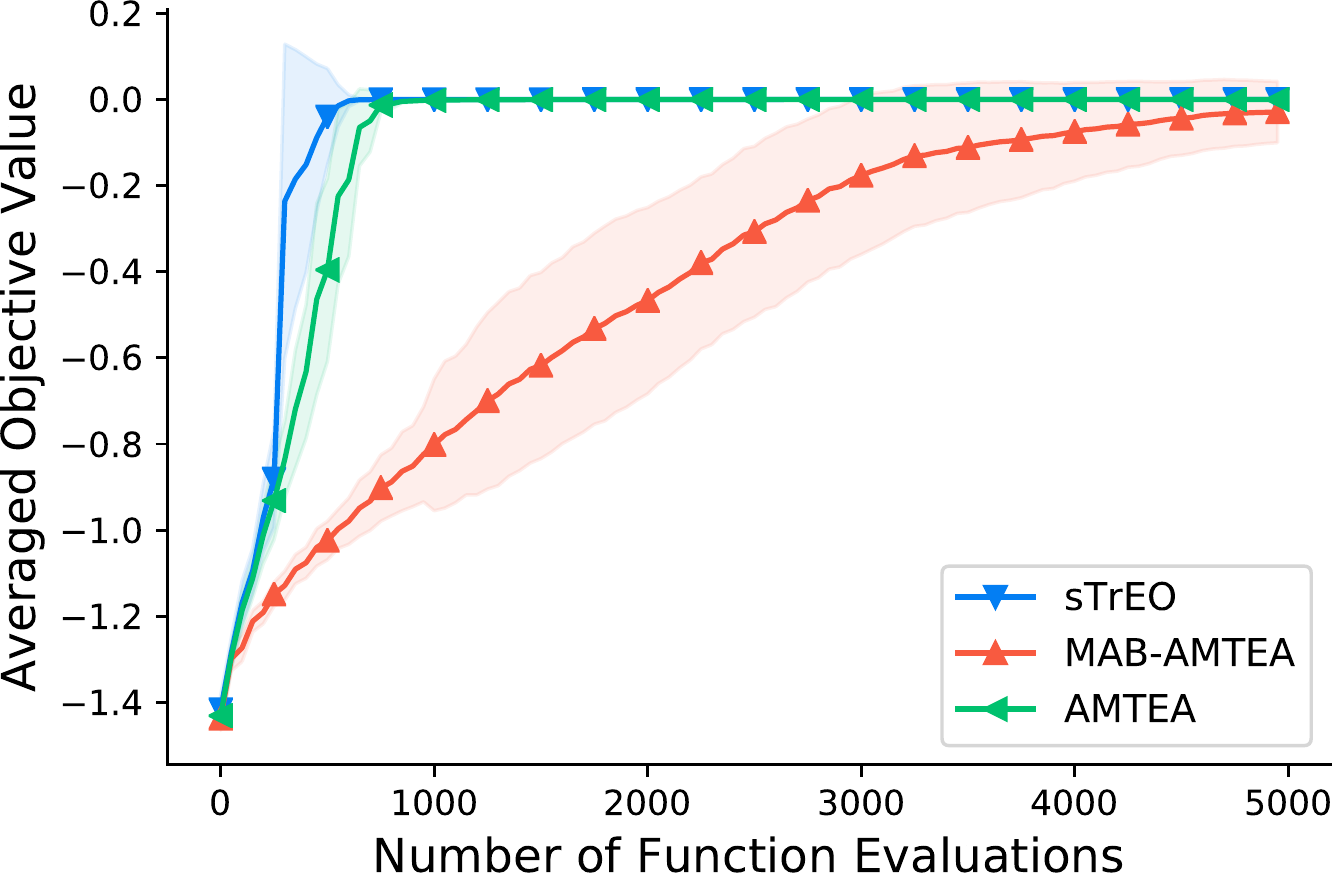}
    \caption{\scriptsize 2000 sources (150 related)}\label{GenRA2000_30}
 \end{subfigure}
   \begin{subfigure}[b]{0.49\columnwidth}
  \centering
    \includegraphics[width=\textwidth]{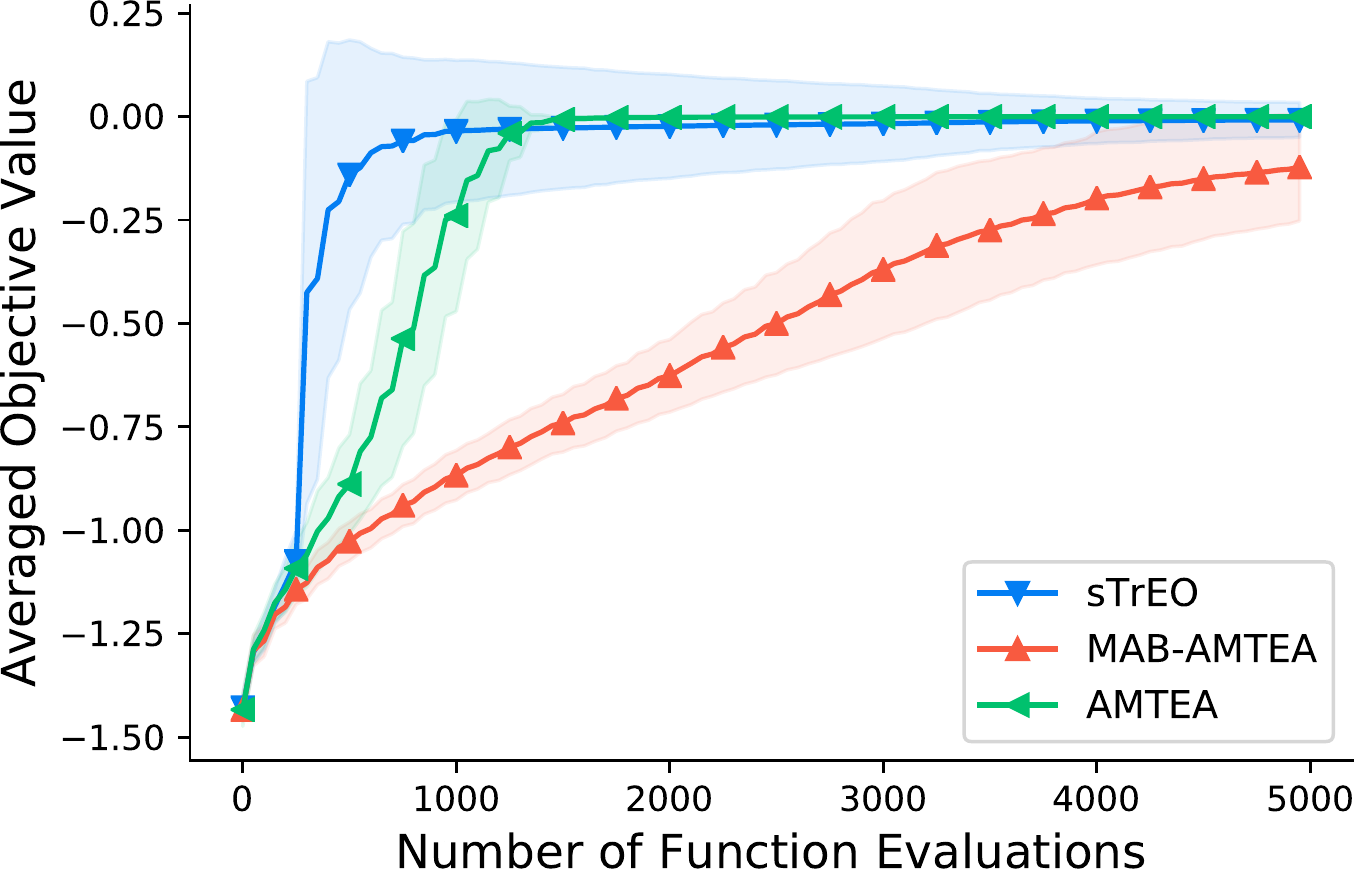}
    \caption{\scriptsize 5000 sources (150 related)}\label{GenRA5000_75}
 \end{subfigure}
 \begin{subfigure}[b]{0.49\columnwidth}
  \centering
    \includegraphics[width=\textwidth]{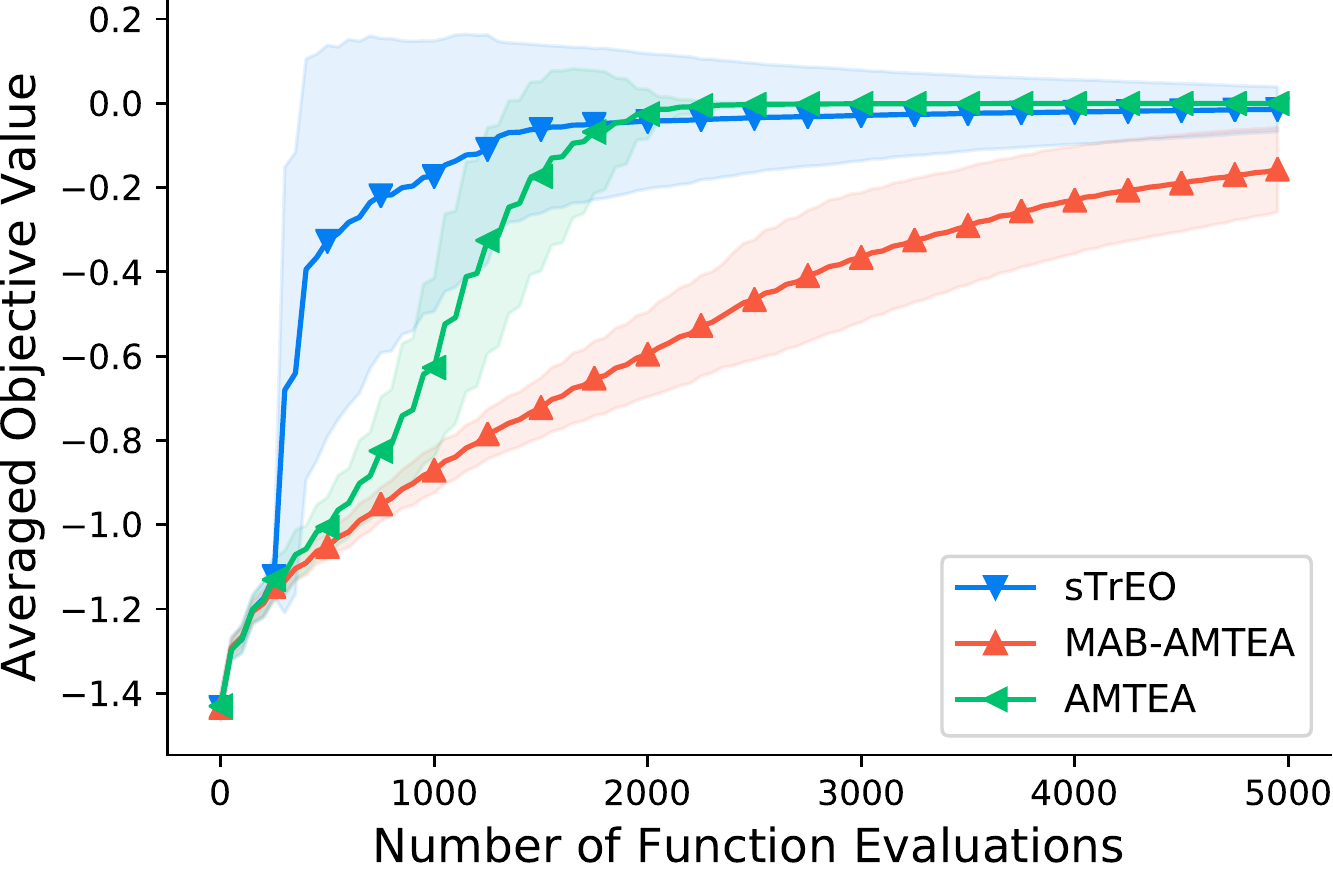}
    \caption{\scriptsize 10000 sources (150 related)}\label{GenRA10000_150}
 \end{subfigure}
 \caption{Further experiments on \emph{online learning agility} of sTrEO for the 20-joint robotic arm example with (a) 1000, (b) 2000, (c) 5000 and (d) 10000 sources. The number of related sources is fixed to 150. The shaded region indicates standard deviations either side of the mean.}\label{FigOlnLrnAgl}
\end{figure}

\begin{figure}
  \begin{subfigure}[b]{0.49\columnwidth}
  \centering
    \includegraphics[width=\textwidth]{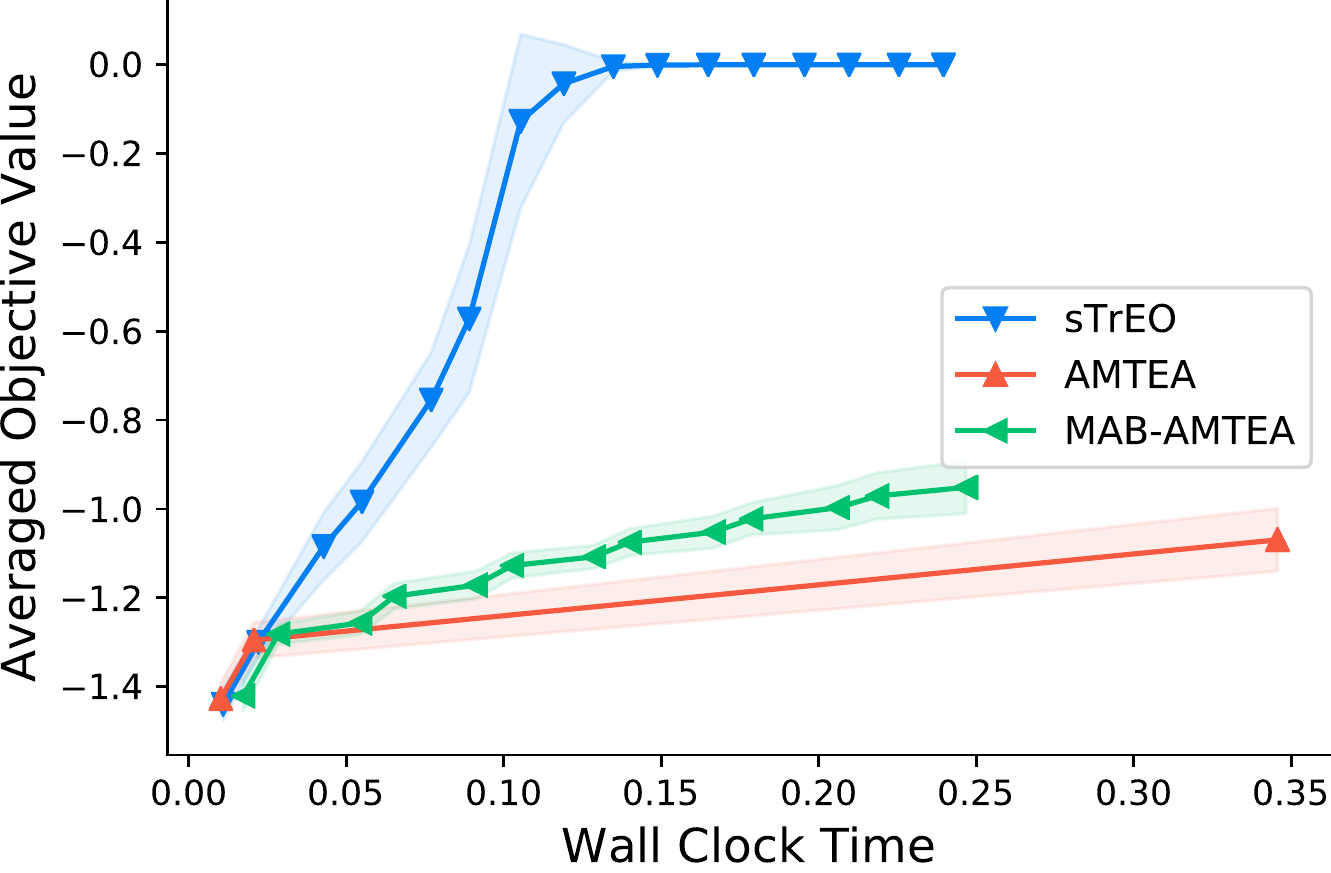}
    \caption{\scriptsize 1000 sources (150 related)}\label{TimeRA1000_15}
 \end{subfigure}
 \begin{subfigure}[b]{0.49\columnwidth}
  \centering
    \includegraphics[width=\textwidth]{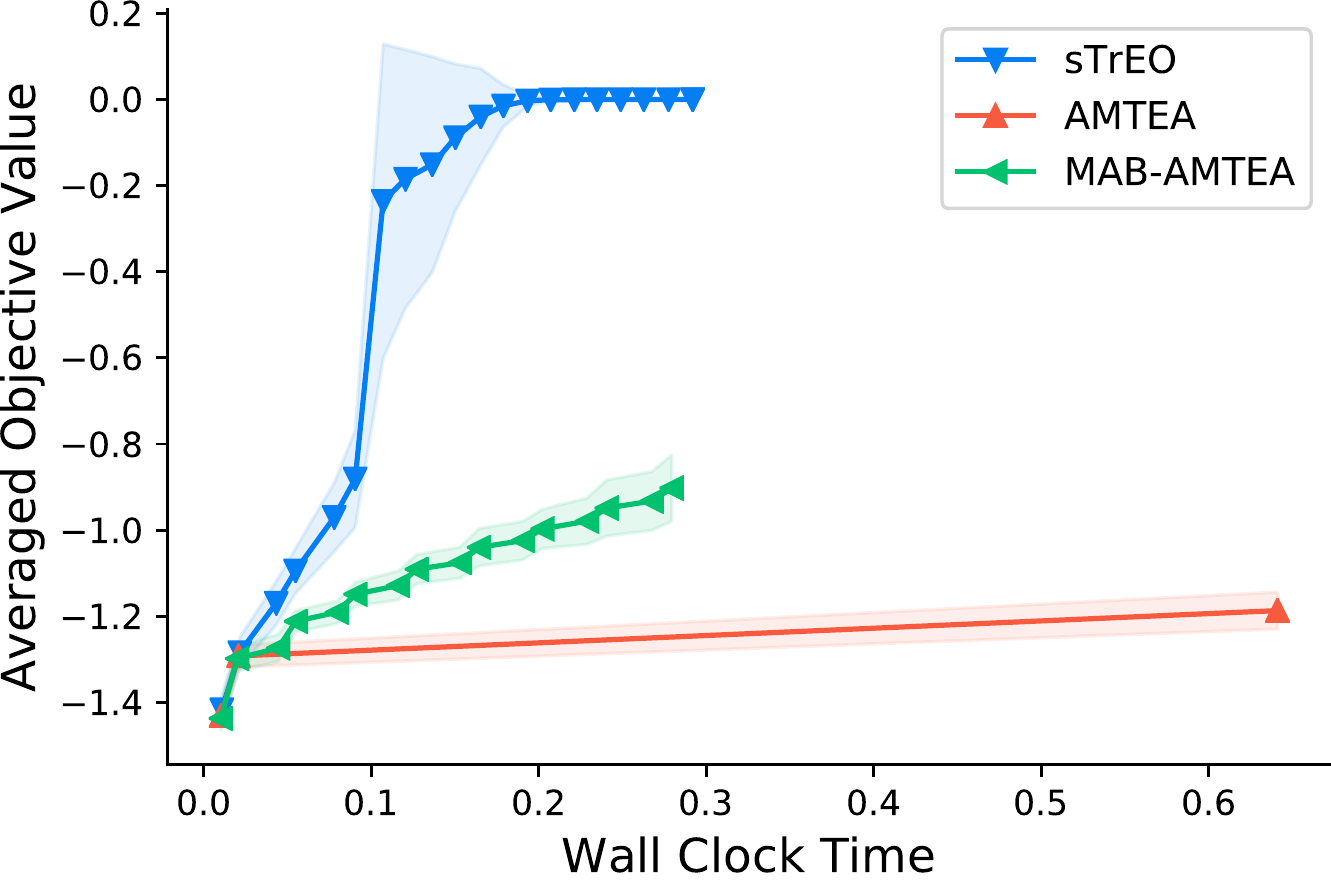}
    \caption{\scriptsize 2000 sources (150 related)}\label{TimeRA2000_30}
 \end{subfigure}
   \begin{subfigure}[b]{0.49\columnwidth}
  \centering
    \includegraphics[width=\textwidth]{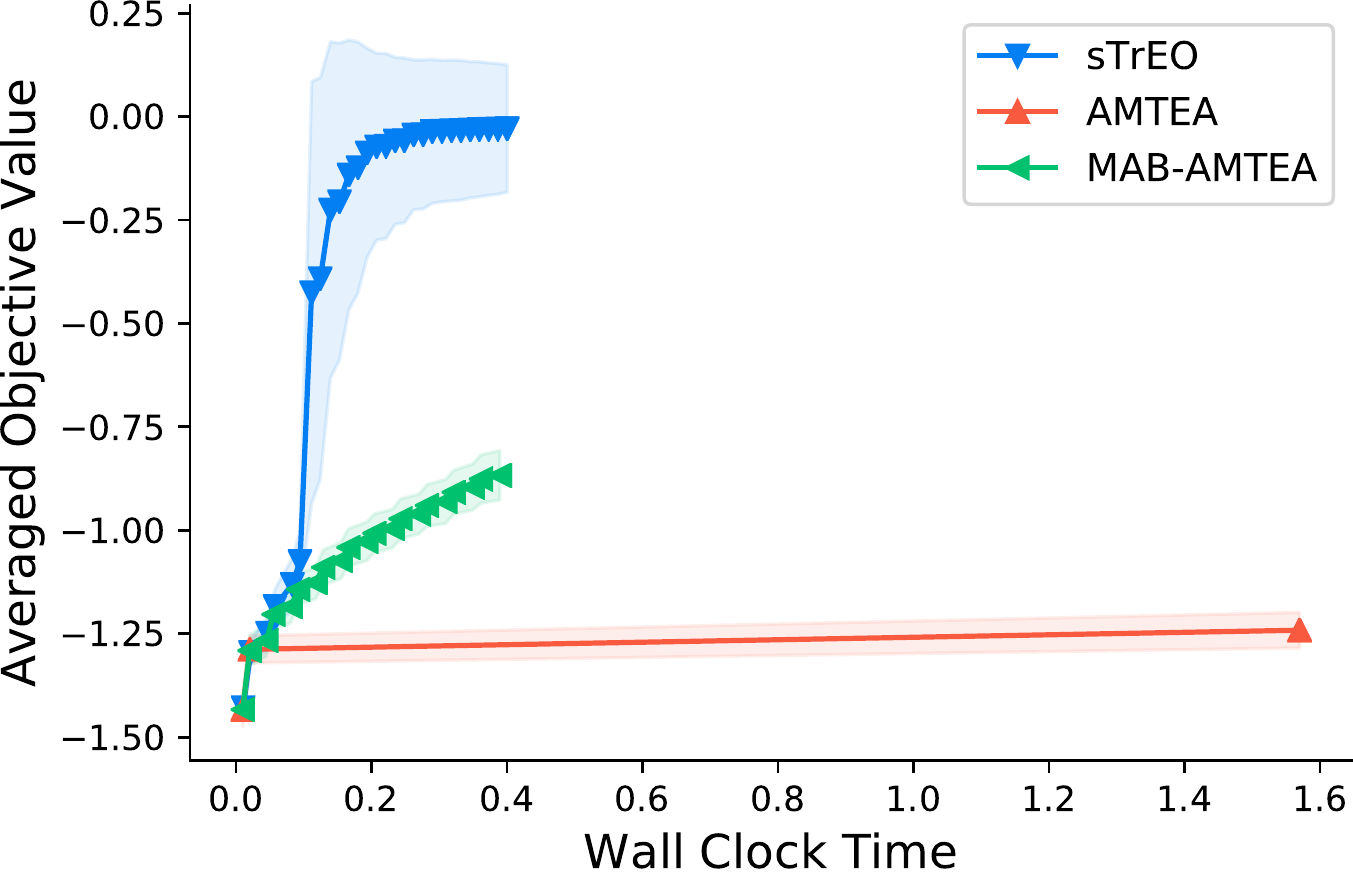}
    \caption{\scriptsize 5000 sources (150 related)}\label{TimeRA5000_75}
 \end{subfigure}
 \begin{subfigure}[b]{0.49\columnwidth}
  \centering
    \includegraphics[width=\textwidth]{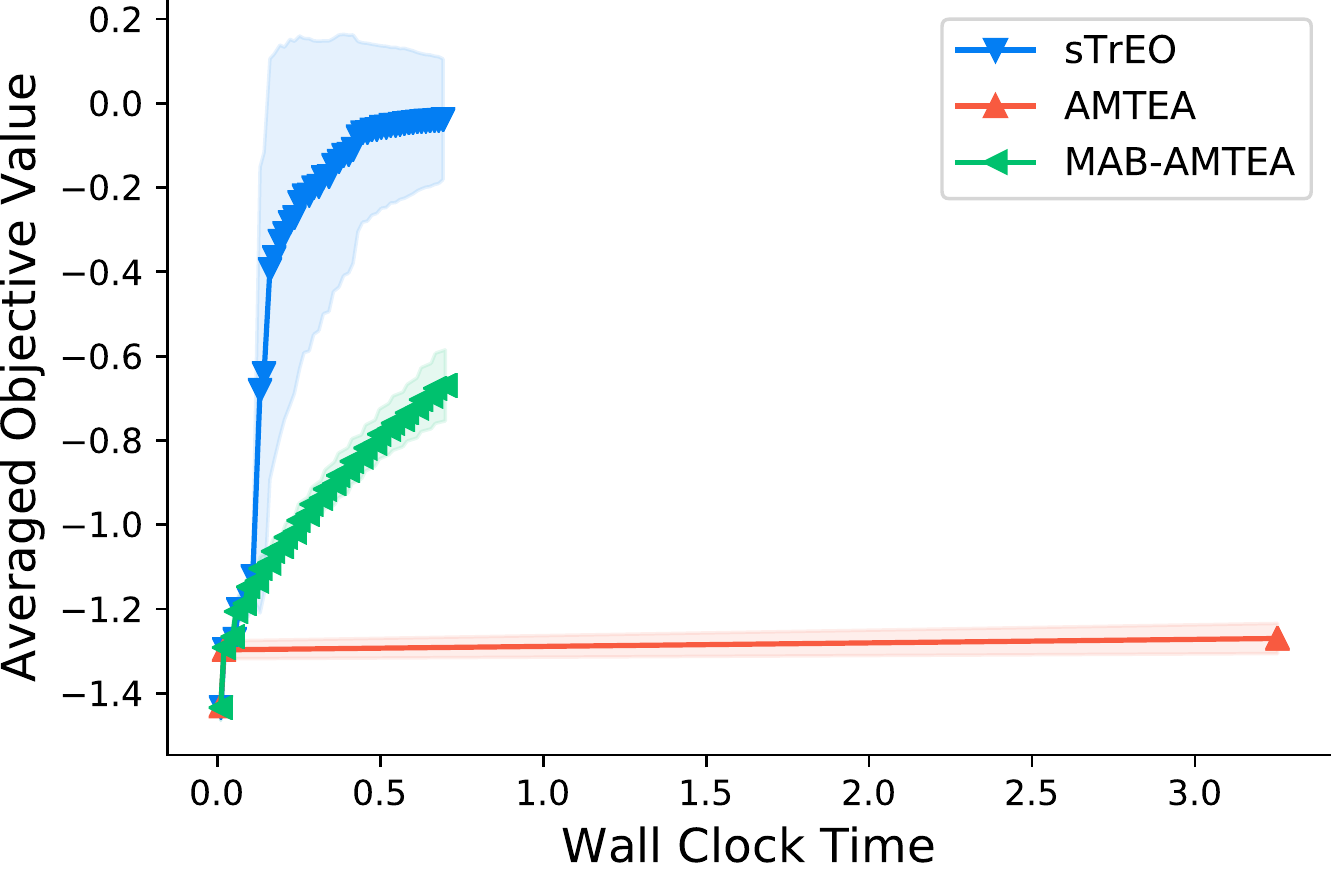}
    \caption{\scriptsize 10000 sources (150 related)}\label{TimeRA10000_150}
 \end{subfigure}
 \caption{Further experiments on \emph{scalability} of sTrEO for the 20-joint robotic arm example with (a) 1000, (b) 2000, (c) 5000 and (d) 10000 sources. The number of related sources is fixed to 150. The shaded region indicates standard deviations either side of the mean.}\label{FigScale}
\end{figure}

Fig. \ref{FigOlnLrnAgl} shows the convergence trends of the three algorithms (averaged over 30 independent runs) with regard to the number of function evaluations, an empirical way to evaluate online learning agility. As can be seen from the figure, despite the growing difference between the number of unrelated and related sources, both AMTEA and sTrEO are able to learn and utilize relevant knowledge in early generations of evolution to converge to optimality. This is however more evident for sTrEO as it acquires a steeper convergence curve, particularly for 5000 and 10000 sources. On the contrary, MAB-AMTEA fails at effectively utilizing relevant sources as it is not able to converge to optimality within the computational budget of 5000 evaluations. With respect to scalability, Fig. \ref{FigScale} shows the convergence trend of the algorithms within sTrEO's convergence time. It can be observed that unlike AMTEA, sTrEO's computational complexity is as efficient as MAB-AMTEA even for the extreme case of 10000 source instances (despite the latter's failure in converging to optimality). AMTEA requires much longer time to converge to optimality. Following the above analysis, we conclude that sTrEO is able to \emph{simultaneously} ensure scalability and online learning agility \emph{beyond} 1000 source task-instances.

\begin{figure}
   \begin{subfigure}[b]{0.49\columnwidth}
  \centering
    \includegraphics[width=\textwidth]{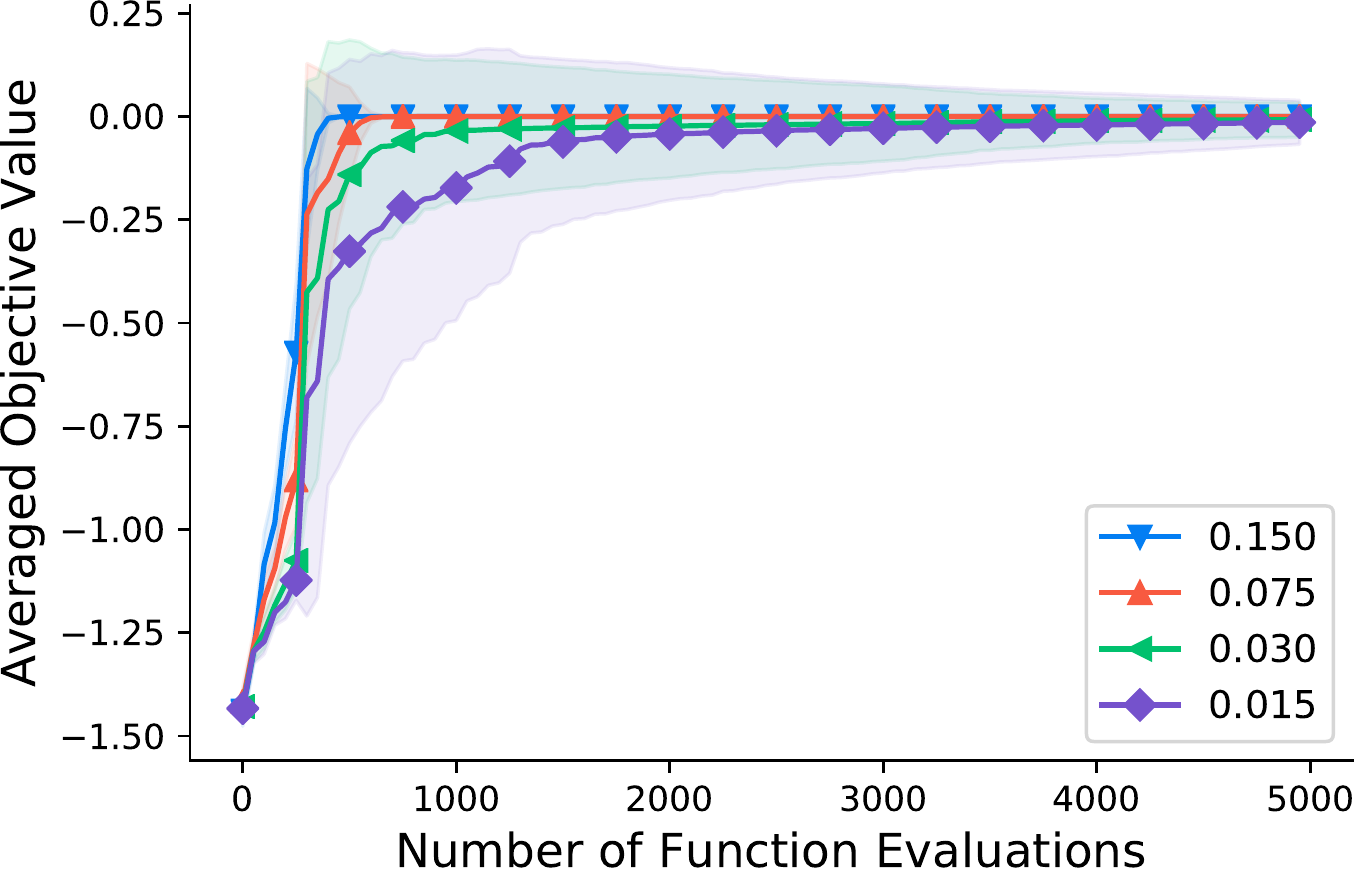}
    \caption{\scriptsize Convergence trends}\label{RelatednessGen}
 \end{subfigure}
 \begin{subfigure}[b]{0.49\columnwidth}
  \centering
    \includegraphics[width=\textwidth]{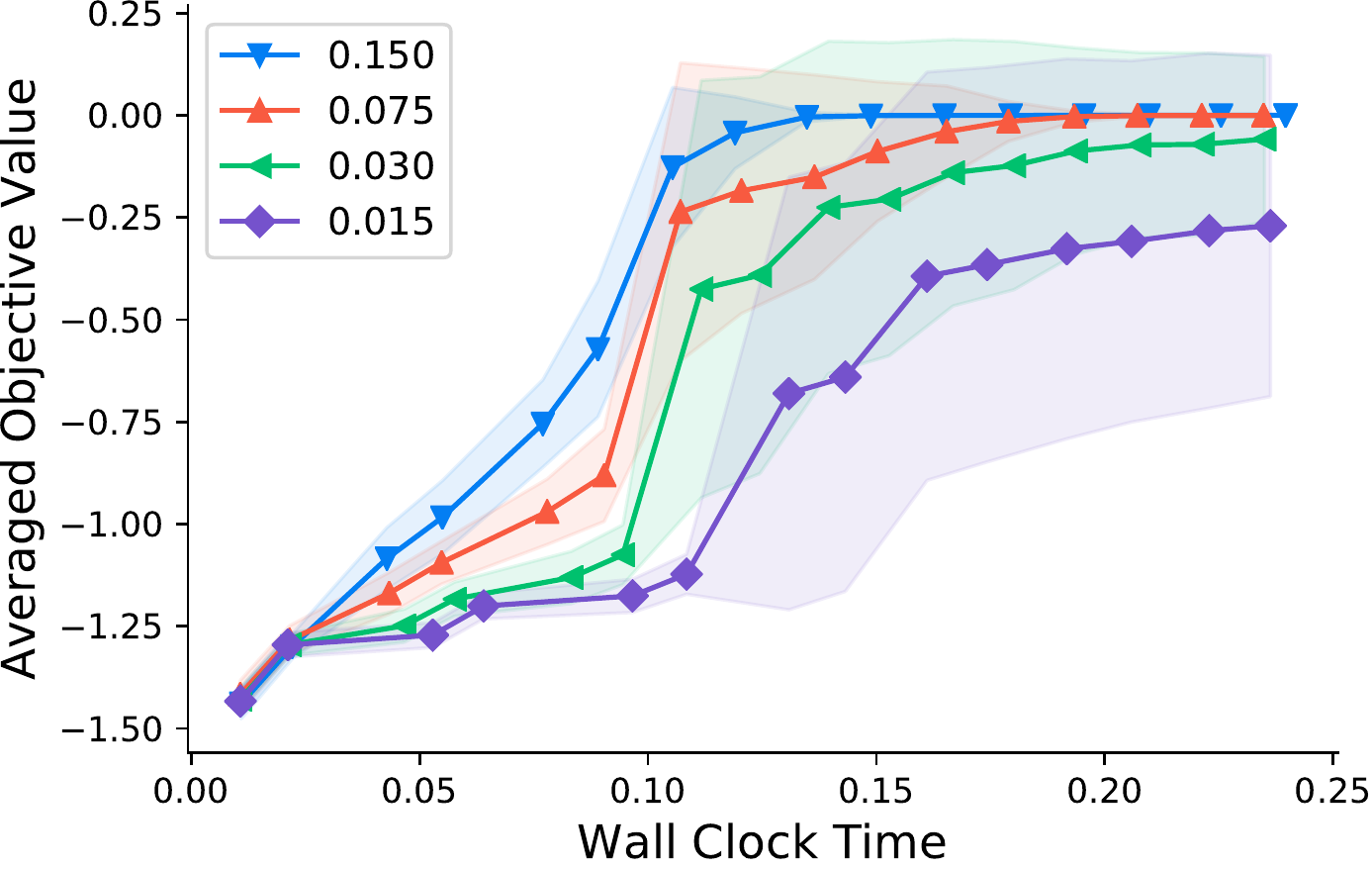}
    \caption{\scriptsize Convergence times}\label{RelatednessTime}
 \end{subfigure}
 \caption{The impact of the ratio of source-target relatedness on sTrEO's performance in terms of (a) number of function evaluations and (b) execution time. The number of related sources is fixed to 150. The shaded region indicates standard deviations either side of the mean.}
    \label{FigRelatedness}
\end{figure}
\begin{figure}
  \begin{subfigure}[b]{0.49\columnwidth}
  \centering
    \includegraphics[width=\textwidth]{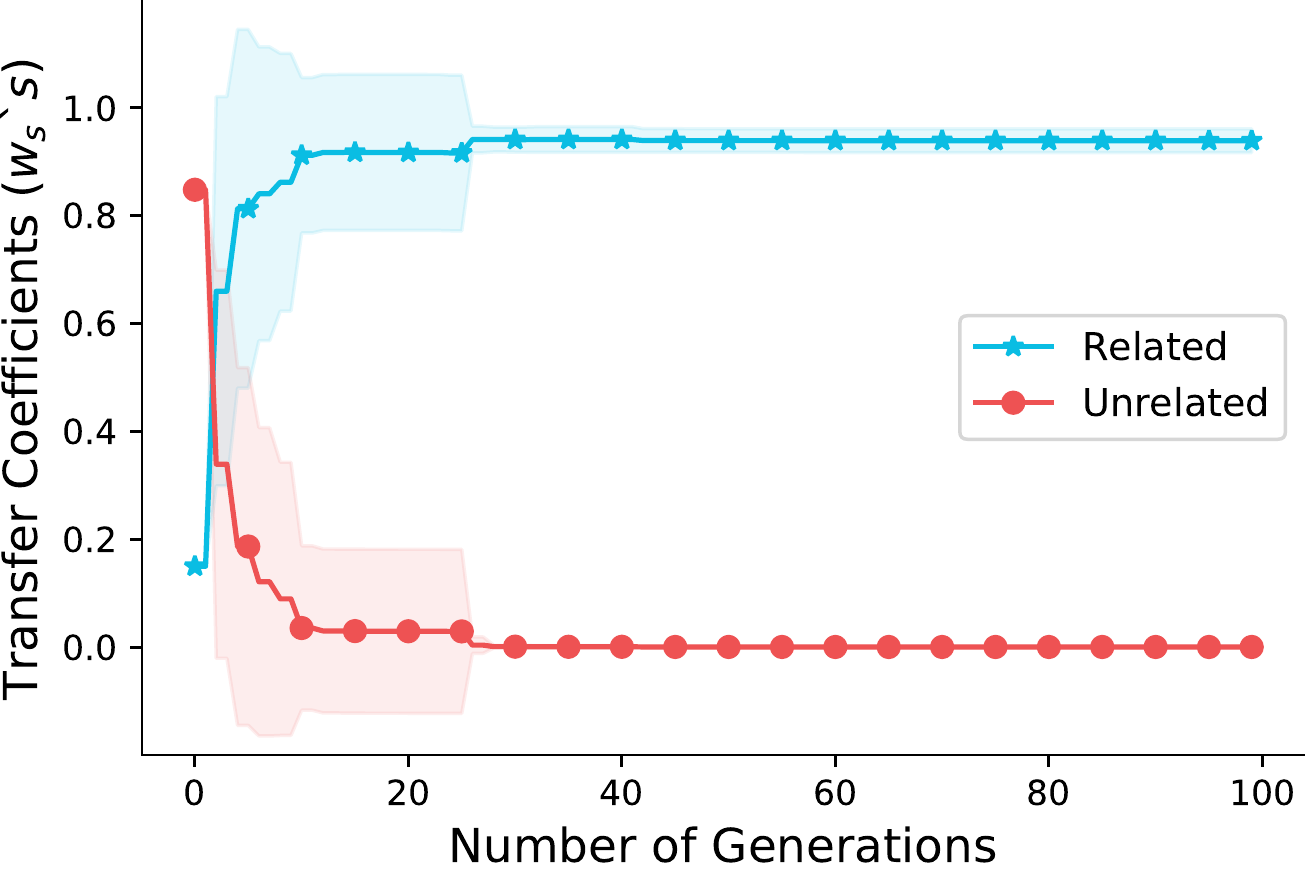}
    \caption{\scriptsize relatedness ratio =$0.15$ }\label{LrnRA1000_15}
 \end{subfigure}
 \begin{subfigure}[b]{0.49\columnwidth}
  \centering
    \includegraphics[width=\textwidth]{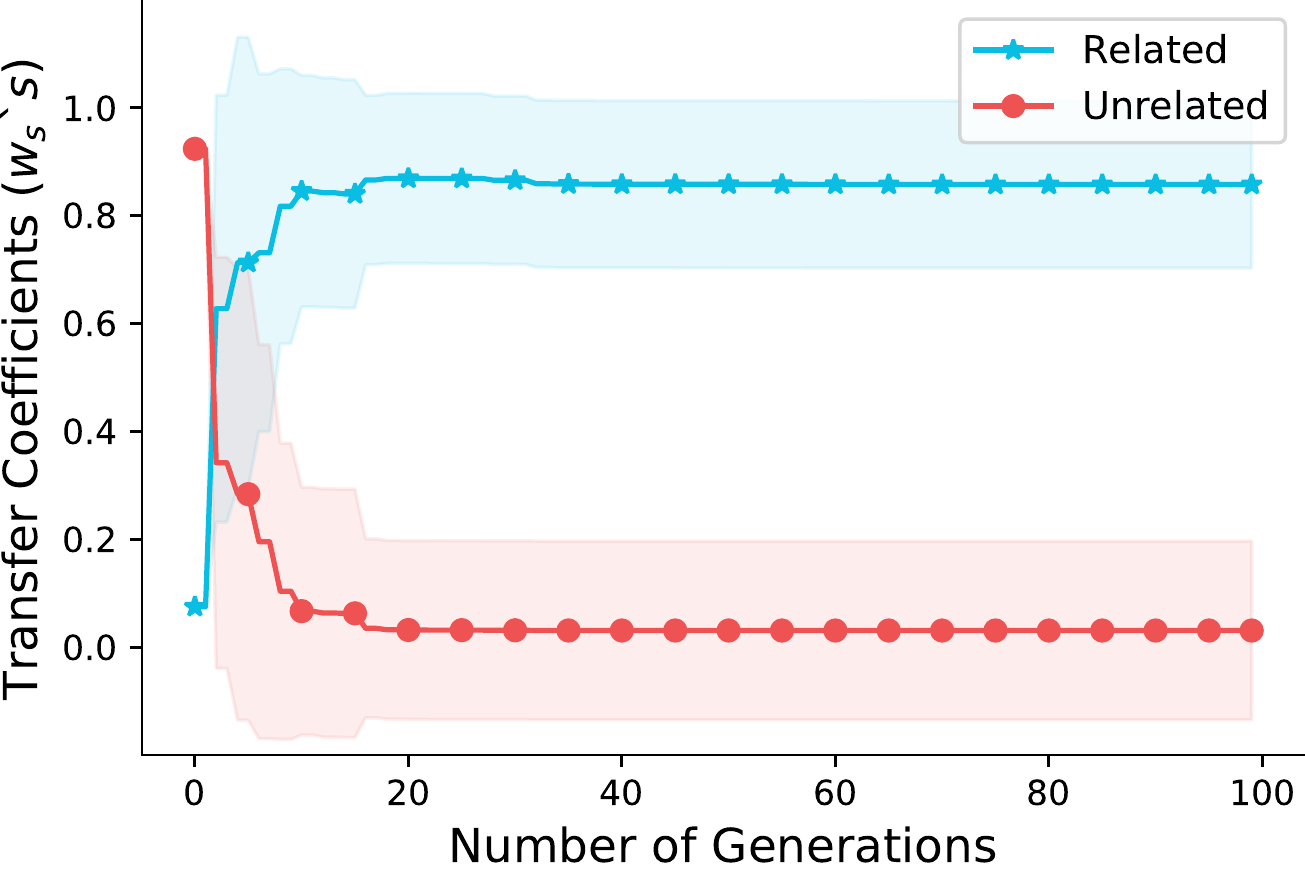}
    \caption{\scriptsize relatedness ratio =$0.075$}\label{LrnRA2000_15}
 \end{subfigure}
   \begin{subfigure}[b]{0.49\columnwidth}
  \centering
    \includegraphics[width=\textwidth]{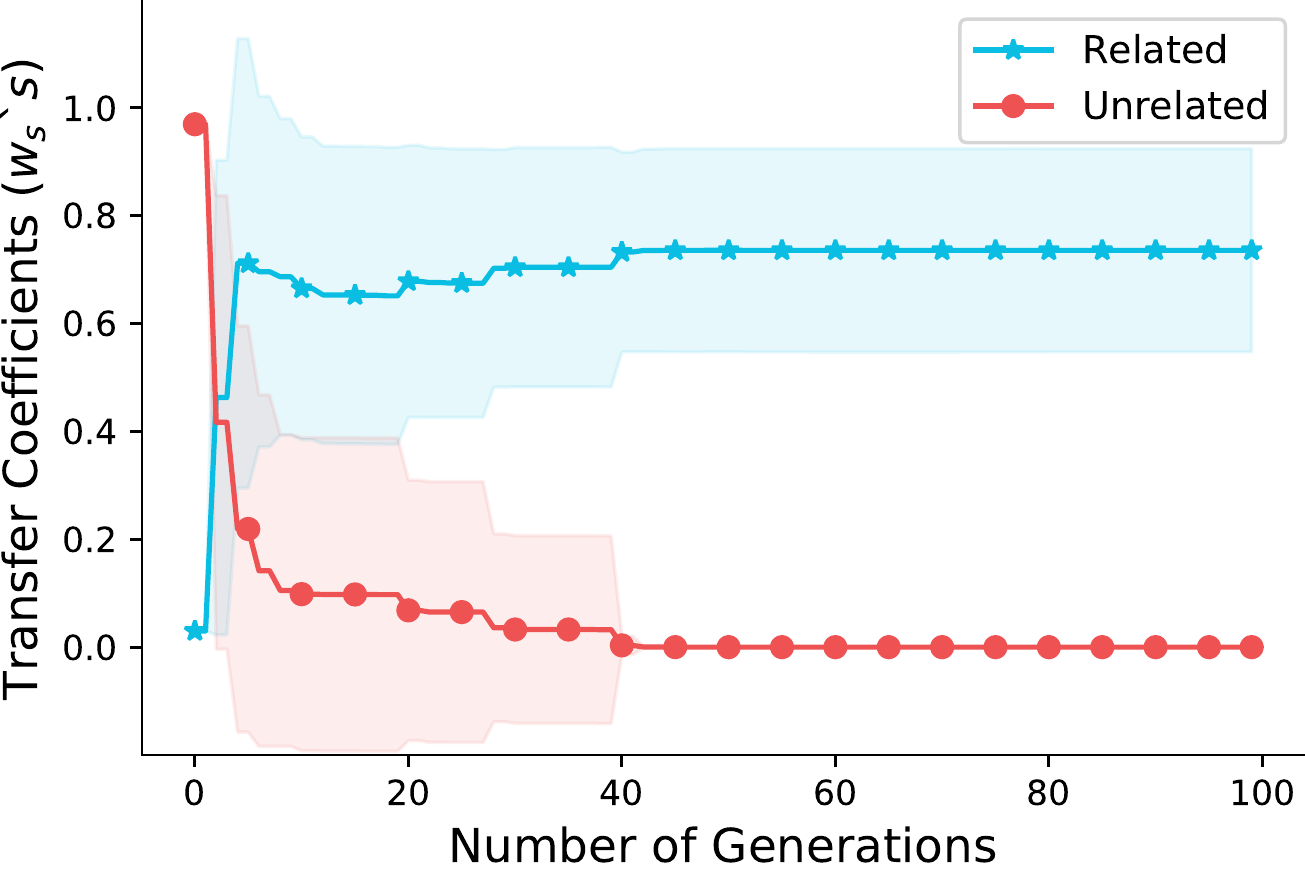}
    \caption{\scriptsize relatedness ratio =$0.03$}\label{LrnRA5000_15}
 \end{subfigure}
 \begin{subfigure}[b]{0.49\columnwidth}
  \centering
    \includegraphics[width=\textwidth]{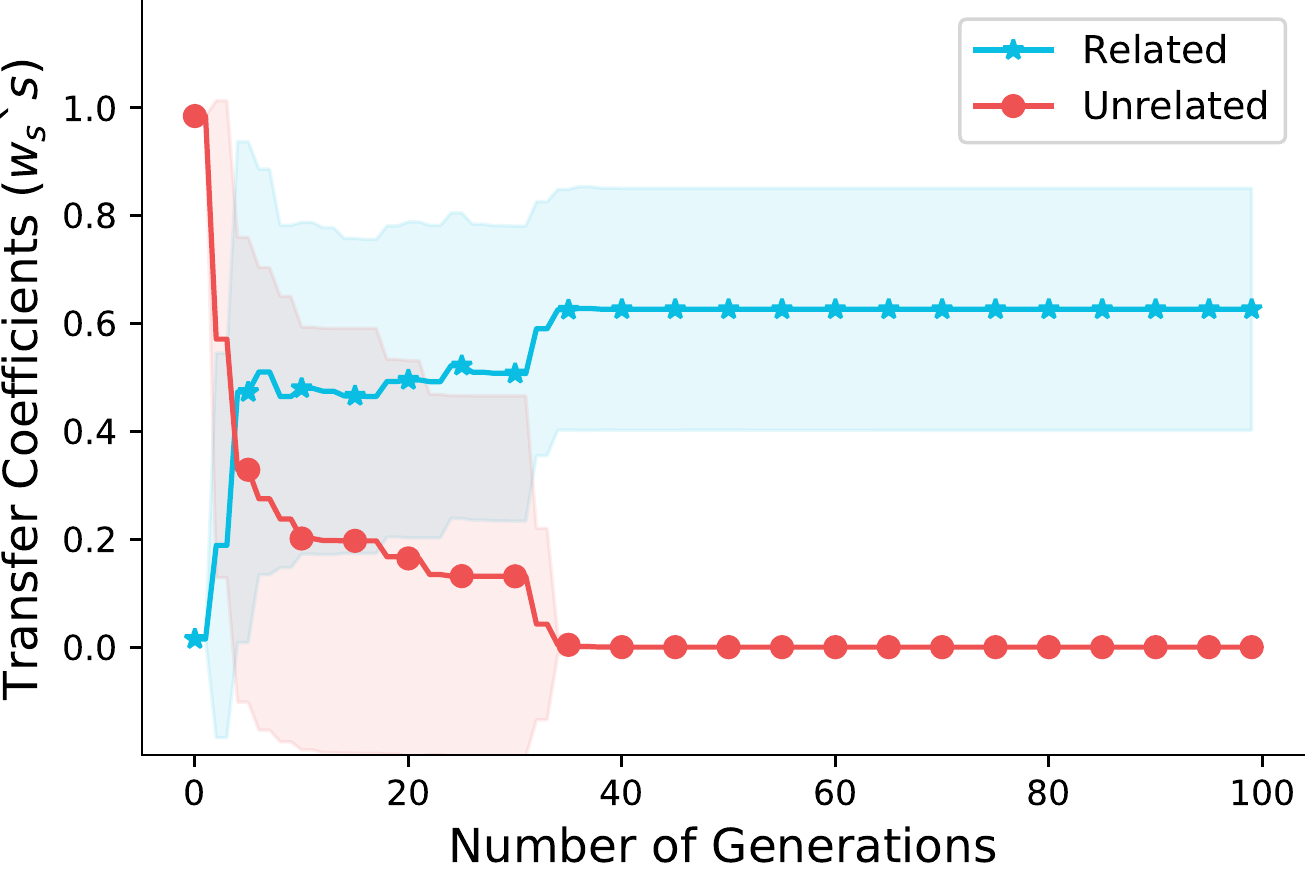}
    \caption{\scriptsize relatedness ratio =$0.015$}\label{LrnRA10000_15}
 \end{subfigure}
 \caption{Learned transfer coefficients of sTrEO given \emph{sparsity} of related sources for the 20-joint robotic arm example with (a) 1000, (b) 2000, (c) 5000 and (d) 10000 sources. The number of related sources is fixed to 150. The shaded region indicates standard deviations either side of the mean.}\label{FigSparsity}
\end{figure}

Fig. \ref{FigRelatedness} shows the impact of the ratio of source-target relatedness on sTrEO's online learning agility, and hence its performance. Fig. \ref{RelatednessGen} depicts the convergence trends of sTrEO relative to the number of function evaluations whereas Fig. \ref{RelatednessGen} presents the convergence trends within sTrEO's convergence time taken for the case of 1000 sources with 150 related instances. As can be seen, sTrEO is sensitive to sparsity of related sources. As the number of unrelated sources increases, the algorithm requires more generations (Fig. \ref{RelatednessGen}), and hence more time (Fig. \ref{RelatednessTime}), to learn related sources, through launching the (1+1)-ES based source-target similarity learning procedure, impeding convergence to optimality. This is better visualized in Fig. \ref{FigSparsity} which depicts the trajectory of learned transfer coefficients aggregated per related and unrelated source instances, respectively, for different source-target relatedness ratios. 



\end{document}